\documentclass{article}
\usepackage{iclr2027_conference,times}
\usepackage[hyphens]{url}
\usepackage{graphicx}
\usepackage{tikz}\usetikzlibrary{arrows.meta,calc,shapes.arrows}   %
\usepackage{caption}
\usepackage{booktabs}
\usepackage{amsmath,amssymb}
\usepackage{xcolor}
\usepackage{colortbl}
\usepackage{multirow}
\usepackage[colorlinks=true,citecolor=blue!60!black,linkcolor=red!60!black,urlcolor=blue!60!black]{hyperref}
\usepackage{placeins}   %

\newcommand{\tightbox}[2]{{\setlength{\fboxsep}{1pt}\colorbox{#1}{#2}}}
\definecolor{tableBest}{HTML}{E5EEF5}
\newcommand{\bestcell}[1]{\cellcolor{tableBest}\textbf{#1}}
\newcommand{\secondcell}[1]{\underline{#1}}

\title{PhysLDM: Latent Diffusion for \\ High-Fidelity Deformable Simulation}

\author{
\vspace{2cm}
\parbox{0.95\textwidth}{%
\centering
\textbf{Yu Zhang}$^{1}$
\hspace{1.5cm}
\textbf{Xudong Xu}$^{2}$
\hspace{1.5cm}
\textbf{Xingang Pan}$^{1}$ \\[3pt]
\normalfont
$^{1}$S-Lab, Nanyang Technological University, Singapore \\
$^{2}$Shanghai Artificial Intelligence Laboratory, Shanghai, China \\[2pt]
\texttt{\{yuzhang,xingang.pan\}@ntu.edu.sg}
\qquad
\texttt{xudongxu9710@gmail.com}
\vspace{-2cm}
}
}

\iclrfinalcopy %

\hypersetup{pdftitle={PhysLDM: Latent Diffusion for High-Fidelity Deformable Simulation},pdfauthor={Yu Zhang, Xudong Xu, Xingang Pan}}
\begin{document}
\addtocontents{toc}{\protect\setcounter{tocdepth}{-1}}

\maketitle
\lhead{}\renewcommand{\headrulewidth}{0pt}

\begin{abstract}
Neural simulation of high-fidelity deformable bodies is a foundational challenge in computer graphics and physical AI. 
Long-horizon prediction for high-resolution 3D volumetric meshes is difficult: autoregressive methods are susceptible to error accumulation, while direct multi-frame prediction at native resolution is computationally prohibitive.
This motivates a compact spatiotemporal latent representation, which remains largely unexplored for mesh-based volumetric physics. Meanwhile, it remains unclear whether deterministic regression or generative diffusion is the more appropriate predictive paradigm.
To address these coupled challenges, we introduce PhysLDM, a unified latent-diffusion paradigm for one-shot volumetric deformable simulation. 
Its core is a holistic spatiotemporal VAE (ST-VAE) that avoids the ``staircase'' artifacts of standard temporal compression (as in common video VAEs) and disentangles rigid motion from local deformation, achieving $\sim$2.48 mm reconstruction precision on meter-scale scenes at up to 78$\times$ token compression.
Based on this reliable latent space, we systematically compare regression and diffusion paradigms. Our experiments uncover a key modeling insight: complex deformable dynamics are often chaotic, and in this regime deterministic regression tends to produce damped, non-physical averages, whereas diffusion better models their distribution. Accordingly, we employ a latent diffusion model that effectively learns from the chaotic data to generate physically plausible trajectories. 
Trained purely kinematically on an Objaverse-scale simulation dataset, a single PhysLDM generalizes zero-shot to unseen OOD object datasets (GSO and Toys4K).
Its differentiability further enables efficient solution of inverse problems and higher-order design optimization.
To our knowledge, PhysLDM is the first high-fidelity spatiotemporal autoencoder and latent-diffusion paradigm for volumetric deformable dynamics, offering a scalable and robust approach to neural simulation. Project Page: \href{https://yucrazing.github.io/physldm/}{https://yucrazing.github.io/physldm/}
\end{abstract}

\begin{figure*}[t]
\centering
\input{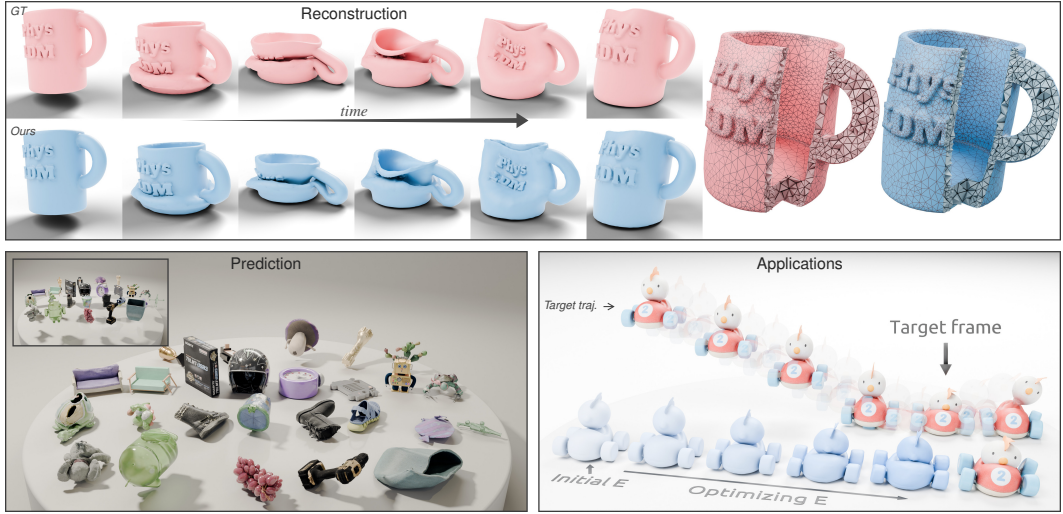}%
\caption{\textbf{PhysLDM overview.} \emph{Top, Reconstruction:} Ground truth (rose) versus ST-VAE reconstruction (blue) for an unseen high-resolution soft mug (11{,}621 nodes, ${>}2\times$ the training cap). The trajectory is compressed by ${\sim}182\times$ with only a 5.3\,mm mean error. Far right: a cross-section of the final frame visualizing the internal tetrahedral structure. \emph{Bottom Left, Prediction:} One-shot prediction of 64-frame trajectories for diverse out-of-distribution (OOD) objects (GSO/Toys4K) across varying materials and initial velocities (inset: initial state; main view: predicted settled state). \emph{Bottom Right, Differentiable Applications:} A classic material estimation task: recovering Young's modulus $E$ (blue sequence) by matching a target state (textured) via gradient descent. Furthermore, designing the throw that best reveals the material is a second-order inverse task (Sec.~\ref{sec:diffapp}). PhysLDM easily solves it with native support for higher-order derivatives. By contrast, all other differentiable simulators we tested either fail or are computationally prohibitive on this task.}
\label{fig:teaser}
\end{figure*}

\section{Introduction}

Simulating the dynamics of deformable solids is essential to fields involving physical interaction, including computer graphics, engineering, and robotics. Furthermore, learning these complex dynamics as predictive physical world models is also considered essential for physical AI. However, fulfilling this vision with traditional simulators remains a fundamental bottleneck: classical high-precision solvers are accurate but slow, and their gradients are hard to obtain. Neural simulation is an emerging alternative: its goal is to learn these complex physical behaviors directly from data, offering fast, differentiable surrogates that bypass the computational bottlenecks of classical solvers \citep{DBLP:conf/icml/Sanchez-Gonzalez20,DBLP:conf/iclr/PfaffFSB21}. Nevertheless, high-fidelity deformable simulation demands a high-resolution geometric representation, typically a volumetric mesh with thousands to tens of thousands of vertices. Efficiently and accurately predicting the long-horizon trajectories of such large-scale systems poses significant challenges for conventional neural architectures.

Achieving this, however, presents several tightly coupled challenges.
First, high-fidelity 3D volumetric meshes contain large, object-dependent numbers of vertices, resulting in extremely high-dimensional system states.
Existing autoregressive (AR) approaches predict the dynamics frame by frame, without jointly modeling the full temporal horizon, and are therefore susceptible to compounding errors and trajectory drift over long rollouts.
One-shot prediction of the entire trajectory provides global temporal context, but operating directly at native resolution is computationally prohibitive.
Consequently, compressing these complex dynamics into a spatiotemporal latent space becomes necessary, yet largely unexplored for mesh-based volumetric physics. This introduces the first challenge: a representational bottleneck. 
In our explorations, we find that standard compression methods (\textit{e.g.}, with uniform temporal strides and local temporal upsamplers, as in standard video-VAE designs \citep{DBLP:conf/cvpr/YuCSL0CH0HE023}) lose high-frequency temporal information and make reconstructed shapes jump unnaturally, producing a ``staircase'' artifact that we analyze in Sec.~\ref{sec:staircase} (Staircase analysis).
Furthermore, directly compressing raw world coordinates forces models to entangle large-scale rigid motion with millimeter-scale local deformations, severely degrading precision.

The second challenge is the optimal predictive paradigm: it remains an open question whether deterministic regression or generative diffusion is better suited for complex physical simulation, and a systematic comparison in the context of 3D volumetric physics is currently absent. Our systematic experiments reveal a critical insight: microscopic non-determinism can be strongly amplified by the inherently chaotic nature of complex deformable dynamics.
For instance, state-of-the-art GPU-based physical solvers \citep{DBLP:journals/tog/HuangCLK24,DBLP:journals/tog/HuangLLKL25} exhibit only slight microscopic non-determinism, yet chaos can amplify this small difference into vastly different final trajectories (Sec.~\ref{sec:chaos}, Chaos analysis). 
In this chaotic regime, nearly identical initial conditions can correspond to multiple macroscopically different yet physically plausible futures. Deterministic regression tends to average over these modes, producing damped trajectories that may match none of them, whereas generative diffusion can represent and sample from the multi-modal distribution of plausible outcomes.

To overcome these tightly coupled representational and predictive limitations, we propose PhysLDM, a latent-diffusion paradigm for one-shot volumetric deformable simulation. The foundation of PhysLDM is a novel holistic spatiotemporal variational autoencoder (ST-VAE) specifically designed for mesh-based volumetric physics. By employing a holistic temporal compression strategy and explicitly decoupling rigid body motion from local deformation, our ST-VAE eliminates ``staircase'' artifacts and preserves fine geometric details. This design achieves a $\sim$2.48\,mm average geometric reconstruction precision for meter-scale scenes at up to a $78\times$ token compression. By employing a Latent Diffusion Model (LDM) \citep{DBLP:conf/cvpr/RombachBLEO22}, PhysLDM models the distribution of physically plausible future trajectories, whereas deterministic regression produces damped averages in our experiments.

A prominent feature of our paradigm is its constitutive-model-agnostic nature. Supervised purely kinematically on an Objaverse-scale dataset, a single PhysLDM demonstrates strong zero-shot generalization capabilities to unseen, out-of-distribution (OOD) datasets. We further demonstrate the utility of PhysLDM as a \textit{differentiable simulator} for inverse problems and design optimization. To our knowledge, PhysLDM yields the first high-fidelity spatiotemporal autoencoder and latent-diffusion paradigm for volumetric deformable dynamics, offering a scalable and robust approach to neural simulation.

In summary, our main contributions include:
\begin{itemize}
\item \textbf{A high-fidelity ST-VAE for 3D volumetric deformable dynamics.} We propose a novel holistic temporal compression scheme for volumetric mesh deformable dynamics, which removes the ``staircase'' artifacts that persist under standard temporal compression. By explicitly decoupling rigid motion and deformation, we achieve \(\sim\)2.48mm average geometric reconstruction precision for meter-scale scenes with up to 78$\times$ token compression.
\item \textbf{A systematic comparison of deterministic regression and generative diffusion paradigms in physical simulation.} We reveal that the inherently chaotic nature of complex deformable dynamics can amplify the non-determinism that GPU solvers exhibit under identical initial conditions, and show that in this chaotic regime diffusion is better suited than deterministic regression, whose outputs collapse toward damped averages.
\item \textbf{A constitutive-model-agnostic latent diffusion paradigm with strong zero-shot generalization.} We present a latent-diffusion paradigm and train a single model, PhysLDM, on Objaverse-scale simulations \citep{DBLP:conf/cvpr/DeitkeSSWMVSEKF23} supervised purely kinematically. Our model transfers zero-shot to unseen OOD object datasets (GSO \citep{DBLP:conf/icra/DownsFKKHRMV22}, Toys4K \citep{DBLP:conf/cvpr/StojanovTR21}) with remarkable robustness.
\item \textbf{A differentiable simulator for inverse problems and design optimization.}
PhysLDM supports end-to-end and higher-order differentiation, enabling efficient material estimation and second-order design optimization tasks that are difficult or computationally prohibitive for existing differentiable simulators.
\end{itemize}

\section{Related Work}

\textbf{Physics-based deformable simulation.}
Physics-based Lagrangian simulation has a rich history in computer graphics, evolving from early elastic models \citep{DBLP:conf/siggraph/TerzopoulosPBF87} to finite element methods (FEM) \citep{DBLP:conf/sca/IrvingTF04}, contact handling \citep{DBLP:journals/tog/BridsonFA02}, and others \citep{DBLP:journals/jvcir/MullerHHR07,DBLP:journals/tog/BouazizMLKP14}. Recently, barrier-based methods such as incremental potential contact (IPC) provide strict intersection-free guarantees, but incur high computational costs \citep{DBLP:journals/tog/LiFSLZPJK20,DBLP:journals/tog/HuangCLK24,DBLP:journals/tog/HuangLLKL25,DBLP:journals/tog/TangHBLL26,DBLP:journals/tog/Ando24}. Conversely, real-time solvers achieve interactive rates at the expense of physical fidelity, and they cannot learn the behavior of high-accuracy solvers \citep{DBLP:journals/tog/ChenLYY24,DBLP:journals/tog/GilesDY25}. Differentiable simulators restore gradients, but either require tedious hand-derived formulations or need to backpropagate through every integration step \citep{DBLP:journals/tog/HuangTGFSPZ24,newton2025}. In contrast, PhysLDM serves as a fast, end-to-end differentiable neural surrogate learned directly from high-accuracy solvers.

\textbf{Neural simulators for deformable dynamics.}
Neural simulators provide a fast, fully differentiable alternative to traditional solvers. The dominant paradigm of neural simulators is the AR rollout, which predicts the next state frame-by-frame. This paradigm has been adopted across diverse physical simulation fields \citep{DBLP:conf/icml/Sanchez-Gonzalez20,DBLP:conf/iclr/UmmenhoferPTK20,DBLP:conf/iclr/PfaffFSB21,zhang2026clothtransformer,DBLP:journals/corr/abs-2605-15305}. However, AR methods lack the global temporal context necessary for long-horizon consistency, potentially leading to temporal incoherence or trajectory drift. The concurrent work of \citet{DBLP:journals/corr/abs-2606-27364} performs one-shot multi-frame prediction, but is restricted to low-resolution surface meshes (training on about $356$ vertices per scene) from a limited set of templates ($35$ different meshes). In contrast, PhysLDM introduces a latent-diffusion paradigm capable of predicting high-resolution volumetric trajectories in a single diffusion pass, demonstrating strong zero-shot generalization to diverse OOD scenes.

\textbf{Spatiotemporal latent representations.}
High-fidelity volumetric simulation involves high-resolution meshes, making direct multi-frame prediction computationally prohibitive at native resolutions. However, spatiotemporal autoencoders remain largely unexplored for high-fidelity, volumetric-mesh deformable-body physical simulation. In fact, existing autoencoders for physical simulation typically compress only the \emph{spatial} domain \citep{DBLP:journals/tog/BarbicJ05,DBLP:journals/cgf/FultonMDLJ19,DBLP:journals/tog/RomeroCPO21,DBLP:conf/iclr/ChenXCCPMCCG23,DBLP:conf/siggrapha/ChangCWCCG23,DBLP:journals/tog/ModiSPSL24,DBLP:conf/siggraph/LiaoSP26,DBLP:journals/cgf/WiewelBT19,DBLP:conf/iclr/HanGPWL22}, with concurrent neural simulators following a similar \mbox{pattern} \citep{zhang2026clothtransformer,DBLP:journals/corr/abs-2605-15305,DBLP:journals/corr/abs-2605-30347}. While spatiotemporal autoencoders have been successfully deployed in other domains, their architectures are tailored to their respective tasks, such as video generation \citep{DBLP:conf/cvpr/YuCSL0CH0HE023,DBLP:conf/iclr/YangTZ00XYHZFYZ25} and other fields \citep{DBLP:conf/iccv/PetrovichBV21,DBLP:conf/cvpr/CaoL0NT24}. Unfortunately, these designs do not transfer directly to deformable physics. For example, transplanting the standard temporal compression of video VAEs produces severe staircase artifacts (as analyzed in Sec.~\ref{sec:staircase}, Staircase analysis). To our knowledge, our ST-VAE is the first spatiotemporal autoencoder designed specifically for high-fidelity deformable simulation.

\textbf{Prediction paradigms: regression versus diffusion.}
For deformable-body physical simulation, deterministic regression and generative diffusion have never been systematically compared. In practice, most work in neural simulation uses regression \citep{DBLP:conf/icml/Sanchez-Gonzalez20,DBLP:conf/iclr/PfaffFSB21,zhang2026clothtransformer,DBLP:journals/corr/abs-2605-09196,DBLP:journals/corr/abs-2605-15305}. Meanwhile, diffusion \citep{DBLP:conf/nips/HoJA20} has instead been adopted where output distributions matter, such as image and video synthesis \citep{DBLP:conf/cvpr/RombachBLEO22,DBLP:conf/nips/HoSGC0F22} and human motion \citep{DBLP:conf/iclr/TevetRGSCB23}. Although the concurrent work of \citet{DBLP:journals/corr/abs-2606-27364} claims that diffusion is superior to deterministic regression, motivated by the diversity of futures under unobserved initial conditions, it never compares one-shot regression with diffusion. By contrast, we systematically compare both paradigms and find that under the chaotic dynamics of complex deformable bodies diffusion is more suitable, and regression may collapse onto non-physical averages.

\section{Methodology}
\label{sec:method}

\subsection{Problem Formulation}
\label{subsec:formulation}

We formulate the physical simulation of volumetric objects as a trajectory prediction task. Let a scene be represented as a tetrahedral mesh with rest vertices $\mathbf{r} \in \mathbb{R}^{N \times 3}$. The initial state of the system is defined as $\mathbf{s}_0 = (\mathbf{r}, \mathbf{R}_0, \mathbf{c}_0, \mathbf{v}_0, \boldsymbol{\omega}_0, E)$, which encompasses the rest geometry, the frame-$0$ pose, linear velocity, and angular velocity, together with the Young's modulus $E$ of the material. Our goal is to model the distribution $p(\mathbf{X} \mid \mathbf{s}_0)$ of future vertex trajectories $\mathbf{X} \in \mathbb{R}^{T \times N \times 3}$ over a horizon of $T$ frames. Rather than modeling the raw world coordinates directly, we factor each frame into a global rigid transform acting on a canonical deformation:
\begin{equation}
\mathbf{x}_t = \mathbf{R}_t\,(\mathbf{r} + \mathbf{d}_t) + \mathbf{c}_t,
\label{eq:factoring}
\end{equation}
where $\mathbf{R}_t \in \mathrm{SO}(3)$ and $\mathbf{c}_t \in \mathbb{R}^3$ denote the per-frame rigid rotation and centroid translation, and $\mathbf{d}_t$ represents the residual canonical (rigid-free) deformation field. $\mathbf{R}_t$ and $\mathbf{c}_t$ are extracted from each ground-truth frame in closed form by a per-frame Procrustes (Kabsch) fit of the rest shape \citep{DBLP:journals/pami/Umeyama91}. This factoring separates the large-scale rigid motion from the small-scale deformation. The rigid transform $(\mathbf{R}_t, \mathbf{c}_t)$ carries the global motion in six degrees of freedom per frame. The canonical deformation field $\mathbf{d}_t$ then only has to capture the local deformation, which is independent of the object's pose.

\subsection{Architecture Overview}
\label{sec:arch}
\begin{figure*}[t]
  \centering
  \includegraphics[width=\textwidth]{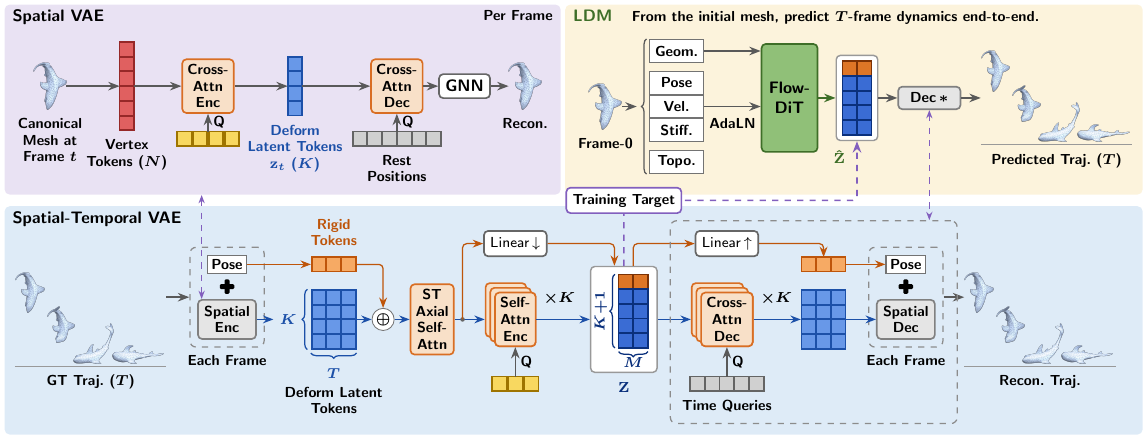}
  \caption{\textbf{Overview of PhysLDM.}
  The framework consists of three main components:
  (1) A \textbf{Spatial VAE} (top left) that encodes the per-frame canonical mesh into a compact set of resolution-independent latent tokens.
  (2) An \textbf{ST-VAE} (bottom) that compresses the temporal trajectory into a single joint latent $\mathbf{Z}$, which serves as the training target of the LDM.
  (3) An \textbf{LDM} (top right) that predicts the entire latent trajectory from the initial state in a single diffusion pass. \textsf{Dec}\,$\ast$ denotes the frozen decoders, which decode the latent into the final full trajectory.}
  \label{fig:framework}
\end{figure*}

As shown in Figure~\ref{fig:framework}, PhysLDM has three main modules: a Spatial VAE, an ST-VAE, and an LDM.

\textbf{Spatial VAE.}
To efficiently process high-resolution volumetric meshes of varying vertex counts, we compress individual frames using a Spatial VAE. Following the representation of \citet{zhang2026clothtransformer}, a set of $K$ learnable query tokens cross-attends the canonical mesh vertices to extract a fixed-size set of $K$ latent tokens $\mathbf{z}_t$, which remains independent of the vertex count $N$. The decoder mirrors the encoder: each rest vertex, embedded with its position, cross-attends the latent tokens and is decoded to its displacement. This module ensures that the latent representation is resolution-independent.

\textbf{ST-VAE.}
The ST-VAE compresses the input $T$-frame trajectory by a factor of $T/M$ into a compact latent $\mathbf{Z}$. Each frame is represented by $K{+}1$ tokens: the $K$ deformation tokens from the spatial VAE and one additional rigid token that embeds the pose ($\mathbf{R}_t$ and $\mathbf{c}_t$) in its channels. This token sequence first passes through axial blocks that alternate self-attention across tokens within a frame and along the temporal axis \citep{DBLP:journals/corr/abs-1912-12180,DBLP:conf/icml/BertasiusWT21,DBLP:conf/iccv/Arnab0H0LS21}, after which it is decoupled into a deformation track and a rigid track. For the deformation track, the standard choice (as in video VAEs) would merge every $T/M$ frames into one latent frame and then decode it with a local temporal upsampler. However, this standard design introduces the staircase artifacts analyzed in Sec.~\ref{sec:staircase} (Staircase analysis). We instead employ a holistic temporal representation where time is summarized by factorized self-attention: for each of the $K$ spatial tokens separately, $M$ learnable queries are prepended to its temporal sequence, attended along the time axis, and kept as the only temporal outputs, giving $K\times M$ tokens in total, which we call holistic temporal tokens. For the rigid track, sharing the holistic tokens with the deformation causes capacity competition and degrades the reconstruction (see ablation in Sec.~\ref{sec:ablations}). The rigid token therefore takes a dedicated linear temporal pathway and recombines with the canonical shape per frame during decoding. Ultimately, the two tracks merge into the single joint latent $\mathbf{Z}$ that serves as the training target of the LDM. The ST-VAE decoder is kept frozen during the LDM stage to decode latents back into full trajectories.

\textbf{LDM.}
Conditioned on the initial state, the LDM predicts the latent of the full $T$-frame trajectory in one pass. The frame-$0$ mesh is injected via cross-attention on its spatial latent, while the pose, the initial velocity, and the stiffness condition the network via adaptive layer normalization (AdaLN) alongside the flow timestep. The mesh topology enters only through the frozen decoder, whose graph neural network (GNN) refines the decoded vertices over the tetrahedral mesh. We employ a flow-matching \citep{DBLP:conf/iclr/LipmanCBNL23} diffusion transformer (DiT) \citep{DBLP:conf/iccv/PeeblesX23} utilizing rectified flow \citep{DBLP:conf/iclr/LiuG023} to model the probability path. Architecture details are in Appendix~\ref{app:arch}.

\subsection{Training}
\label{sec:training}
Our model is trained in three stages. The first stage trains the spatial VAE with a Charbonnier (smoothed $L_1$) position-reconstruction loss and a Kullback--Leibler (KL) loss (see the spatial-VAE-pretraining ablation in Appendix~\ref{app:moreabl}). The second stage trains the ST-VAE on top of the pretrained spatial VAE. Its loss consists mainly of deformation and rigid reconstruction losses over all $T$ frames and a KL loss, together with a latent-anchor loss to the spatial manifold. The third stage trains the LDM on the pretrained, frozen ST-VAE, with a flow-matching velocity mean-squared error (MSE) loss. Complete loss forms, weights, and training details are given in Appendix~\ref{app:arch}.

\textbf{Constitutive-Model-Free Supervision.}
Notably, our framework relies entirely on kinematic supervision. Every reconstruction loss is computed directly from geometry, namely vertex positions, velocities, and rigid poses. No constitutive models are used during training or inference. This purely data-driven approach makes PhysLDM agnostic to the underlying simulator's energy model, allowing it to seamlessly scale across diverse materials, and giving it the potential to learn from real-world captured trajectories where explicit energy models are unavailable (Appendix~\ref{app:agnostic}).

\section{Experiments}
\label{sec:exp}

\paragraph{Data and setup.}
We generate high-accuracy physical simulation data with a GPU FEM solver based on IPC \citep{DBLP:journals/tog/HuangCLK24,DBLP:journals/tog/HuangLLKL25}. For each scene, we randomly select $n$ objects ($n \le 3$) from more than $30{,}000$ Objaverse objects, tetrahedralize them with fTetWild \citep{DBLP:journals/tog/HuSWZP20} into volumetric meshes, and assign each object a random initial velocity ($v_0$) and a random stiffness (Young's modulus, $E$). The simulation time step is $1/60$\,s and each trajectory has $64$ frames. The Objaverse scenes ($63{,}397$ single-object and $16{,}078$ multi-body trajectories) are split into training, validation, and held-out test sets at a ratio of $90\%:5\%:5\%$. For OOD evaluation, we generate scenes in the same way from the objects of two other datasets, GSO and Toys4K. The ST-VAE and the LDM are trained on scenes with $N\in[500,5{,}000]$ vertices. The ST-VAE and LDM results in this paper are trained on single-object scenes ($n{=}1$). Our extended model is trained on multi-object scenes ($n\le3$; see the multi-body extension in Appendix~\ref{app:multibody}). In the controlled ST-VAE comparisons, we use ST-VAE Small ($K{=}512$, $T{=}32$, $M{=}4$, $95$M parameters). In the LDM experiments, we use ST-VAE Base ($K{=}512$, $T{=}64$, $M{=}8$, $124$M parameters; Table~\ref{tab:tiers}). Both configurations compress a trajectory by $8N/513$ in token count, which reaches $78\times$ at the $5{,}000$-vertex cap. Training details are in Appendix~\ref{app:arch}. Full data details, evaluation setup, and metric definitions are in Appendix~\ref{app:metrics}.

\paragraph{Staircase analysis.}
\label{sec:staircase}
Standard temporal compression methods (such as video VAEs) uniformly compress consecutive frames into fixed latent blocks along the time axis and decode them with a local temporal upsampler, but this approach is not suitable for 3D volumetric deformable dynamics trajectory reconstruction: it causes the object's deformation to hold and then snap to the next state, creating an unnatural ``staircase'' artifact that severely degrades visual fidelity. While this timing error is easily perceptible in video playback, it is difficult to identify from static multi-frame snapshots alone; therefore, we quantify it via the deformation-change correlation (DCC), defined in Appendix~\ref{app:metrics-t1}. Although DCC does not solely measure overall reconstruction quality on its own, it plays a decisive role in evaluating the physical plausibility of the motion when reconstruction precision is comparable. A detailed analysis of the staircase artifact is given in Appendix~\ref{app:staircase-loc}.

\paragraph{Chaos analysis.}
\label{sec:chaos}
Although physics simulators are commonly assumed to produce deterministic trajectories once initial states and parameters are fully specified, this is not guaranteed for GPU solvers. As shown in Fig.~\ref{fig:chaos}, running the same scene five times from byte-identical inputs yields visibly different final poses, with corresponding vertices landing up to $442$\,mm apart. This divergence stems from two factors: first, the underlying hardware non-determinism of GPU solvers introduces negligible initial variations; second, the inherently chaotic nature of complex deformable dynamics can rapidly amplify these microscopic errors into macroscopic trajectory divergence. Consequently, training data is not a deterministic function of initial conditions. We also verify this at scale on the $300$ Objaverse held-out scenes: running each scene twice from identical initial conditions, the final-frame vertex trajectories of $23.7\%$ of these meter-scale scenes differ by more than $1\%$ of the object size (about $1$\,cm), and those of $7.7\%$ differ by more than $5\%$ (about $5$\,cm) (Table~\ref{tab:resim}). This chaotic nature suggests why regression models struggle: faced with multiple branching futures, a squared-error loss would drive the model toward the conditional mean, collapsing incompatible outcomes into a damped and non-physical average. A generative approach like our LDM naturally avoids this by sampling plausible trajectories. A more detailed chaos analysis is given in Appendix~\ref{app:chaos}.

\begin{table}[!t]\centering\footnotesize
\setlength{\tabcolsep}{1.8pt}\renewcommand{\arraystretch}{1.1}\setlength{\aboverulesep}{0.15ex}\setlength{\belowrulesep}{0.25ex}
\caption{ST-VAE reconstruction comparison across the three evaluation sets, on $300$ scenes per set within the training range and $300$ scenes per set beyond the $5{,}000$-vertex training cap. DCC: deformation-change correlation; Recon: deformation reconstruction error (mm); rot / trans: rigid rotation (deg) and translation (mm) errors; full definitions in Appendix~\ref{app:metrics}. \tightbox{tableBest}{\textbf{best}} and \underline{second-best} per column within each block. The full comparison with all baselines is reported in Table~\ref{tab:s2full}. $^\dagger$No factoring of Eq.~\eqref{eq:factoring}; errors computed after removing the rigid motion from its output (Appendix~\ref{app:metrics}).}
\label{tab:s2main}
\begin{tabular*}{\linewidth}{@{\extracolsep{\fill}}c l cccccc ccc@{}}
\toprule
 & & \multicolumn{6}{c}{$500\le N\le 5{,}000$} & \multicolumn{3}{c}{\makebox[0pt][c]{$5{,}000<N\le10{,}000$}} \\
\cmidrule(lr){3-8}\cmidrule(lr){9-11}
 & & DCC$\uparrow$ & \multicolumn{3}{c}{Recon$\downarrow$} & \multicolumn{2}{c}{Rigid$\downarrow$} & Recon$\downarrow$ & \multicolumn{2}{c}{Rigid$\downarrow$} \\
\cmidrule(lr){3-3}\cmidrule(lr){4-6}\cmidrule(lr){7-8}\cmidrule(lr){9-9}\cmidrule(lr){10-11}
 & ST-VAE & mean & mean & med & P90 & rot & trans & mean & rot & trans \\
\midrule
\multirow{3}{*}{\rotatebox[origin=c]{90}{\scriptsize\shortstack{Objaverse\\Held-out}}} & Standard & 0.665 & \secondcell{5.32} & \secondcell{3.47} & \secondcell{12.01} & \bestcell{0.270} & \secondcell{0.57} & \secondcell{6.14} & \secondcell{0.185} & \secondcell{0.50} \\
 & Direct world-coord.$^\dagger$ & \secondcell{0.880} & 6.50 & 4.21 & 15.60 & 0.951 & 3.89 & 7.99 & 0.846 & 4.14 \\
 & \textbf{Holistic (ours)} & \bestcell{0.949} & \bestcell{2.48} & \bestcell{1.62} & \bestcell{8.08} & \secondcell{0.276} & \bestcell{0.53} & \bestcell{4.71} & \bestcell{0.169} & \bestcell{0.47} \\
\midrule
\multirow{3}{*}{\rotatebox[origin=c]{90}{GSO}} & Standard & 0.641 & \secondcell{3.46} & \secondcell{1.64} & \secondcell{8.83} & \secondcell{0.161} & \secondcell{0.48} & \secondcell{4.30} & \secondcell{0.149} & \secondcell{0.42} \\
 & Direct world-coord.$^\dagger$ & \secondcell{0.836} & 5.56 & 2.98 & 13.35 & 0.640 & 3.65 & 5.17 & 0.678 & 3.12 \\
 & \textbf{Holistic (ours)} & \bestcell{0.941} & \bestcell{1.91} & \bestcell{1.02} & \bestcell{5.88} & \bestcell{0.158} & \bestcell{0.45} & \bestcell{3.39} & \bestcell{0.147} & \bestcell{0.37} \\
\midrule
\multirow{3}{*}{\rotatebox[origin=c]{90}{Toys4K}} & Standard & 0.663 & \secondcell{5.49} & 3.83 & \secondcell{11.11} & \secondcell{0.310} & \secondcell{0.62} & \secondcell{6.86} & \secondcell{0.189} & \secondcell{0.51} \\
 & Direct world-coord.$^\dagger$ & \secondcell{0.858} & 6.96 & \secondcell{3.82} & 17.05 & 1.191 & 4.47 & 8.16 & 0.901 & 4.55 \\
 & \textbf{Holistic (ours)} & \bestcell{0.930} & \bestcell{3.12} & \bestcell{2.05} & \bestcell{8.29} & \bestcell{0.301} & \bestcell{0.61} & \bestcell{5.44} & \bestcell{0.174} & \bestcell{0.49} \\
\bottomrule
\end{tabular*}
\end{table}

\newlength{\qualh}\setlength{\qualh}{1.55cm}
\begin{figure*}[!t]
\centering
\setlength{\tabcolsep}{2pt}\renewcommand{\arraystretch}{0.6}
\resizebox{\textwidth}{!}{\begin{tabular}{@{}c@{\,}c@{\hspace{5pt}}c@{\hspace{5pt}}c@{\hspace{5pt}}c@{}}
\rotatebox{90}{\small GT} & \makebox[0pt][l]{\includegraphics[height=\qualh]{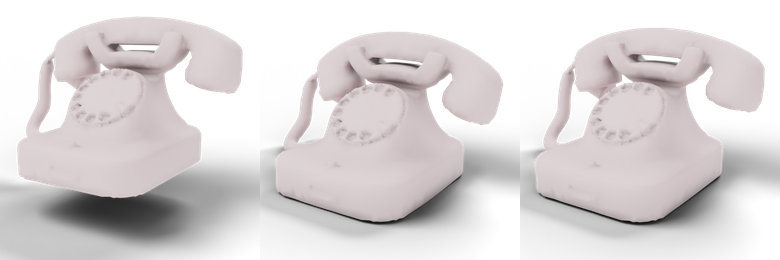}}\makebox[0pt][l]{\raisebox{1.5pt}{\hspace{1.5pt}\small $N{=}6106$}}\hspace{3.0000\qualh} & \makebox[0pt][l]{\includegraphics[height=\qualh]{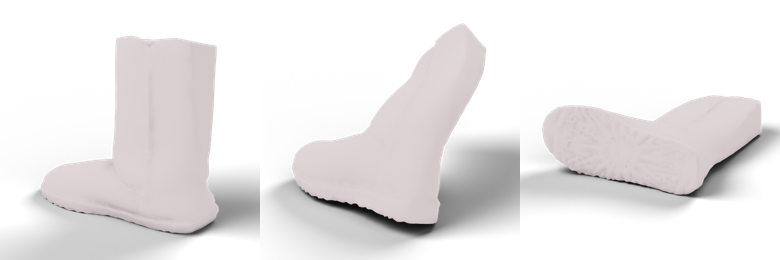}}\makebox[0pt][l]{\raisebox{1.5pt}{\hspace{1.5pt}\small $N{=}6569$}}\hspace{3.0000\qualh} & \makebox[0pt][l]{\includegraphics[height=\qualh]{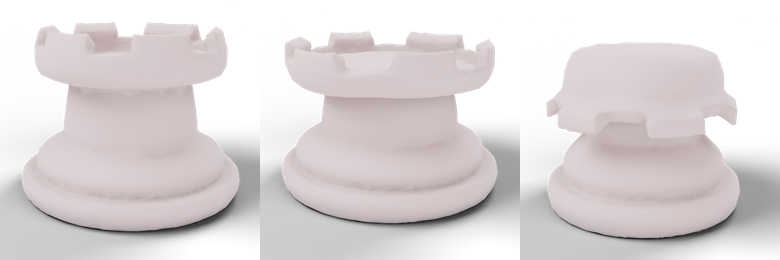}}\makebox[0pt][l]{\raisebox{1.5pt}{\hspace{1.5pt}\small $N{=}6322$}}\hspace{3.0000\qualh} & \makebox[0pt][l]{\includegraphics[height=\qualh]{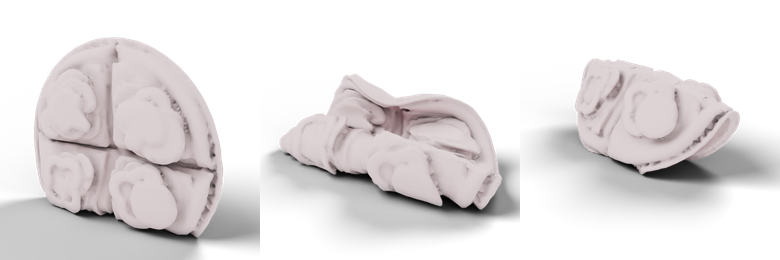}}\makebox[0pt][l]{\raisebox{1.5pt}{\hspace{1.5pt}\small $N{=}9403$}}\hspace{3.0000\qualh} \\
\rotatebox{90}{\small\shortstack{\textbf{Holistic}\\\textbf{(ours)}}} & \includegraphics[height=\qualh]{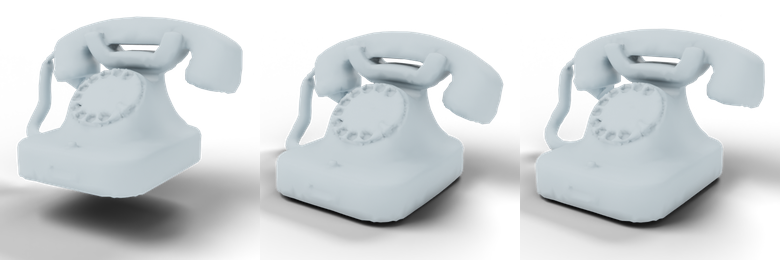} & \includegraphics[height=\qualh]{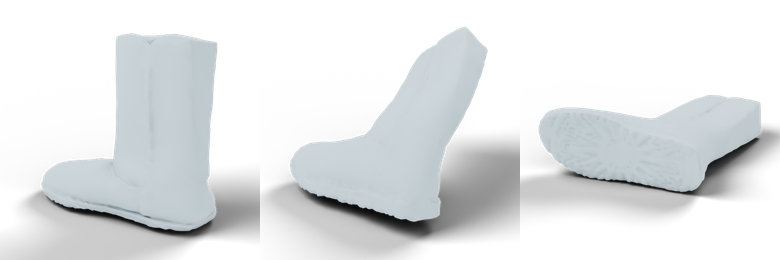} & \includegraphics[height=\qualh]{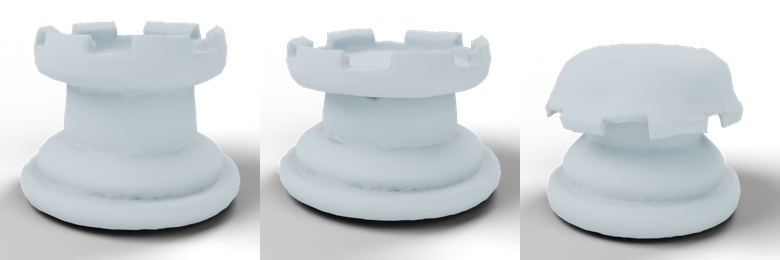} & \includegraphics[height=\qualh]{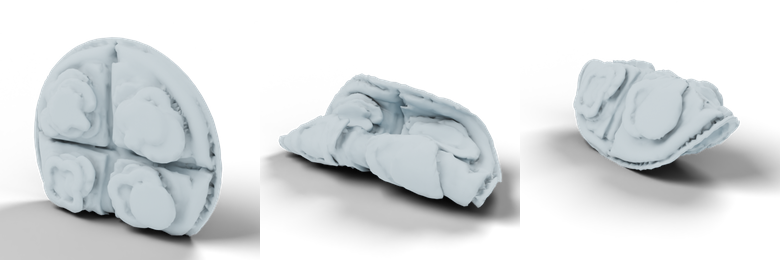} \\
\rotatebox{90}{\small Standard} & \includegraphics[height=\qualh]{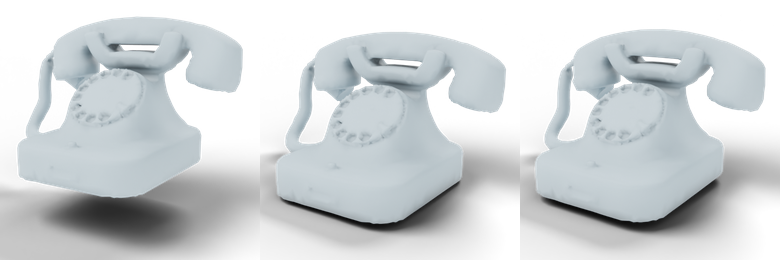} & \includegraphics[height=\qualh]{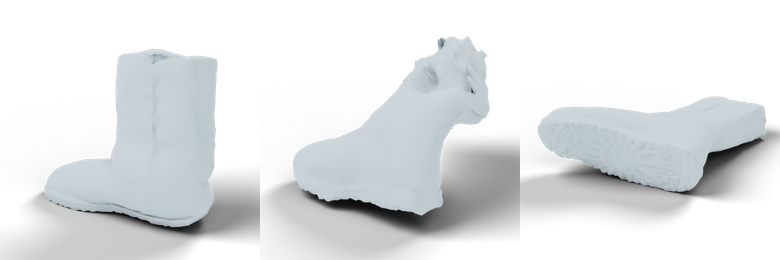} & \includegraphics[height=\qualh]{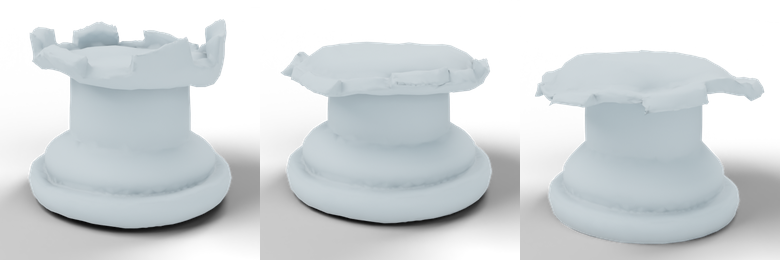} & \includegraphics[height=\qualh]{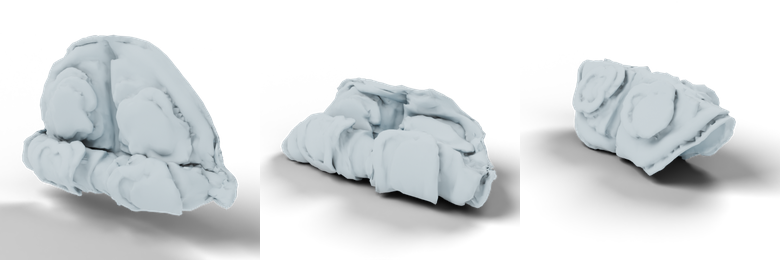} \\
\rotatebox{90}{\small\shortstack{Direct\\world-coord.}} & \makebox[0pt][l]{\includegraphics[height=\qualh]{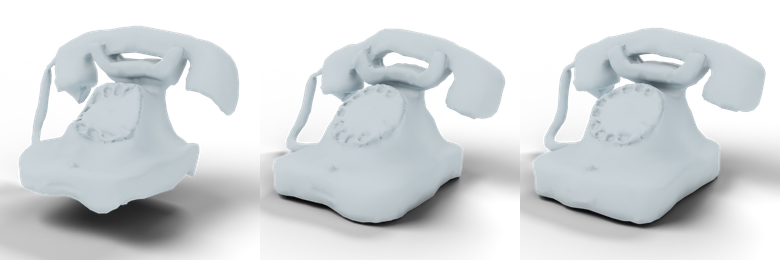}}\makebox[0pt][l]{\hspace{0.8pt}\tikz[baseline=0pt]{\node[inner sep=0pt,inner xsep=3pt,font=\normalsize\itshape,text=black!65] (t) at ($(1.5000\qualh,0)+(-1.5pt,0)$) {time};\draw[black!65,line width=1.6pt,line cap=round] (0,0) -- (t.west);\draw[black!65,line width=1.6pt,line cap=round,-{Straight Barb[length=11pt,width=13pt,angle'=60]}] (t.east) -- ($(3.0000\qualh,0)+(-3pt,0)$);}}\hspace{3.0000\qualh} & \includegraphics[height=\qualh]{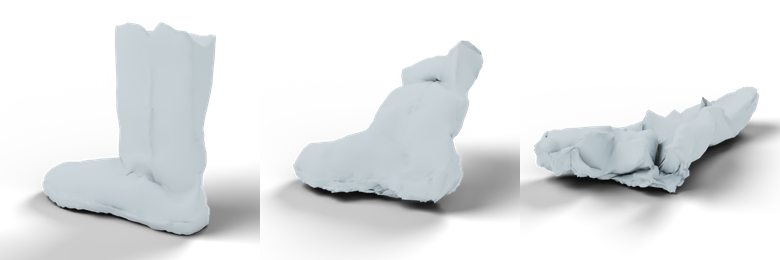} & \includegraphics[height=\qualh]{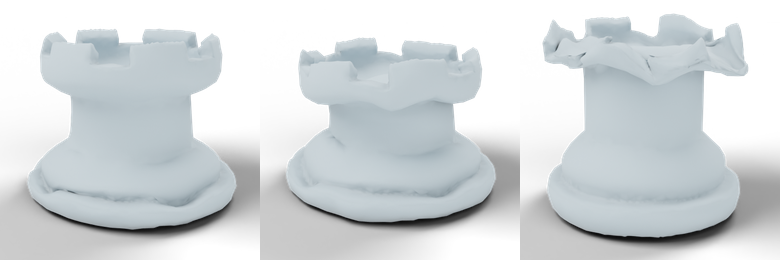} & \includegraphics[height=\qualh]{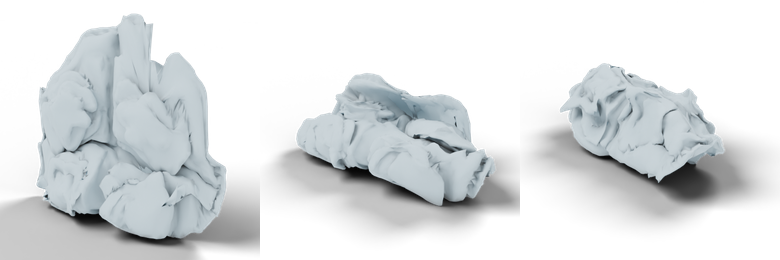} \\
\end{tabular}}

\caption{ST-VAE reconstruction on four scenes (a held-out Objaverse phone, a GSO boot, a Toys4K chess piece, and a GSO toy), showing three frames sampled along the time axis. Rows: ground truth (rose), ours, and two baselines. The standard baseline looks reasonable in these static frames, yet it suffers from the staircase artifact, which is hard to see in snapshots but is exposed by the DCC in Table~\ref{tab:s2main}. The direct world-coordinate baseline produces noisy geometric detail. Scene details are listed in Table~\ref{tab:qual_recon_cases}. More visualizations of ST-VAE reconstruction are shown in Figs.~\ref{fig:qual_recon_app} and~\ref{fig:qual_recon_app2}.}
\label{fig:qual_recon}
\end{figure*}

\paragraph{ST-VAE Reconstruction Results.}
Table~\ref{tab:s2main} and Fig.~\ref{fig:qual_recon} evaluate our holistic ST-VAE against two alternative designs: standard temporal compression (the design of video VAEs) and a direct world-coordinate baseline (which applies holistic compression without the factoring of Eq.~\eqref{eq:factoring}). All models share a matched compression rate and are assessed across the same three evaluation sets (Appendix~\ref{app:baselines}). The results yield three primary observations. First, holistic temporal compression eliminates staircase artifacts. Standard compression suffers from the staircase artifact (Sec.~\ref{sec:staircase}, Staircase analysis), which shows up as its clearly lower DCC. Our holistic ST-VAE preserves accurate motion timing, achieving a DCC exceeding $0.93$ across all evaluation sets. Furthermore, augmenting the standard baseline with auxiliary velocity or spectral losses, or placing the latent frames adaptively in time, does not mitigate this limitation (Table~\ref{tab:s2full}). Second, decoupling rigid motion from deformation enhances precision. The direct world-coordinate baseline underperforms the holistic ST-VAE across all deformation and rigid motion metrics, and shows noisy and cluttered geometric detail in the visual comparison (Fig.~\ref{fig:qual_recon}). Third, the holistic ST-VAE achieves superior precision across diverse scenarios. Under up to $78\times$ compression, the ST-VAE yields a mean geometric reconstruction error of just $2.48$\,mm on held-out scenes, representing about half the error of the standard baseline, with a median vertex error below $2$\,mm. It achieves the best performance across nearly all metrics and datasets, except for one rigid-rotation entry where it is a close second. This performance holds robustly zero-shot on OOD sets like GSO and Toys4K, where its errors stay at the same level as in distribution, and on high-resolution meshes exceeding the $5{,}000$-vertex training cap, where its mean error is still only a few millimeters. The least accurate held-out cases, geometries with many long, thin features, are analyzed in Appendix~\ref{app:extreme}.

\begin{figure*}[!t]
\centering
\setlength{\tabcolsep}{2pt}\renewcommand{\arraystretch}{0.6}
\resizebox{\textwidth}{!}{\begin{tabular}{@{}c@{\,}c@{\hspace{5pt}}c@{\hspace{5pt}}c@{\hspace{5pt}}c@{}}
\rotatebox{90}{\small GT} & \makebox[0pt][l]{\includegraphics[height=\qualh]{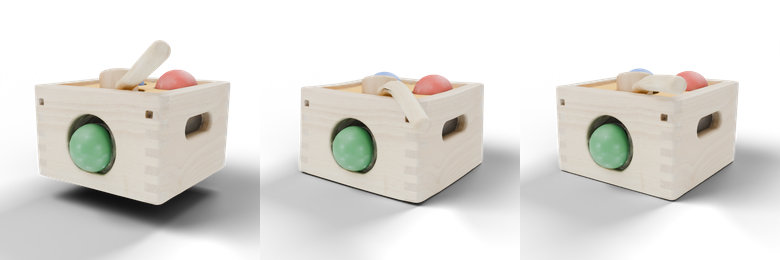}}\makebox[0pt][l]{\raisebox{1.5pt}{\hspace{1.5pt}\small $N{=}7767$}}\hspace{3.0000\qualh} & \makebox[0pt][l]{\includegraphics[height=\qualh]{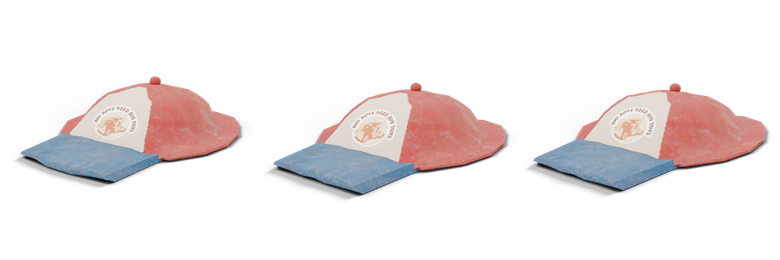}}\makebox[0pt][l]{\raisebox{1.5pt}{\hspace{1.5pt}\small $N{=}2435$}}\hspace{3.0000\qualh} & \makebox[0pt][l]{\includegraphics[height=\qualh]{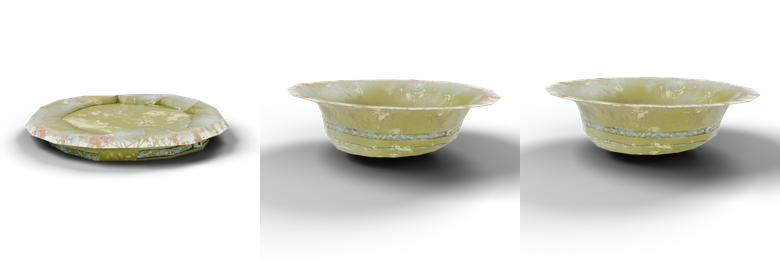}}\makebox[0pt][l]{\raisebox{1.5pt}{\hspace{1.5pt}\small $N{=}5877$}}\hspace{3.0000\qualh} & \makebox[0pt][l]{\includegraphics[height=\qualh]{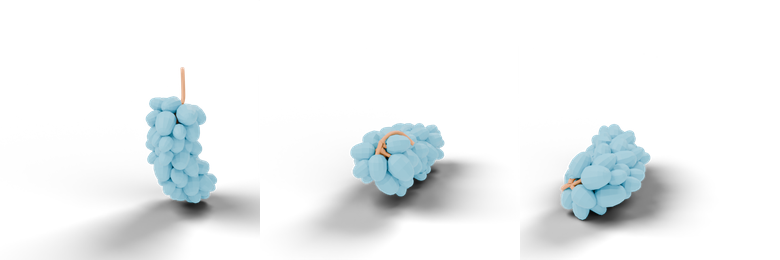}}\makebox[0pt][l]{\raisebox{1.5pt}{\hspace{1.5pt}\small $N{=}8767$}}\hspace{3.0000\qualh} \\
\rotatebox{90}{\small\shortstack{\textbf{Diffusion}\\\textbf{(ours)}}} & \includegraphics[height=\qualh]{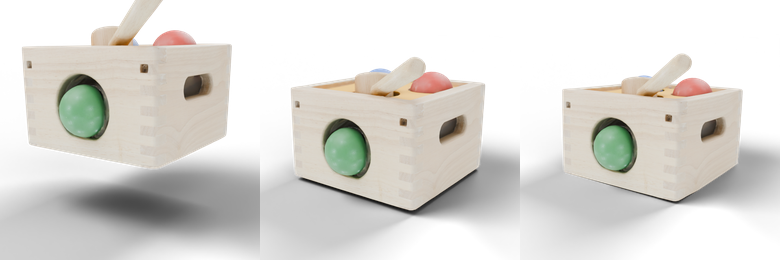} & \includegraphics[height=\qualh]{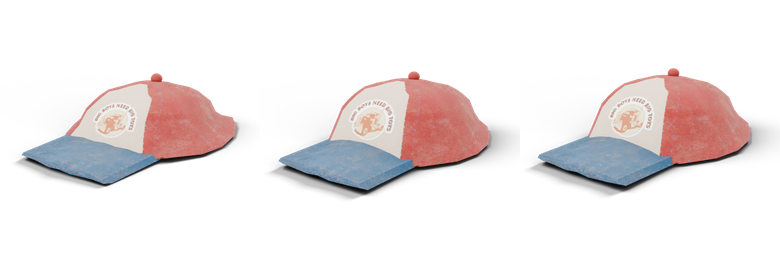} & \includegraphics[height=\qualh]{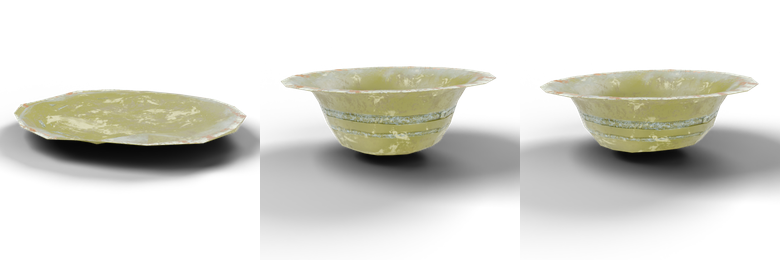} & \includegraphics[height=\qualh]{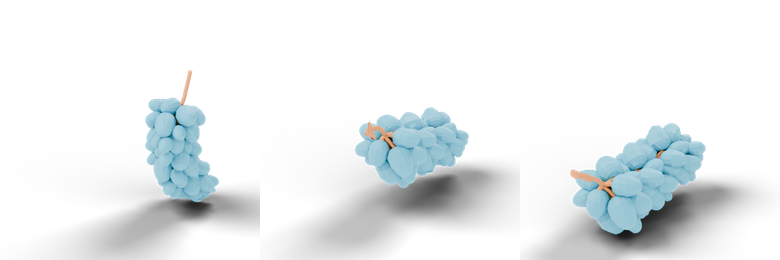} \\
\rotatebox{90}{\small Regression} & \includegraphics[height=\qualh]{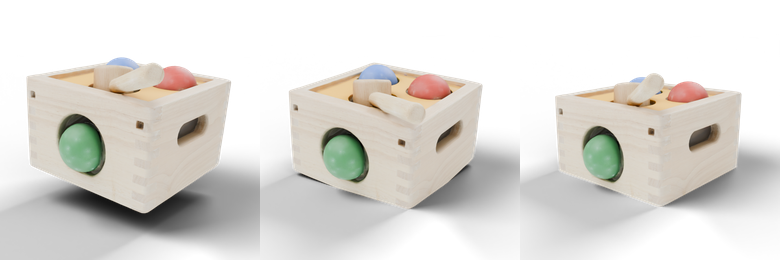} & \includegraphics[height=\qualh]{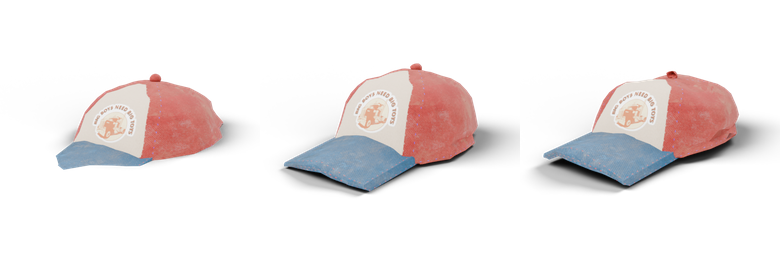} & \includegraphics[height=\qualh]{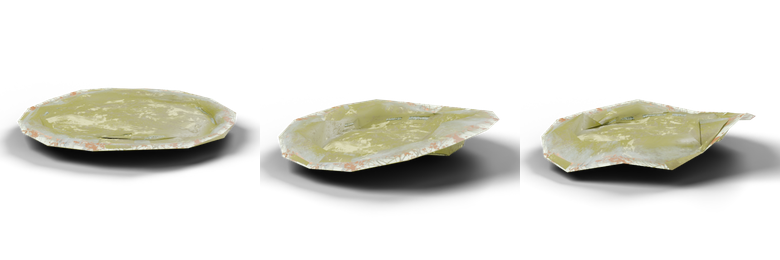} & \includegraphics[height=\qualh]{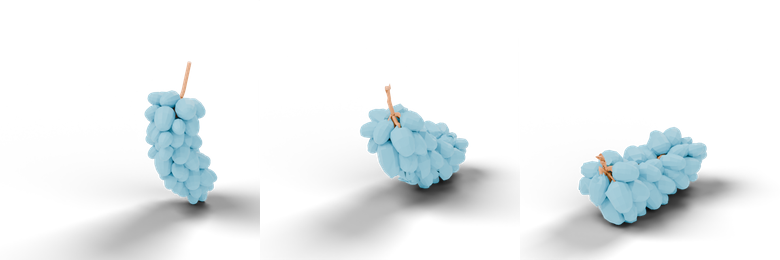} \\
\rotatebox{90}{\small\shortstack{MGN-style\\AR}} & \makebox[0pt][l]{\includegraphics[height=\qualh]{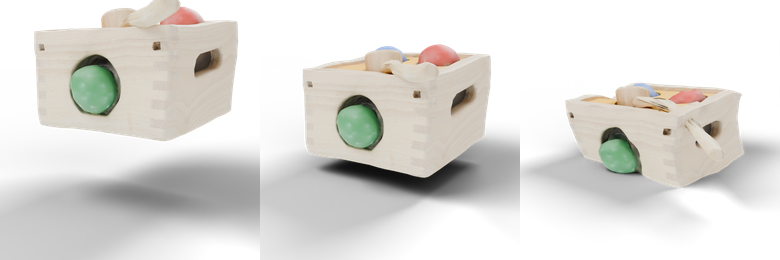}}\makebox[0pt][l]{\hspace{0.8pt}\tikz[baseline=0pt]{\node[inner sep=0pt,inner xsep=3pt,font=\normalsize\itshape,text=black!65] (t) at ($(1.5000\qualh,0)+(-1.5pt,0)$) {time};\draw[black!65,line width=1.6pt,line cap=round] (0,0) -- (t.west);\draw[black!65,line width=1.6pt,line cap=round,-{Straight Barb[length=11pt,width=13pt,angle'=60]}] (t.east) -- ($(3.0000\qualh,0)+(-3pt,0)$);}}\hspace{3.0000\qualh} & \includegraphics[height=\qualh]{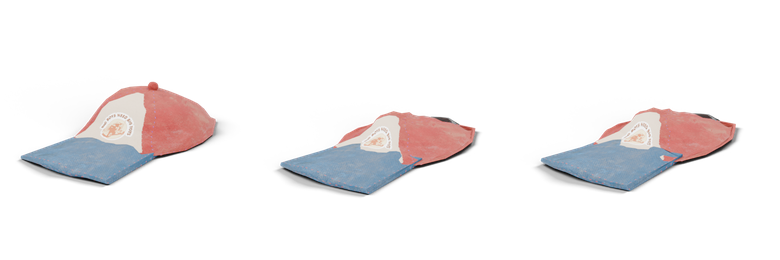} & \includegraphics[height=\qualh]{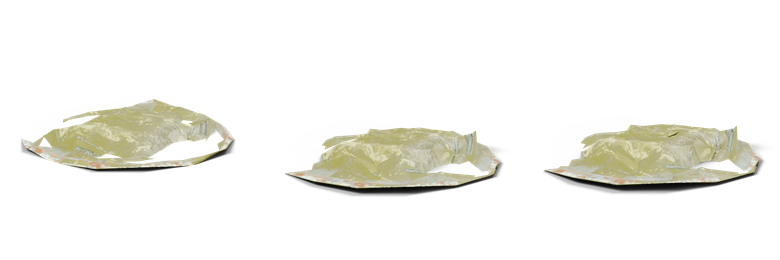} & \includegraphics[height=\qualh]{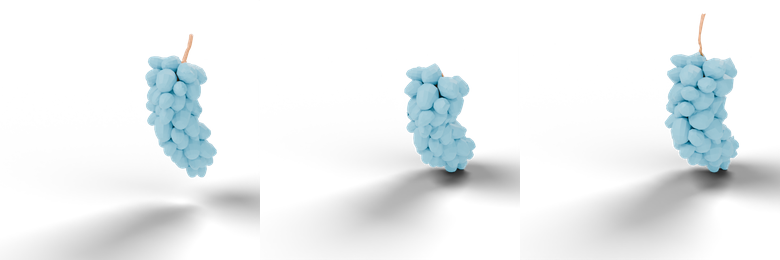} \\
\end{tabular}}

\caption{Prediction on four scenes (a GSO punch-drop toy, a held-out Objaverse cap and bowl, and Toys4K grapes), showing three frames sampled along the time axis. Rows: ground truth, \textbf{Diffusion (ours)}, Regression, and the MGN-style AR baseline. The regression method predicts geometry with visible artifacts and jitter. MGN-style AR drifts over time. Scene details are listed in Table~\ref{tab:qual_pred_cases}. More visualizations of prediction are shown in Fig.~\ref{fig:qual_pred_app}.}
\label{fig:qual_pred}
\end{figure*}

\paragraph{Prediction Results.}
Table~\ref{tab:s3phys} and Fig.~\ref{fig:qual_pred} compare three models predicting a $64$-frame trajectory from the same initial state: our LDM, a capacity-matched deterministic regression model (both using one-shot prediction and the same ST-VAE), and a frame-by-frame AR graph network based on MeshGraphNets (MGN) (Appendix~\ref{app:predbaselines}). Because complex deformable dynamics are inherently chaotic, world-space error alone is insufficient to evaluate prediction quality, as the ground truth represents only one of many valid outcomes. Therefore, Table~\ref{tab:s3phys} primarily relies on physics-consistency checks (Appendix~\ref{app:metrics}; the full set of checks is in Table~\ref{tab:s3full}). Across all metrics and datasets, the LDM significantly outperforms both baselines. We highlight three observations from the results. First, the LDM consistently outperforms deterministic regression in generating physically plausible trajectories. The regression model produces damped outputs with geometric artifacts and jitter (e.g., $22.7\%$ volume violations versus $7.9\%$ for the LDM on held-out scenes). In contrast, the LDM reduces these artifacts and better preserves the deformation-to-rigid energy ratio. Second, one-shot prediction provides the global temporal context necessary to prevent the compounding errors typically seen in autoregressive models. Consequently, the AR baseline drifts over time and accumulates nearly three times the world-space error of the LDM ($80.6$ against $28.2$\,mm). Third, this strong performance transfers zero-shot to OOD datasets (GSO and Toys4K), demonstrating the robust generalization of our approach. Finally, the LDM remains highly efficient, running at $40$ to $80$ frames per second on a single GPU, even for high-resolution meshes exceeding the training cap (Appendix~\ref{app:runtime}).

\begin{table}[!t]\centering\footnotesize
\setlength{\tabcolsep}{1.2pt}\renewcommand{\arraystretch}{0.95}\setlength{\aboverulesep}{0.15ex}\setlength{\belowrulesep}{0.25ex}
\caption{Prediction comparison across the three evaluation sets, on the same $300$ scenes per set within the training vertex range ($500\le N\le5{,}000$) as Table~\ref{tab:s2main}. Vol.: tetrahedral volume violation \%; SI: self-intersecting surface triangles \%; KE: deformation-to-rigid kinetic-energy ratio $KE_{\mathrm{def}}/KE_{\mathrm{rig}}$ relative to ground truth (closer to $1$ is better); Err: median per-vertex error in world coordinates (mm); full definitions in Appendix~\ref{app:metrics}. The full set of physics checks is reported in Table~\ref{tab:s3full}. \tightbox{tableBest}{\textbf{best}} and \underline{second-best} per column.}
\label{tab:s3phys}
\begin{tabular*}{\linewidth}{@{\extracolsep{\fill}}l cccc cccc cccc@{}}
\toprule
 & \multicolumn{4}{c}{Objaverse Held-out} & \multicolumn{4}{c}{GSO} & \multicolumn{4}{c}{Toys4K} \\
\cmidrule(lr){2-5}\cmidrule(lr){6-9}\cmidrule(lr){10-13}
Prediction Model & Vol.$\downarrow$ & SI$\downarrow$ & KE$\rightarrow$1 & Err$\downarrow$ & Vol.$\downarrow$ & SI$\downarrow$ & KE$\rightarrow$1 & Err$\downarrow$ & Vol.$\downarrow$ & SI$\downarrow$ & KE$\rightarrow$1 & Err$\downarrow$ \\
\midrule
\textbf{Diffusion (ours)} & \bestcell{7.9} & \bestcell{0.24} & \bestcell{0.80} & \bestcell{28.2} & \bestcell{2.0} & \bestcell{0.00} & \bestcell{0.86} & \bestcell{14.0} & \bestcell{10.3} & \bestcell{0.32} & \bestcell{0.80} & \bestcell{35.6} \\
Regression & \secondcell{22.7} & \secondcell{0.79} & 0.72 & \secondcell{37.5} & \secondcell{5.1} & \secondcell{0.01} & 0.72 & \secondcell{19.9} & \secondcell{24.3} & \secondcell{1.07} & 0.69 & \secondcell{42.6} \\
MGN-style AR & 55.1 & 3.53 & \secondcell{0.77} & 80.6 & 47.1 & 0.68 & \secondcell{0.76} & 78.4 & 57.5 & 4.73 & \secondcell{0.77} & 79.1 \\
\bottomrule
\end{tabular*}
\end{table}

The diffusion results in Table~\ref{tab:s3phys} use a single sample per scene, which cannot show whether diffusion captures the several possible outcomes of a chaotic scene. To test whether diffusion benefits from modeling several outcomes, we compare the two models separately on chaotic and non-chaotic held-out scenes and draw eight diffusion samples per scene (Appendix~\ref{app:chaos71}). Diffusion outperforms regression on both subsets. One may worry that this only reflects a stronger model. However, the diffusion samples spread much more widely on the chaotic scenes, and this spread follows the difference between two runs of the solver. Since the model is the same on both subsets, this effect comes from the generative paradigm and not from a stronger model. A soft key dropped onto the ground illustrates it (Fig.~\ref{fig:keys}). Run twice, the solver lets the key fall over in two different directions. Every diffusion sample faithfully falls toward one of the two correct directions, while regression predicts a non-physical average that matches neither run.

\subsection{Ablations}
\label{sec:ablations}

\paragraph{ST-VAE.}
We remove the ST-VAE and let a DiT of the same capacity as our LDM denoise the vertex$\times$frame tokens directly (Table~\ref{tab:latent}). Without the latent, one-shot diffusion at native resolution is computationally prohibitive. Memory per sample grows with the vertex count. Even on a $140$\,GiB H200, the model at batch~$1$ runs out of memory at $2{,}600$ vertices. With the latent, the LDM works on a fixed number of tokens and its cost is independent of the vertex count. The vertex-dependent cost moves to the ST-VAE stage, whose training memory grows linearly with $N$ (Table~\ref{tab:latent}), so on the same GPU the training cap rises from about $2{,}500$ vertices without the latent to about $14{,}000$ for ST-VAE Base. The latent also improves the quality of the generated geometry. A raw-coordinate model predicts every vertex directly, so any noise left by the sampler shows up as per-vertex jitter. Our model decodes all vertices from the latent, so the same noise only shifts the shape slightly and coherently (illustrated in Appendix~\ref{app:physiformer}).

\paragraph{The dedicated rigid pathway in ST-VAE.}
In the ST-VAE (Sec.~\ref{sec:arch}), the rigid token bypasses the holistic temporal compression and takes its own linear temporal pathway. To test this design, we route the rigid token through the shared holistic temporal compression instead. In our experiments, this one change causes two failures: the translation error explodes, and the reconstructed deformation shrinks drastically (detailed diagnostics in Appendix~\ref{app:moreabl}). Further ablations (spatial-VAE pretraining and the decoder GNN) are reported in Appendix~\ref{app:moreabl}.

\subsection{Differentiable Applications}
\label{sec:diffapp}

We demonstrate the end-to-end differentiability of PhysLDM on two tasks of increasing difficulty and compare against two differentiable simulators, DiffIPC \citep{DBLP:journals/tog/HuangTGFSPZ24} and Newton/Warp \citep{newton2025}. The detailed comparison, together with the setup and results, is given in Appendix~\ref{app:diffapp}.

\paragraph{Material Estimation.}
The bottom-right block of Fig.~\ref{fig:teaser} shows a material estimation task. Given a single simulated trajectory of a body with unknown stiffness, we recover its Young's modulus $E$ by gradient descent through the model, since the mismatch to the observed trajectory is differentiable in $E$ end to end. Because Young's modulus is usually measured on a logarithmic scale, we optimize $E$ in $\log_{10}$ space, where an error of $1$\,dex means a factor of $10$. To cover a wide range of stiffness, we repeat the task on seven bodies with moduli from $3\times10^{4}$ to $3\times10^{6}$\,Pa, each starting from an initial guess up to $1.1$\,dex away. In all seven cases, our model recovers $E$ within $0.06$\,dex in $40$ Adam steps, which takes about $40$ seconds per run. DiffIPC solves the task only after its solver parameters are tuned by hand, and then needs $26$ minutes on $48$ threads. Newton/Warp fails in the backward pass, because backpropagating through its enormous number of substeps explodes the gradient.

\paragraph{Beyond Material Estimation: Designing the Throw.}
Based on the first task, we may ask a further question: which throw reveals the stiffness best, so that the estimation above becomes easier? Easier means that a change of $E$ produces a larger change of the trajectory, so the trajectory carries more information about the stiffness. We therefore maximize the sensitivity of the predicted trajectory to $E$ over the release velocity by gradient ascent through the model. The objective is a derivative, so its gradient is a second-order quantity. Our model supports higher-order derivatives through the automatic differentiation of PyTorch, so it provides this gradient directly, and $30$ Adam steps finish the design in $7$ minutes. Replayed in the ground-truth simulator, the optimized throw makes the rigid motion $13\times$ more sensitive to the stiffness. DiffIPC offers only first-order derivatives and cannot be applied directly to this second-order task. A workaround is to obtain the second-order gradient by nested finite differences, but then DiffIPC is about $635\times$ slower than our model per gradient step. The whole design would take several days, which is not acceptable in practice. Newton/Warp does not produce a usable first-order gradient on this scene, so a second-order one is out of reach.

\section{Conclusion}
We introduce PhysLDM, a latent diffusion paradigm for predicting the dynamics of high-fidelity volumetric deformable bodies. To overcome the computational limits of processing high-resolution meshes, we developed a novel holistic ST-VAE that explicitly disentangles global rigid motion from local deformations. This architecture eliminates the staircase artifacts inherent to standard temporal compression, ensuring highly accurate trajectory reconstruction even for high-resolution, out-of-distribution meshes. Operating on this compact latent representation, we demonstrate that while deterministic regression produces damped, non-physical averages in this chaotic regime, our generative approach predicts physically consistent trajectories in a single diffusion pass. Finally, trained purely kinematically, PhysLDM delivers robust zero-shot generalization and provides end-to-end differentiability, unlocking complex inverse design optimizations that remain intractable for traditional solvers.

\textbf{Limitations and Future Work.} Currently, our model focuses on predicting $64$-frame trajectories in a single diffusion pass. Extending this one-shot approach into an autoregressive framework presents a promising direction for future research.

\bibliographystyle{iclr2027_conference}
\bibliography{refs}

@inproceedings{DBLP:conf/icml/Sanchez-Gonzalez20,
  author       = {Alvaro Sanchez{-}Gonzalez and
                  Jonathan Godwin and
                  Tobias Pfaff and
                  Rex Ying and
                  Jure Leskovec and
                  Peter W. Battaglia},
  title        = {Learning to Simulate Complex Physics with Graph Networks},
  booktitle    = {Proceedings of the 37th International Conference on Machine Learning,
                  {ICML} 2020, 13-18 July 2020, Virtual Event},
  series       = {Proceedings of Machine Learning Research},
  volume       = {119},
  pages        = {8459--8468},
  publisher    = {{PMLR}},
  year         = {2020},
  url          = {http://proceedings.mlr.press/v119/sanchez-gonzalez20a.html},
  bibsource    = {dblp computer science bibliography, https://dblp.org}
}

@inproceedings{DBLP:conf/iclr/PfaffFSB21,
  author       = {Tobias Pfaff and
                  Meire Fortunato and
                  Alvaro Sanchez{-}Gonzalez and
                  Peter W. Battaglia},
  title        = {Learning Mesh-Based Simulation with Graph Networks},
  booktitle    = {9th International Conference on Learning Representations, {ICLR} 2021,
                  Virtual Event, Austria, May 3-7, 2021},
  publisher    = {OpenReview.net},
  year         = {2021},
  url          = {https://openreview.net/forum?id=roNqYL0\_XP},
  bibsource    = {dblp computer science bibliography, https://dblp.org}
}

@inproceedings{zhang2026clothtransformer,
  title        = {ClothTransformer: Unified Latent-Space Transformers for Scalable Cloth Simulation},
  author       = {Yu Zhang and Yidi Shao and Wenqi Ouyang and Yushi Lan and Zhexin Liang and Chengrui Wu and Xudong Xu and Xingang Pan},
  booktitle    = {Advances in Neural Information Processing Systems (NeurIPS)},
  year         = {2026},
  url          = {https://arxiv.org/abs/2605.27852},
}

@inproceedings{DBLP:conf/cvpr/YuCSL0CH0HE023,
  author       = {Lijun Yu and
                  Yong Cheng and
                  Kihyuk Sohn and
                  Jos{\'{e}} Lezama and
                  Han Zhang and
                  Huiwen Chang and
                  Alexander G. Hauptmann and
                  Ming{-}Hsuan Yang and
                  Yuan Hao and
                  Irfan Essa and
                  Lu Jiang},
  title        = {{MAGVIT:} Masked Generative Video Transformer},
  booktitle    = {{IEEE/CVF} Conference on Computer Vision and Pattern Recognition,
                  {CVPR} 2023, Vancouver, BC, Canada, June 17-24, 2023},
  pages        = {10459--10469},
  publisher    = {{IEEE}},
  year         = {2023},
  url          = {https://doi.org/10.1109/CVPR52729.2023.01008},
  doi          = {10.1109/CVPR52729.2023.01008},
  bibsource    = {dblp computer science bibliography, https://dblp.org}
}

@inproceedings{DBLP:conf/iclr/YangTZ00XYHZFYZ25,
  author       = {Zhuoyi Yang and
                  Jiayan Teng and
                  Wendi Zheng and
                  Ming Ding and
                  Shiyu Huang and
                  Jiazheng Xu and
                  Yuanming Yang and
                  Wenyi Hong and
                  Xiaohan Zhang and
                  Guanyu Feng and
                  Da Yin and
                  Yuxuan Zhang and
                  Weihan Wang and
                  Yean Cheng and
                  Bin Xu and
                  Xiaotao Gu and
                  Yuxiao Dong and
                  Jie Tang},
  title        = {CogVideoX: Text-to-Video Diffusion Models with An Expert Transformer},
  booktitle    = {The Thirteenth International Conference on Learning Representations,
                  {ICLR} 2025, Singapore, April 24-28, 2025},
  publisher    = {OpenReview.net},
  year         = {2025},
  url          = {https://openreview.net/forum?id=LQzN6TRFg9},
  bibsource    = {dblp computer science bibliography, https://dblp.org}
}

@article{DBLP:journals/tog/HuangCLK24,
  author       = {Kemeng Huang and
                  Floyd M. Chitalu and
                  Huancheng Lin and
                  Taku Komura},
  title        = {{GIPC:} Fast and Stable Gauss-Newton Optimization of {IPC} Barrier
                  Energy},
  journal      = {{ACM} Trans. Graph.},
  volume       = {43},
  number       = {2},
  pages        = {23:1--23:18},
  year         = {2024},
  url          = {https://doi.org/10.1145/3643028},
  doi          = {10.1145/3643028},
  bibsource    = {dblp computer science bibliography, https://dblp.org}
}

@article{DBLP:journals/tog/HuangLLKL25,
  author       = {Kemeng Huang and
                  Xinyu Lu and
                  Huancheng Lin and
                  Taku Komura and
                  Minchen Li},
  title        = {StiffGIPC: Advancing {GPU} {IPC} for Stiff Affine-Deformable Simulation},
  journal      = {{ACM} Trans. Graph.},
  volume       = {44},
  number       = {3},
  pages        = {31:1--31:20},
  year         = {2025},
  url          = {https://doi.org/10.1145/3735126},
  doi          = {10.1145/3735126},
  bibsource    = {dblp computer science bibliography, https://dblp.org}
}

@inproceedings{DBLP:conf/iccv/PetrovichBV21,
  author       = {Mathis Petrovich and
                  Michael J. Black and
                  G{\"{u}}l Varol},
  title        = {Action-Conditioned 3D Human Motion Synthesis with Transformer {VAE}},
  booktitle    = {2021 {IEEE/CVF} International Conference on Computer Vision, {ICCV}
                  2021, Montreal, QC, Canada, October 10-17, 2021},
  pages        = {10965--10975},
  publisher    = {{IEEE}},
  year         = {2021},
  url          = {https://doi.org/10.1109/ICCV48922.2021.01080},
  doi          = {10.1109/ICCV48922.2021.01080},
  bibsource    = {dblp computer science bibliography, https://dblp.org}
}

@inproceedings{DBLP:conf/cvpr/RombachBLEO22,
  author       = {Robin Rombach and
                  Andreas Blattmann and
                  Dominik Lorenz and
                  Patrick Esser and
                  Bj{\"{o}}rn Ommer},
  title        = {High-Resolution Image Synthesis with Latent Diffusion Models},
  booktitle    = {{IEEE/CVF} Conference on Computer Vision and Pattern Recognition,
                  {CVPR} 2022, New Orleans, LA, USA, June 18-24, 2022},
  pages        = {10674--10685},
  publisher    = {{IEEE}},
  year         = {2022},
  url          = {https://doi.org/10.1109/CVPR52688.2022.01042},
  doi          = {10.1109/CVPR52688.2022.01042},
  bibsource    = {dblp computer science bibliography, https://dblp.org}
}

@inproceedings{DBLP:conf/cvpr/DeitkeSSWMVSEKF23,
  author       = {Matt Deitke and
                  Dustin Schwenk and
                  Jordi Salvador and
                  Luca Weihs and
                  Oscar Michel and
                  Eli VanderBilt and
                  Ludwig Schmidt and
                  Kiana Ehsani and
                  Aniruddha Kembhavi and
                  Ali Farhadi},
  title        = {Objaverse: {A} Universe of Annotated 3D Objects},
  booktitle    = {{IEEE/CVF} Conference on Computer Vision and Pattern Recognition,
                  {CVPR} 2023, Vancouver, BC, Canada, June 17-24, 2023},
  pages        = {13142--13153},
  publisher    = {{IEEE}},
  year         = {2023},
  url          = {https://doi.org/10.1109/CVPR52729.2023.01263},
  doi          = {10.1109/CVPR52729.2023.01263},
  bibsource    = {dblp computer science bibliography, https://dblp.org}
}

@inproceedings{DBLP:conf/icra/DownsFKKHRMV22,
  author       = {Laura Downs and
                  Anthony Francis and
                  Nate Koenig and
                  Brandon Kinman and
                  Ryan Hickman and
                  Krista Reymann and
                  Thomas Barlow McHugh and
                  Vincent Vanhoucke},
  title        = {Google Scanned Objects: {A} High-Quality Dataset of 3D Scanned Household
                  Items},
  booktitle    = {2022 International Conference on Robotics and Automation, {ICRA} 2022,
                  Philadelphia, PA, USA, May 23-27, 2022},
  pages        = {2553--2560},
  publisher    = {{IEEE}},
  year         = {2022},
  url          = {https://doi.org/10.1109/ICRA46639.2022.9811809},
  doi          = {10.1109/ICRA46639.2022.9811809},
  bibsource    = {dblp computer science bibliography, https://dblp.org}
}

@inproceedings{DBLP:conf/cvpr/StojanovTR21,
  author       = {Stefan Stojanov and
                  Anh Thai and
                  James M. Rehg},
  title        = {Using Shape To Categorize: Low-Shot Learning With an Explicit Shape
                  Bias},
  booktitle    = {{IEEE} Conference on Computer Vision and Pattern Recognition, {CVPR}
                  2021, virtual, June 19-25, 2021},
  pages        = {1798--1808},
  publisher    = {Computer Vision Foundation / {IEEE}},
  year         = {2021},
  url          = {https://openaccess.thecvf.com/content/CVPR2021/html/Stojanov\_Using\_Shape\_To\_Categorize\_Low-Shot\_Learning\_With\_an\_Explicit\_Shape\_CVPR\_2021\_paper.html},
  doi          = {10.1109/CVPR46437.2021.00184},
  bibsource    = {dblp computer science bibliography, https://dblp.org}
}

@article{DBLP:journals/tog/LiFSLZPJK20,
  author       = {Minchen Li and
                  Zachary Ferguson and
                  Teseo Schneider and
                  Timothy R. Langlois and
                  Denis Zorin and
                  Daniele Panozzo and
                  Chenfanfu Jiang and
                  Danny M. Kaufman},
  title        = {Incremental potential contact: intersection-and inversion-free, large-deformation
                  dynamics},
  journal      = {{ACM} Trans. Graph.},
  volume       = {39},
  number       = {4},
  pages        = {49},
  year         = {2020},
  url          = {https://doi.org/10.1145/3386569.3392425},
  doi          = {10.1145/3386569.3392425},
  bibsource    = {dblp computer science bibliography, https://dblp.org}
}

@article{DBLP:journals/tog/TangHBLL26,
  author       = {Xuan Tang and
                  Kemeng Huang and
                  Gilbert Bernstein and
                  Minchen Li and
                  Tzumao Li},
  title        = {{YASPS:} {A} Symbolic Framework for Extensible, High-Performance {IPC}
                  Simulation},
  journal      = {{ACM} Trans. Graph.},
  volume       = {45},
  number       = {4},
  pages        = {142:1--142:35},
  year         = {2026},
  url          = {https://doi.org/10.1145/3811327},
  doi          = {10.1145/3811327},
  bibsource    = {dblp computer science bibliography, https://dblp.org}
}

@article{DBLP:journals/tog/Ando24,
  author       = {Ryoichi Ando},
  title        = {A Cubic Barrier with Elasticity-Inclusive Dynamic Stiffness},
  journal      = {{ACM} Trans. Graph.},
  volume       = {43},
  number       = {6},
  pages        = {224:1--224:13},
  year         = {2024},
  url          = {https://doi.org/10.1145/3687908},
  doi          = {10.1145/3687908},
  bibsource    = {dblp computer science bibliography, https://dblp.org}
}

@article{DBLP:journals/tog/ChenLYY24,
  author       = {Anka He Chen and
                  Ziheng Liu and
                  Yin Yang and
                  Cem Yuksel},
  title        = {Vertex Block Descent},
  journal      = {{ACM} Trans. Graph.},
  volume       = {43},
  number       = {4},
  pages        = {116:1--116:16},
  year         = {2024},
  url          = {https://doi.org/10.1145/3658179},
  doi          = {10.1145/3658179},
  bibsource    = {dblp computer science bibliography, https://dblp.org}
}

@article{DBLP:journals/tog/GilesDY25,
  author       = {Chris Giles and
                  Elie Diaz and
                  Cem Yuksel},
  title        = {Augmented Vertex Block Descent},
  journal      = {{ACM} Trans. Graph.},
  volume       = {44},
  number       = {4},
  pages        = {90:1--90:12},
  year         = {2025},
  url          = {https://doi.org/10.1145/3731195},
  doi          = {10.1145/3731195},
  bibsource    = {dblp computer science bibliography, https://dblp.org}
}

@inproceedings{DBLP:conf/iclr/UmmenhoferPTK20,
  author       = {Benjamin Ummenhofer and
                  Lukas Prantl and
                  Nils Thuerey and
                  Vladlen Koltun},
  title        = {Lagrangian Fluid Simulation with Continuous Convolutions},
  booktitle    = {8th International Conference on Learning Representations, {ICLR} 2020,
                  Addis Ababa, Ethiopia, April 26-30, 2020},
  publisher    = {OpenReview.net},
  year         = {2020},
  url          = {https://openreview.net/forum?id=B1lDoJSYDH},
  bibsource    = {dblp computer science bibliography, https://dblp.org}
}

@article{DBLP:journals/corr/abs-2606-27364,
  author       = {Yiming Chen and
                  Yushi Lan and
                  Andrea Vedaldi},
  title        = {PhysiFormer: Learning to Simulate Mechanics in World Space},
  journal      = {CoRR},
  volume       = {abs/2606.27364},
  year         = {2026},
  url          = {https://doi.org/10.48550/arXiv.2606.27364},
  doi          = {10.48550/ARXIV.2606.27364},
  eprinttype   = {arXiv},
  eprint       = {2606.27364},
  bibsource    = {dblp computer science bibliography, https://dblp.org}
}

@article{DBLP:journals/corr/abs-2605-09196,
  author       = {Zhiyang Dou and
                  Minghao Guo and
                  Haixu Wu and
                  Doug Roble and
                  Tuur Stuyck and
                  Wojciech Matusik},
  title        = {RigidFormer: Learning Rigid Dynamics using Transformers},
  journal      = {CoRR},
  volume       = {abs/2605.09196},
  year         = {2026},
  url          = {https://doi.org/10.48550/arXiv.2605.09196},
  doi          = {10.48550/ARXIV.2605.09196},
  eprinttype   = {arXiv},
  eprint       = {2605.09196},
  bibsource    = {dblp computer science bibliography, https://dblp.org}
}

@article{DBLP:journals/corr/abs-2605-15305,
  author       = {Caoliwen Wang and
                  Minghao Guo and
                  Siyuan Chen and
                  Heng Zhang and
                  Mengdi Wang and
                  Xingyu Ni and
                  Hanson Sun and
                  Kunyi Wang and
                  Zherong Pan and
                  Kui Wu and
                  Lingjie Liu and
                  Yin Yang and
                  Chenfanfu Jiang and
                  Taku Komura and
                  Wojciech Matusik and
                  Peter Yichen Chen},
  title        = {WorldParticle: Unified World Simulation of Lagrangian Particle Dynamics
                  via Transformer},
  journal      = {CoRR},
  volume       = {abs/2605.15305},
  year         = {2026},
  url          = {https://doi.org/10.48550/arXiv.2605.15305},
  doi          = {10.48550/ARXIV.2605.15305},
  eprinttype   = {arXiv},
  eprint       = {2605.15305},
  bibsource    = {dblp computer science bibliography, https://dblp.org}
}

@inproceedings{DBLP:conf/iclr/TevetRGSCB23,
  author       = {Guy Tevet and
                  Sigal Raab and
                  Brian Gordon and
                  Yonatan Shafir and
                  Daniel Cohen{-}Or and
                  Amit Haim Bermano},
  title        = {Human Motion Diffusion Model},
  booktitle    = {The Eleventh International Conference on Learning Representations,
                  {ICLR} 2023, Kigali, Rwanda, May 1-5, 2023},
  publisher    = {OpenReview.net},
  year         = {2023},
  url          = {https://openreview.net/forum?id=SJ1kSyO2jwu},
  bibsource    = {dblp computer science bibliography, https://dblp.org}
}

@inproceedings{DBLP:conf/cvpr/CaoL0NT24,
  author       = {Wei Cao and
                  Chang Luo and
                  Biao Zhang and
                  Matthias Nie{\ss}ner and
                  Jiapeng Tang},
  title        = {Motion2VecSets: 4D Latent Vector Set Diffusion for Non-Rigid Shape
                  Reconstruction and Tracking},
  booktitle    = {{IEEE/CVF} Conference on Computer Vision and Pattern Recognition,
                  {CVPR} 2024, Seattle, WA, USA, June 16-22, 2024},
  pages        = {20496--20506},
  publisher    = {{IEEE}},
  year         = {2024},
  url          = {https://doi.org/10.1109/CVPR52733.2024.01937},
  doi          = {10.1109/CVPR52733.2024.01937},
  bibsource    = {dblp computer science bibliography, https://dblp.org}
}

@article{DBLP:journals/cgf/WiewelBT19,
  author       = {Steffen Wiewel and
                  Moritz Becher and
                  Nils Th{\"{u}}rey},
  title        = {Latent Space Physics: Towards Learning the Temporal Evolution of Fluid
                  Flow},
  journal      = {Comput. Graph. Forum},
  volume       = {38},
  number       = {2},
  pages        = {71--82},
  year         = {2019},
  url          = {https://doi.org/10.1111/cgf.13620},
  doi          = {10.1111/CGF.13620},
  bibsource    = {dblp computer science bibliography, https://dblp.org}
}

@inproceedings{DBLP:conf/iclr/HanGPWL22,
  author       = {Xu Han and
                  Han Gao and
                  Tobias Pfaff and
                  Jian{-}Xun Wang and
                  Liping Liu},
  title        = {Predicting Physics in Mesh-reduced Space with Temporal Attention},
  booktitle    = {The Tenth International Conference on Learning Representations, {ICLR}
                  2022, Virtual Event, April 25-29, 2022},
  publisher    = {OpenReview.net},
  year         = {2022},
  url          = {https://openreview.net/forum?id=XctLdNfCmP},
  bibsource    = {dblp computer science bibliography, https://dblp.org}
}

@article{DBLP:journals/tog/BarbicJ05,
  author       = {Jernej Barbic and
                  Doug L. James},
  title        = {Real-Time subspace integration for St. Venant-Kirchhoff deformable
                  models},
  journal      = {{ACM} Trans. Graph.},
  volume       = {24},
  number       = {3},
  pages        = {982--990},
  year         = {2005},
  url          = {https://doi.org/10.1145/1073204.1073300},
  doi          = {10.1145/1073204.1073300},
  bibsource    = {dblp computer science bibliography, https://dblp.org}
}

@article{DBLP:journals/cgf/FultonMDLJ19,
  author       = {Lawson Fulton and
                  Vismay Modi and
                  David Duvenaud and
                  David I. W. Levin and
                  Alec Jacobson},
  title        = {Latent-space Dynamics for Reduced Deformable Simulation},
  journal      = {Comput. Graph. Forum},
  volume       = {38},
  number       = {2},
  pages        = {379--391},
  year         = {2019},
  url          = {https://doi.org/10.1111/cgf.13645},
  doi          = {10.1111/CGF.13645},
  bibsource    = {dblp computer science bibliography, https://dblp.org}
}

@article{DBLP:journals/tog/RomeroCPO21,
  author       = {Cristian Romero and
                  Dan Casas and
                  Jes{\'{u}}s P{\'{e}}rez and
                  Miguel A. Otaduy},
  title        = {Learning contact corrections for handle-based subspace dynamics},
  journal      = {{ACM} Trans. Graph.},
  volume       = {40},
  number       = {4},
  pages        = {131:1--131:12},
  year         = {2021},
  url          = {https://doi.org/10.1145/3450626.3459875},
  doi          = {10.1145/3450626.3459875},
  bibsource    = {dblp computer science bibliography, https://dblp.org}
}

@inproceedings{DBLP:conf/iclr/ChenXCCPMCCG23,
  author       = {Peter Yichen Chen and
                  Jinxu Xiang and
                  Dong Heon Cho and
                  Yue Chang and
                  G. A. Pershing and
                  Henrique Teles Maia and
                  Maurizio M. Chiaramonte and
                  Kevin T. Carlberg and
                  Eitan Grinspun},
  title        = {{CROM:} Continuous Reduced-Order Modeling of PDEs Using Implicit Neural
                  Representations},
  booktitle    = {The Eleventh International Conference on Learning Representations,
                  {ICLR} 2023, Kigali, Rwanda, May 1-5, 2023},
  publisher    = {OpenReview.net},
  year         = {2023},
  url          = {https://openreview.net/forum?id=FUORz1tG8Og},
  bibsource    = {dblp computer science bibliography, https://dblp.org}
}

@inproceedings{DBLP:conf/siggrapha/ChangCWCCG23,
  author       = {Yue Chang and
                  Peter Yichen Chen and
                  Zhecheng Wang and
                  Maurizio M. Chiaramonte and
                  Kevin Carlberg and
                  Eitan Grinspun},
  editor       = {June Kim and
                  Ming C. Lin and
                  Bernd Bickel},
  title        = {LiCROM: Linear-Subspace Continuous Reduced Order Modeling with Neural
                  Fields},
  booktitle    = {{SIGGRAPH} Asia 2023 Conference Papers, {SA} 2023, Sydney, NSW, Australia,
                  December 12-15, 2023},
  pages        = {111:1--111:12},
  publisher    = {{ACM}},
  year         = {2023},
  url          = {https://doi.org/10.1145/3610548.3618158},
  doi          = {10.1145/3610548.3618158},
  bibsource    = {dblp computer science bibliography, https://dblp.org}
}

@article{DBLP:journals/tog/ModiSPSL24,
  author       = {Vismay Modi and
                  Nicholas Sharp and
                  Or Perel and
                  Shinjiro Sueda and
                  David I. W. Levin},
  title        = {Simplicits: Mesh-Free, Geometry-Agnostic Elastic Simulation},
  journal      = {{ACM} Trans. Graph.},
  volume       = {43},
  number       = {4},
  pages        = {117:1--117:11},
  year         = {2024},
  url          = {https://doi.org/10.1145/3658184},
  doi          = {10.1145/3658184},
  bibsource    = {dblp computer science bibliography, https://dblp.org}
}

@inproceedings{DBLP:conf/siggraph/LiaoSP26,
  author       = {Li Liao and
                  Pengfei Shen and
                  Yifan Peng},
  editor       = {Christopher P. Redmann and
                  Justin Solomon and
                  Mirela Ben{-}Chen},
  title        = {Boundary-aware Neural Model Reduction for PDEs},
  booktitle    = {Proceedings of the Special Interest Group on Computer Graphics and
                  Interactive Techniques Conference, {SIGGRAPH} Conference Papers 2026,
                  Los Angeles, CA, USA, July 19-23, 2026},
  pages        = {5:1--5:11},
  publisher    = {{ACM}},
  year         = {2026},
  url          = {https://doi.org/10.1145/3799902.3811153},
  doi          = {10.1145/3799902.3811153},
  bibsource    = {dblp computer science bibliography, https://dblp.org}
}

@article{DBLP:journals/corr/abs-2605-30347,
  author       = {Chen Geng and
                  Guangzhao He and
                  Yue Gao and
                  Yunzhi Zhang and
                  Shangzhe Wu and
                  Jiajun Wu},
  title        = {NeuROK: Generative 4D Neural Object Kinematics},
  journal      = {CoRR},
  volume       = {abs/2605.30347},
  year         = {2026},
  url          = {https://doi.org/10.48550/arXiv.2605.30347},
  doi          = {10.48550/ARXIV.2605.30347},
  eprinttype   = {arXiv},
  eprint       = {2605.30347},
  bibsource    = {dblp computer science bibliography, https://dblp.org}
}

@inproceedings{DBLP:conf/iccv/Arnab0H0LS21,
  author       = {Anurag Arnab and
                  Mostafa Dehghani and
                  Georg Heigold and
                  Chen Sun and
                  Mario Lucic and
                  Cordelia Schmid},
  title        = {ViViT: {A} Video Vision Transformer},
  booktitle    = {2021 {IEEE/CVF} International Conference on Computer Vision, {ICCV}
                  2021, Montreal, QC, Canada, October 10-17, 2021},
  pages        = {6816--6826},
  publisher    = {{IEEE}},
  year         = {2021},
  url          = {https://doi.org/10.1109/ICCV48922.2021.00676},
  doi          = {10.1109/ICCV48922.2021.00676},
  bibsource    = {dblp computer science bibliography, https://dblp.org}
}

@inproceedings{DBLP:conf/icml/BertasiusWT21,
  author       = {Gedas Bertasius and
                  Heng Wang and
                  Lorenzo Torresani},
  editor       = {Marina Meila and
                  Tong Zhang},
  title        = {Is Space-Time Attention All You Need for Video Understanding?},
  booktitle    = {Proceedings of the 38th International Conference on Machine Learning,
                  {ICML} 2021, 18-24 July 2021, Virtual Event},
  series       = {Proceedings of Machine Learning Research},
  volume       = {139},
  pages        = {813--824},
  publisher    = {{PMLR}},
  year         = {2021},
  url          = {http://proceedings.mlr.press/v139/bertasius21a.html},
  bibsource    = {dblp computer science bibliography, https://dblp.org}
}

@inproceedings{DBLP:conf/nips/HoJA20,
  author       = {Jonathan Ho and
                  Ajay Jain and
                  Pieter Abbeel},
  editor       = {Hugo Larochelle and
                  Marc'Aurelio Ranzato and
                  Raia Hadsell and
                  Maria{-}Florina Balcan and
                  Hsuan{-}Tien Lin},
  title        = {Denoising Diffusion Probabilistic Models},
  booktitle    = {Advances in Neural Information Processing Systems 33: Annual Conference
                  on Neural Information Processing Systems 2020, NeurIPS 2020, December
                  6-12, 2020, virtual},
  year         = {2020},
  url          = {https://proceedings.neurips.cc/paper/2020/hash/4c5bcfec8584af0d967f1ab10179ca4b-Abstract.html},
  bibsource    = {dblp computer science bibliography, https://dblp.org}
}

@inproceedings{DBLP:conf/iccv/PeeblesX23,
  author       = {William Peebles and
                  Saining Xie},
  title        = {Scalable Diffusion Models with Transformers},
  booktitle    = {{IEEE/CVF} International Conference on Computer Vision, {ICCV} 2023,
                  Paris, France, October 1-6, 2023},
  pages        = {4172--4182},
  publisher    = {{IEEE}},
  year         = {2023},
  url          = {https://doi.org/10.1109/ICCV51070.2023.00387},
  doi          = {10.1109/ICCV51070.2023.00387},
  bibsource    = {dblp computer science bibliography, https://dblp.org}
}

@inproceedings{DBLP:conf/iclr/LipmanCBNL23,
  author       = {Yaron Lipman and
                  Ricky T. Q. Chen and
                  Heli Ben{-}Hamu and
                  Maximilian Nickel and
                  Matthew Le},
  title        = {Flow Matching for Generative Modeling},
  booktitle    = {The Eleventh International Conference on Learning Representations,
                  {ICLR} 2023, Kigali, Rwanda, May 1-5, 2023},
  publisher    = {OpenReview.net},
  year         = {2023},
  url          = {https://openreview.net/forum?id=PqvMRDCJT9t},
  bibsource    = {dblp computer science bibliography, https://dblp.org}
}

@inproceedings{DBLP:conf/iclr/LiuG023,
  author       = {Xingchao Liu and
                  Chengyue Gong and
                  Qiang Liu},
  title        = {Flow Straight and Fast: Learning to Generate and Transfer Data with
                  Rectified Flow},
  booktitle    = {The Eleventh International Conference on Learning Representations,
                  {ICLR} 2023, Kigali, Rwanda, May 1-5, 2023},
  publisher    = {OpenReview.net},
  year         = {2023},
  url          = {https://openreview.net/forum?id=XVjTT1nw5z},
  bibsource    = {dblp computer science bibliography, https://dblp.org}
}

@article{DBLP:journals/corr/abs-1912-12180,
  author       = {Jonathan Ho and
                  Nal Kalchbrenner and
                  Dirk Weissenborn and
                  Tim Salimans},
  title        = {Axial Attention in Multidimensional Transformers},
  journal      = {CoRR},
  volume       = {abs/1912.12180},
  year         = {2019},
  url          = {http://arxiv.org/abs/1912.12180},
  eprinttype   = {arXiv},
  eprint       = {1912.12180},
  bibsource    = {dblp computer science bibliography, https://dblp.org}
}

@inproceedings{DBLP:conf/siggraph/TerzopoulosPBF87,
  author       = {Demetri Terzopoulos and
                  John C. Platt and
                  Alan H. Barr and
                  Kurt W. Fleischer},
  editor       = {Maureen C. Stone},
  title        = {Elastically deformable models},
  booktitle    = {Proceedings of the 14th Annual Conference on Computer Graphics and
                  Interactive Techniques, {SIGGRAPH} 1987, Anaheim, California, USA,
                  July 27-31, 1987},
  pages        = {205--214},
  publisher    = {{ACM}},
  year         = {1987},
  url          = {https://doi.org/10.1145/37401.37427},
  doi          = {10.1145/37401.37427},
  bibsource    = {dblp computer science bibliography, https://dblp.org}
}

@article{DBLP:journals/tog/BridsonFA02,
  author       = {Robert Bridson and
                  Ronald Fedkiw and
                  John Anderson},
  title        = {Robust treatment of collisions, contact and friction for cloth animation},
  journal      = {{ACM} Trans. Graph.},
  volume       = {21},
  number       = {3},
  pages        = {594--603},
  year         = {2002},
  url          = {https://doi.org/10.1145/566654.566623},
  doi          = {10.1145/566654.566623},
  bibsource    = {dblp computer science bibliography, https://dblp.org}
}

@inproceedings{DBLP:conf/sca/IrvingTF04,
  author       = {Geoffrey Irving and
                  Joseph Teran and
                  Ronald Fedkiw},
  editor       = {Norman I. Badler and
                  Mathieu Desbrun and
                  Ronan Boulic and
                  Dinesh K. Pai},
  title        = {Invertible finite elements for robust simulation of large deformation},
  booktitle    = {Proceedings of the 2004 {ACM} SIGGRAPH/Eurographics Symposium on Computer
                  Animation, Grenoble, France, August 27-29, 2004},
  pages        = {131--140},
  publisher    = {The Eurographics Association},
  year         = {2004},
  url          = {https://doi.org/10.2312/SCA/SCA04/131-140},
  doi          = {10.2312/SCA/SCA04/131-140},
  bibsource    = {dblp computer science bibliography, https://dblp.org}
}

@article{DBLP:journals/jvcir/MullerHHR07,
  author       = {Matthias M{\"{u}}ller and
                  Bruno Heidelberger and
                  Marcus Hennix and
                  John Ratcliff},
  title        = {Position based dynamics},
  journal      = {J. Vis. Commun. Image Represent.},
  volume       = {18},
  number       = {2},
  pages        = {109--118},
  year         = {2007},
  url          = {https://doi.org/10.1016/j.jvcir.2007.01.005},
  doi          = {10.1016/J.JVCIR.2007.01.005},
  bibsource    = {dblp computer science bibliography, https://dblp.org}
}

@article{DBLP:journals/tog/BouazizMLKP14,
  author       = {Sofien Bouaziz and
                  Sebastian Martin and
                  Tiantian Liu and
                  Ladislav Kavan and
                  Mark Pauly},
  title        = {Projective dynamics: fusing constraint projections for fast simulation},
  journal      = {{ACM} Trans. Graph.},
  volume       = {33},
  number       = {4},
  pages        = {154:1--154:11},
  year         = {2014},
  url          = {https://doi.org/10.1145/2601097.2601116},
  doi          = {10.1145/2601097.2601116},
  bibsource    = {dblp computer science bibliography, https://dblp.org}
}

@inproceedings{DBLP:conf/nips/HoSGC0F22,
  author       = {Jonathan Ho and
                  Tim Salimans and
                  Alexey A. Gritsenko and
                  William Chan and
                  Mohammad Norouzi and
                  David J. Fleet},
  editor       = {Sanmi Koyejo and
                  S. Mohamed and
                  A. Agarwal and
                  Danielle Belgrave and
                  K. Cho and
                  A. Oh},
  title        = {Video Diffusion Models},
  booktitle    = {Advances in Neural Information Processing Systems 35: Annual Conference
                  on Neural Information Processing Systems 2022, NeurIPS 2022, New Orleans,
                  LA, USA, November 28 - December 9, 2022},
  year         = {2022},
  url          = {http://papers.nips.cc/paper\_files/paper/2022/hash/39235c56aef13fb05a6adc95eb9d8d66-Abstract-Conference.html},
  bibsource    = {dblp computer science bibliography, https://dblp.org}
}

@article{DBLP:journals/tog/HuangTGFSPZ24,
  author       = {Zizhou Huang and
                  Davi Colli Tozoni and
                  Arvi Gjoka and
                  Zachary Ferguson and
                  Teseo Schneider and
                  Daniele Panozzo and
                  Denis Zorin},
  title        = {Differentiable solver for time-dependent deformation problems with
                  contact},
  journal      = {{ACM} Trans. Graph.},
  volume       = {43},
  number       = {3},
  pages        = {31:1--31:30},
  year         = {2024},
  url          = {https://doi.org/10.1145/3657648},
  doi          = {10.1145/3657648},
  bibsource    = {dblp computer science bibliography, https://dblp.org}
}

@inproceedings{DBLP:conf/cvpr/KendallGC18,
  author       = {Alex Kendall and
                  Yarin Gal and
                  Roberto Cipolla},
  title        = {Multi-Task Learning Using Uncertainty to Weigh Losses for Scene Geometry
                  and Semantics},
  booktitle    = {2018 {IEEE} Conference on Computer Vision and Pattern Recognition,
                  {CVPR} 2018, Salt Lake City, UT, USA, June 18-22, 2018},
  pages        = {7482--7491},
  publisher    = {Computer Vision Foundation / {IEEE} Computer Society},
  year         = {2018},
  url          = {http://openaccess.thecvf.com/content\_cvpr\_2018/html/Kendall\_Multi-Task\_Learning\_Using\_CVPR\_2018\_paper.html},
  doi          = {10.1109/CVPR.2018.00781},
  bibsource    = {dblp computer science bibliography, https://dblp.org}
}

@inproceedings{DBLP:conf/nips/KingmaSJCCSW16,
  author       = {Diederik P. Kingma and
                  Tim Salimans and
                  Rafal J{\'{o}}zefowicz and
                  Xi Chen and
                  Ilya Sutskever and
                  Max Welling},
  editor       = {Daniel D. Lee and
                  Masashi Sugiyama and
                  Ulrike von Luxburg and
                  Isabelle Guyon and
                  Roman Garnett},
  title        = {Improving Variational Autoencoders with Inverse Autoregressive Flow},
  booktitle    = {Advances in Neural Information Processing Systems 29: Annual Conference
                  on Neural Information Processing Systems 2016, December 5-10, 2016,
                  Barcelona, Spain},
  pages        = {4736--4744},
  year         = {2016},
  url          = {https://proceedings.neurips.cc/paper/2016/hash/ddeebdeefdb7e7e7a697e1c3e3d8ef54-Abstract.html},
  bibsource    = {dblp computer science bibliography, https://dblp.org}
}

@article{DBLP:journals/pami/Umeyama91,
  author       = {Shinji Umeyama},
  title        = {Least-Squares Estimation of Transformation Parameters Between Two
                  Point Patterns},
  journal      = {{IEEE} Trans. Pattern Anal. Mach. Intell.},
  volume       = {13},
  number       = {4},
  pages        = {376--380},
  year         = {1991},
  url          = {https://doi.org/10.1109/34.88573},
  doi          = {10.1109/34.88573},
  bibsource    = {dblp computer science bibliography, https://dblp.org}
}

@misc{newton2025,
  author       = {{The Newton Contributors}},
  title        = {Newton: {GPU}-accelerated physics simulation for robotics and simulation research},
  year         = {2025},
  howpublished = {\url{https://github.com/newton-physics/newton}}
}

@article{DBLP:journals/tog/HuSWZP20,
  author       = {Yixin Hu and
                  Teseo Schneider and
                  Bolun Wang and
                  Denis Zorin and
                  Daniele Panozzo},
  title        = {Fast tetrahedral meshing in the wild},
  journal      = {{ACM} Trans. Graph.},
  volume       = {39},
  number       = {4},
  pages        = {117},
  year         = {2020},
  url          = {https://doi.org/10.1145/3386569.3392385},
  doi          = {10.1145/3386569.3392385},
  bibsource    = {dblp computer science bibliography, https://dblp.org}
}

@inproceedings{DBLP:conf/icra/KimDHG22,
  author       = {Chung Min Kim and
                  Michael Danielczuk and
                  Isabella Huang and
                  Ken Goldberg},
  title        = {IPC-GraspSim: Reducing the Sim2Real Gap for Parallel-Jaw Grasping
                  with the Incremental Potential Contact Model},
  booktitle    = {2022 International Conference on Robotics and Automation, {ICRA} 2022,
                  Philadelphia, PA, USA, May 23-27, 2022},
  pages        = {6180--6187},
  publisher    = {{IEEE}},
  year         = {2022},
  url          = {https://doi.org/10.1109/ICRA46639.2022.9811777},
  doi          = {10.1109/ICRA46639.2022.9811777},
  bibsource    = {dblp computer science bibliography, https://dblp.org}
}

@inproceedings{DBLP:conf/nips/LiDYLZLJZH25,
  author       = {Yuyang Li and
                  Wenxin Du and
                  Chang Yu and
                  Puhao Li and
                  Zihang Zhao and
                  Tengyu Liu and
                  Chenfanfu Jiang and
                  Yixin Zhu and
                  Siyuan Huang},
  editor       = {Danielle Belgrave and
                  Cheng Zhang and
                  Laura N. Montoya and
                  Hsuan{-}Tien Lin and
                  Razvan Pascanu and
                  Piotr Koniusz and
                  Marzyeh Ghassemi and
                  Nancy Chen and
                  Iv{\'{a}}n Vladimir Meza Ru{\'{\i}}z and
                  Arturo Loaiza{-}Bonilla},
  title        = {Taccel: Scaling Up Vision-based Tactile Robotics via High-performance
                  {GPU} Simulation},
  booktitle    = {Advances in Neural Information Processing Systems 38: Annual Conference
                  on Neural Information Processing Systems 2025, NeurIPS 2025, San Diego,
                  CA, USA, December 2-7, 2025 / Mexico City, Mexico, November 30 - December
                  5, 2025},
  year         = {2025},
  url          = {http://papers.nips.cc/paper\_files/paper/2025/hash/87d91d52272c3166315ca20b18519b9e-Abstract-Conference.html},
  bibsource    = {dblp computer science bibliography, https://dblp.org}
}

@inproceedings{DBLP:conf/sgp/SorkineA07,
  author       = {Olga Sorkine and
                  Marc Alexa},
  editor       = {Alexander G. Belyaev and
                  Michael Garland},
  title        = {As-rigid-as-possible surface modeling},
  booktitle    = {Proceedings of the Fifth Eurographics Symposium on Geometry Processing,
                  Barcelona, Spain, July 4-6, 2007},
  series       = {{ACM} International Conference Proceeding Series},
  volume       = {257},
  pages        = {109--116},
  publisher    = {Eurographics Association},
  year         = {2007},
  url          = {https://doi.org/10.2312/SGP/SGP07/109-116},
  doi          = {10.2312/SGP/SGP07/109-116},
  bibsource    = {dblp computer science bibliography, https://dblp.org}
}

@article{DBLP:journals/tog/SmithGK18,
  author       = {Breannan Smith and
                  Fernando de Goes and
                  Theodore Kim},
  title        = {Stable Neo-Hookean Flesh Simulation},
  journal      = {{ACM} Trans. Graph.},
  volume       = {37},
  number       = {2},
  pages        = {12},
  year         = {2018},
  url          = {https://doi.org/10.1145/3180491},
  doi          = {10.1145/3180491},
  bibsource    = {dblp computer science bibliography, https://dblp.org}
}

@inproceedings{DBLP:conf/iccv/JiangHZYWL25,
  author       = {Hanxiao Jiang and
                  Hao{-}Yu Hsu and
                  Kaifeng Zhang and
                  Hsin{-}Ni Yu and
                  Shenlong Wang and
                  Yunzhu Li},
  title        = {PhysTwin: Physics-Informed Reconstruction and Simulation of Deformable
                  Objects from Videos},
  booktitle    = {{IEEE/CVF} International Conference on Computer Vision, {ICCV} 2025,
                  Honolulu, HI, USA, October 19-25, 2025},
  pages        = {7219--7230},
  publisher    = {{IEEE}},
  year         = {2025},
  url          = {https://doi.org/10.1109/ICCV51701.2025.00678},
  doi          = {10.1109/ICCV51701.2025.00678},
  bibsource    = {dblp computer science bibliography, https://dblp.org}
}

\clearpage
\appendix
\section*{Appendix}
\S A specifies the architecture, losses, and training of all three stages. \S B gives the data, evaluation setup, and metric definitions. \S C details the staircase analysis of standard temporal compression. \S D the chaos experiment on a traditional GPU simulator. \S E describes the baselines. \S F collects further ablations. \S G reports inference runtime. \S H shows additional qualitative results. \S I analyzes the extreme cases. \S J the multi-body extension. \S K the setup and results of the differentiable applications. \S L the comparison with concurrent work. \S M tests the constitutive-model-agnostic property on a second energy model and on real captured trajectories.

\addtocontents{toc}{\protect\setcounter{tocdepth}{2}}
\renewcommand{\contentsname}{Contents}
\tableofcontents
\bigskip

\FloatBarrier
\section{Architecture and Training Details}
\label{app:arch}

Our model is trained in three stages, one per component: the spatial VAE for $183$k iterations, ST-VAE Base for $150$k iterations, and the LDM for $278$k iterations. The three stages together take about $7$ days on $8$ H200 GPUs.

\subsection{Stage 1: Spatial VAE}
The spatial VAE is an $8$-layer transformer of hidden width $512$ ($44$M parameters). Its input is one frame in canonical coordinates, i.e.\ the rest shape plus the deformation $\mathbf{d}_t$ of Eq.~\eqref{eq:factoring} with the rigid motion removed. The encoder cross-attends $K{=}512$ learned query tokens against the $N$ vertex tokens of the frame and outputs $K$ latent tokens of width $D{=}48$, independent of $N$. We denote the latent of frame $t$ by $\mathbf{z}_t$. The decoder mirrors the encoder: the $N$ rest vertices, embedded with their positions, cross-attend the latent tokens and are decoded back to their $N$ displacements. The decoder additionally takes the mesh topology as input, because a shallow GNN over the tetrahedral mesh then refines the decoded displacements. The spatial VAE is trained first with a Charbonnier position loss and a KL term.

\subsection{Stage 2: ST-VAE}
The per-frame latents $\{\mathbf{z}_t\}_{t=1}^{T}$ are processed by a stack of factorized space--time attention blocks \citep{DBLP:journals/corr/abs-1912-12180,DBLP:conf/icml/BertasiusWT21,DBLP:conf/iccv/Arnab0H0LS21}. The holistic temporal compression then summarizes the time axis: $M$ learned query tokens attend over each spatial token's $T$-frame sequence, and only these $M$ outputs are kept as the temporal representation, while the $T$ per-frame outputs are discarded. The rigid-motion token bypasses this attention compression and is compressed by a separate linear temporal path. The two are kept apart because the rigid signal is orders of magnitude larger than the deformation and, when the two share one compression path, takes over the shared latent and starves the deformation track (Sec.~\ref{sec:ablations}, Appendix~\ref{app:moreabl}). The two tracks then merge: the joint latent is $\mathbf{Z}\in\mathbb{R}^{M\times(K+1)\times D}$ of Sec.~\ref{sec:arch}, i.e.\ $M$ temporal tokens for each of the $K$ deformation tokens and the rigid token, each of width $D{=}48$. The decoder mirrors the encoder: per-frame queries attend over the $M$ temporal tokens to recover a latent for every frame, which the spatial VAE decoder turns into vertex positions. Table~\ref{tab:tiers} lists the two ST-VAE configurations used in this paper, ST-VAE Small for the controlled comparisons of Table~\ref{tab:s2main} and ST-VAE Base for the LDM experiments (Sec.~\ref{sec:exp}). Both configurations compress a trajectory by the same factor, counted in tokens. The input has one token per vertex per frame, $T\times N$ tokens in total, and the latent has $M\times(K{+}1)$ tokens. Since $T/M{=}8$ and $K{+}1{=}513$ in both configurations, the token compression ratio is $8N/513$, which reaches $78\times$ at the $5{,}000$-vertex cap.

\begin{table}[t]
\centering
\small
\caption{The two ST-VAE configurations used in this paper. ST-VAE Small is used for the reconstruction comparisons of Table~\ref{tab:s2main}. ST-VAE Base is the ST-VAE behind PhysLDM, i.e.\ the LDM experiments and the applications.}
\label{tab:tiers}
\begin{tabular}{@{}lcc@{}}
\toprule
 & ST-VAE Small & ST-VAE Base \\
\midrule
Query tokens $K$ & 512 & 512 \\
Window $T$ & 32 & 64 \\
Temporal tokens $M$ & 4 & 8 \\
Temporal compression & $8\times$ & $8\times$ \\
Hidden width & 512 & 512 \\
Joint latent $M\times(K{+}1)\times D$ & $4\times513\times48$ & $8\times513\times48$ \\
Parameters (temporal stage) & $51$M & $80$M \\
Parameters (total, incl.\ spatial VAE) & $95$M & $124$M \\
Spatial VAE & frozen & jointly tuned \\
Role & controlled comparisons & all downstream stages \\
\bottomrule
\end{tabular}
\end{table}

\subsection{Stage 3: LDM}
The LDM and the regression model receive the same conditioning: the frame-$0$ spatial latent through cross-attention, and the initial pose, the initial velocity $v_0$, and the Young's modulus $E$ through adaptive layer normalization. The LDM is a DiT with $12$ blocks of width $768$ ($170$M parameters) operating on the $4{,}104$ latent tokens of ST-VAE Base ($M\times(K{+}1)=8\times513$ tokens of width $D{=}48$). It is trained with rectified flow and sampled by Euler integration. The regression model is a capacity-matched transformer that emits the full latent trajectory in one pass. Training the spatial VAE, ST-VAE Base, and the LDM takes about $7$ days in total on $8$ H200 GPUs.

\subsection{Loss details}
All position losses use a Charbonnier penalty on the per-vertex error norm,
\begin{equation}
\mathcal{L}_{\mathrm{charb}}(\hat{\mathbf{x}},\mathbf{x}) \;=\; \frac{1}{|V|}\sum_{i\in V}\sqrt{\lVert\hat{\mathbf{x}}_i-\mathbf{x}_i\rVert^2+\epsilon^2},\qquad \epsilon=10^{-3}\ (=1\,\mathrm{mm}),
\end{equation}

\emph{Stage 1: Spatial VAE.} $\mathcal{L}_{\mathrm{S1}}=\mathcal{L}_{\mathrm{charb}}+\beta_1\mathcal{L}_{\mathrm{KL}}$ with a standard Gaussian KL term and $\beta_1=10^{-5}$.

\emph{Stage 2: ST-VAE.} The objective combines the deformation reconstruction of all $T$ frames, the rigid track, and a KL term:
\begin{equation}
\mathcal{L}_{\mathrm{S2}} = \sum_{j\in\{\mathrm{deform},R,c\}} \Big(\tfrac{1}{2}e^{-s_j}\mathcal{L}_j + \tfrac{1}{2}s_j\Big) + \beta_2\,\mathcal{L}_{\mathrm{KL}}^{\mathrm{fb}},
\end{equation}
where $\mathcal{L}_{\mathrm{deform}}$ is $\mathcal{L}_{\mathrm{charb}}$ averaged over the $T$ decoded frames, $\mathcal{L}_R$ and $\mathcal{L}_c$ are mean-squared errors on the rigid token's rotation and translation, and the three are balanced by the learned uncertainty weights $s_j$ of \citet{DBLP:conf/cvpr/KendallGC18}. $\mathcal{L}_{\mathrm{KL}}^{\mathrm{fb}}$ is the KL term with per-channel free bits \citep{DBLP:conf/nips/KingmaSJCCSW16}.

\emph{Stage 3: LDM.} The latent is standardized per modality (rigid and deformation channels separately). With flow time $\tau$, noise $\boldsymbol{\epsilon}\sim\mathcal{N}(0,\mathbf{I})$ and the straight path $\mathbf{x}_\tau=(1{-}\tau)\mathbf{z}+\tau\boldsymbol{\epsilon}$, the LDM minimizes a token-weighted velocity regression
\begin{equation}
\mathcal{L}_{\mathrm{DiT}} = \mathbb{E}_{\tau,\boldsymbol{\epsilon}}\Big[\big\langle w_k\,\lVert\hat{\mathbf{v}}_k-(\boldsymbol{\epsilon}-\mathbf{z})_k\rVert^2\big\rangle_k\Big],\qquad
w_k=\begin{cases}\lambda_{\mathrm{rigid}} & k\ \text{a rigid token}\\ \lambda_{\mathrm{deform}} & \text{otherwise},\end{cases}
\end{equation}
where $k$ indexes the $(K{+}1)\times M$ latent tokens and $\lambda_{\mathrm{rigid}}$ is a hand-tuned weight that up-weights the rigid token relative to the deformation tokens ($\lambda_{\mathrm{deform}}{=}1$). The regression model minimizes the same token-weighted mean-squared error directly on the latent.

\FloatBarrier
\section{Data, Evaluation Setup, and Metrics}
\label{app:metrics}
\subsection{Data details}
\label{app:data}
All simulation data are generated with StiffGIPC \citep{DBLP:journals/tog/HuangLLKL25}, a GPU IPC solver. IPC solvers are widely adopted in scenarios that demand high simulation accuracy, such as embodied AI \citep{DBLP:conf/icra/KimDHG22,DBLP:conf/nips/LiDYLZLJZH25}. Such a solver runs offline. Learning its behavior and turning it into a fast, fully differentiable neural simulator is therefore of practical importance. All 3D meshes used for training come from the Objaverse dataset \citep{DBLP:conf/cvpr/DeitkeSSWMVSEKF23}. Each asset is voxel-remeshed to a watertight surface, tetrahedralized with fTetWild \citep{DBLP:journals/tog/HuSWZP20}, reduced to its largest connected component, and rescaled to a unit bounding box. Meshes with more than $10{,}000$ vertices are discarded, and the ST-VAE and the LDM are trained on scenes with $500$--$5{,}000$ vertices. Every scene drops one meter-scale solid onto a ground plane. The object is released with its lowest point $0.85$ to $1.65$\,m above the ground. Each object receives a random initial velocity $v_0$ and a random stiffness $E$ (a horizontal $v_0\sim U(0,2)$\,m/s and a Young's modulus $E$ log-uniform in $[3\times10^{4},3\times10^{6}]$\,Pa) under a Stable Neo-Hookean material. Multi-body scenes follow the same setting with two or three objects, each with its own $E$ and $v_0$, thrown toward each other so that they collide in mid-air before reaching the ground. In total, the dataset contains $63{,}397$ single-object and $16{,}078$ multi-body simulated trajectories. The simulation time step is $1/60$\,s, and each trajectory has $64$ frames. Held-out scenes are excluded from all three training stages, and every reported number is computed on these held-out scenes or on the OOD sets. For OOD evaluation we generate $1{,}015$ GSO and $3{,}814$ Toys4K scenes under the identical simulation setting. They never appear in training.
\subsection{Evaluation setup}
(a) \emph{ST-VAE reconstruction (Table~\ref{tab:s2main}).} For each evaluation set (held-out, GSO, Toys4K), we randomly draw $300$ scenes with $500\le N\le5{,}000$ vertices for the main columns and $300$ scenes with $N>5{,}000$ vertices for the large-mesh columns. Every model is evaluated on the same scenes. (b) \emph{Prediction (Table~\ref{tab:s3phys}).} The same $300$ scenes per set within $500\le N\le5{,}000$. The LDM uses $50$ Euler steps. Every metric is a per-scene value, reported as the median across scenes to limit the influence of a few extreme values.

\subsection{\texorpdfstring{Metrics of Table~\ref{tab:s2main}}{Metrics of the reconstruction table}}
\label{app:metrics-t1}
\begin{itemize}
\item \textbf{DCC} (deformation-change correlation): the deformation change of a trajectory at frame $t$ is the mean vertex displacement between two consecutive frames in canonical (rigid-free) coordinates, $v(t)=\mathrm{mean}_v\lVert \mathbf{d}_{t+1,v}-\mathbf{d}_{t,v}\rVert$, where $\mathbf{d}_{t,v}$ is the position of vertex $v$ in the deformation field $\mathbf{d}_t$. The deformation-change curve is $v(t)$ over the trajectory. DCC is the Pearson correlation between the deformation-change curves of the reconstruction and of the ground truth. It is reported as the mean across scenes. It detects the staircase artifact directly, since holds suppress the speed and snaps spike it. This metric is amplitude-invariant, so it does not reflect reconstruction quality on its own and should be considered together with the other metrics.
\item \textbf{Recon} (geometric reconstruction error; mean / med / P90): the encode--decode error of the deformation field, i.e.\ the per-vertex Euclidean error of the rigid-free deformation $\mathbf{d}_t$ averaged over vertices and frames of a scene (mm). The three columns are the mean, median, and 90th percentile across scenes. The 90th percentile reflects the hardest $10\%$ of scenes. For the direct world-coordinate baseline, which has no deformation field of its own, a rigid transform is fitted to each output frame by the Kabsch algorithm, and the residual after removing it is scored as the deformation error.
\item \textbf{rot} (rotation error): the geodesic $\mathrm{SO}(3)$ distance between the predicted and ground-truth per-frame rotation (relative to frame $0$), averaged over the frames of a scene and then over scenes (degrees). For the direct world-coordinate baseline, the rotation and translation are those of the per-frame fitted rigid transform.
\item \textbf{trans} (translation error): the centroid translation error (relative to frame $0$), averaged over frames and then over scenes (mm).
\end{itemize}

\subsection{\texorpdfstring{Metrics of Table~\ref{tab:s3phys}}{Metrics of the prediction table}}
\label{app:metrics_s3}
All of these metrics are computed from the predicted and ground-truth trajectories alone and do not require knowledge of the underlying constitutive model. The violation rates are defined as follows. A physical quantity (speed or acceleration) is computed for every vertex at every frame, once from the prediction and once from the ground truth. Each such vertex-frame pair is one sample. A sample is counted as a violation if the predicted value is more than twice, or less than half, the ground-truth value. The violation rate is the fraction of violated samples. Samples whose ground-truth value is too small to be meaningful are excluded by a magnitude threshold.
\begin{itemize}
\item \textbf{Volume viol.} (volume violation rate): this measures whether the prediction preserves volume where the ground truth does. For every tetrahedron at every frame, we compute how much its volume differs from its rest volume, once in the ground truth and once in the prediction. We look only at the cases where the ground truth changes the volume by less than $3\%$, because there the ground truth preserves volume. Among these cases, a violation is one where the prediction changes the volume by more than $10\%$. The rate is the fraction of violations.
\item \textbf{Self-int.} (self-intersection rate): the percentage of surface triangles of the tetrahedral mesh that cross another surface triangle, averaged over frames and reported as the median across scenes. The ground truth has no self-intersection in any frame, so its value is $0$.
\item \textbf{Vel.\ viol.} and \textbf{Acc.\ viol.} (velocity and acceleration violation rates): the violation rate of the per-vertex speed and acceleration magnitudes. The acceleration threshold excludes free fall and settled frames.
\item \textbf{$KE_{\mathrm{def}}/KE_{\mathrm{rig}}$} (deformation-to-rigid kinetic energy ratio): a rigid transform $(\mathbf{R}_t,\mathbf{c}_t)$ is fitted to every frame by the Kabsch algorithm, which gives the rigid trajectory $\mathbf{r}_{t,v}=\mathbf{R}_t\bar{\mathbf{x}}_v+\mathbf{c}_t$ of the rest shape $\bar{\mathbf{x}}$. The vertex velocity $\mathbf{v}_{t,v}$ (central difference of $\mathbf{x}_{t,v}$) is split into the rigid velocity $\mathbf{v}^{\mathrm{rig}}_{t,v}$ (central difference of $\mathbf{r}_{t,v}$) and the deformation residual $\mathbf{v}^{\mathrm{def}}_{t,v}=\mathbf{v}_{t,v}-\mathbf{v}^{\mathrm{rig}}_{t,v}$. With lumped vertex masses $m_v$ and the impact window $W$ ($\pm 10$ frames around first contact), the two kinetic energies are
\begin{equation*}
KE_{\mathrm{def}}=\tfrac{1}{2}\sum_{t\in W}\sum_v m_v\lVert\mathbf{v}^{\mathrm{def}}_{t,v}\rVert^2,\qquad
KE_{\mathrm{rig}}=\tfrac{1}{2}\sum_{t\in W}\sum_v m_v\lVert\mathbf{v}^{\mathrm{rig}}_{t,v}\rVert^2,
\end{equation*}
and the reported value is $(KE_{\mathrm{def}}/KE_{\mathrm{rig}})_{\mathrm{pred}}\,/\,(KE_{\mathrm{def}}/KE_{\mathrm{rig}})_{\mathrm{GT}}$, taken as the median across scenes. Scenes where the ground truth barely deforms (deformation below $1\%$ of the kinetic energy) are skipped. At impact, the kinetic energy of the object is divided between the motion of the object as a whole and the internal motion that deforms it. How much goes into deformation depends on how soft the object is and how hard it hits. The ground truth gives the correct proportion for the same object and the same impact, so comparing the two ratios shows whether the prediction moves energy into deformation the way the GPU solver does. A value below $1$ means the prediction deforms less than the ground truth and moves more rigidly. A value above $1$ would mean unphysical deformation or jitter. In Table~\ref{tab:s3phys} every model is below $1$, and the regression model is the lowest on all three evaluation sets, consistent with the damped averages discussed in Sec.~\ref{sec:exp}.
\item \textbf{World err} (world-coordinate error): the per-vertex Euclidean error in world coordinates against the ground-truth trajectory, averaged over vertices and frames, reported as the median across scenes (mm). Because deformable dynamics are chaotic, the ground truth may be only one of many possible outcomes, so this number is a reference and not a full measure of prediction quality.
\end{itemize}

\FloatBarrier
\section{Staircase Analysis Details}
\label{app:staircase-loc}

This section diagnoses the staircase artifact. It appears in four baselines of Table~\ref{tab:s2full}: the standard baseline, its two variants (velocity loss and spectral loss), and adaptive compression. The architectures of these baselines are described in Appendix~\ref{app:recon-baselines}.

\subsection{Symptom and diagnosis}
\begin{figure*}[t]
\centering
\includegraphics[width=\textwidth]{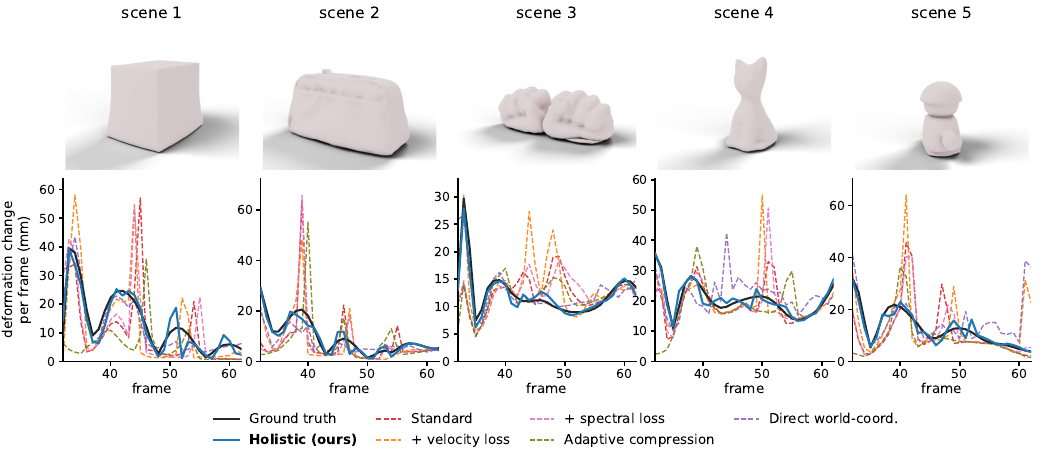}
\caption{The staircase artifact on five GSO scenes. Top: ground truth at the impact frame. Bottom: deformation-change curve (defined in Appendix~\ref{app:metrics-t1}) for ground truth and the six ST-VAEs of Table~\ref{tab:s2full} over the impact window. DCC in Table~\ref{tab:s2main} is the correlation of this curve with the ground-truth curve. On every scene the standard baseline, its two variants, and the adaptive-compression baseline stay below ground truth for several frames and then overshoot it in one frame, while the holistic ST-VAE follows the ground-truth curve. The direct world-coordinate baseline also follows the ground-truth curve, because it uses the same holistic temporal compression.}
\label{fig:panel}
\end{figure*}

\begin{figure}[t]
\centering
\includegraphics[width=\linewidth]{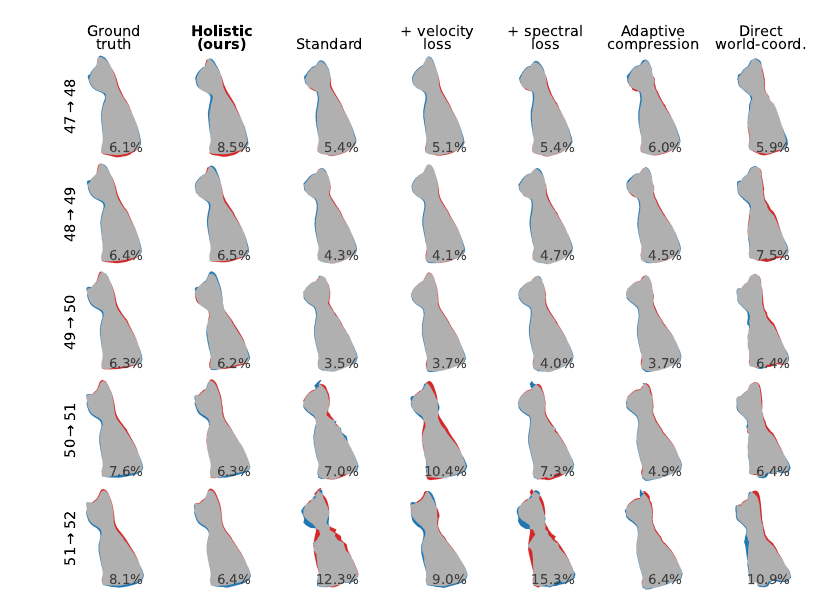}
\caption{Consecutive-frame silhouette difference on scene 4 of Fig.~\ref{fig:panel} (a cat) for five frame pairs around the impact (red: previous frame only; blue: next frame only; grey: overlap; number: XOR over union). Ground truth and the holistic ST-VAE change steadily from frame to frame. The standard baseline and its two variants change much less than the ground truth for the first three pairs and then much more in the last two: the reconstruction holds and then snaps. The adaptive-compression baseline changes too little over all five pairs and releases the motion even later. The direct world-coordinate baseline, which uses the holistic temporal compression, follows the ground truth until the last pair.}
\label{fig:stair_strip}
\end{figure}

\begin{table}[t]\centering\small
\setlength{\tabcolsep}{4pt}
\caption{DCC (deformation-change correlation with ground truth, computed as in Table~\ref{tab:s2main}) on the five GSO scenes of Fig.~\ref{fig:panel}. Scenes 1 to 5 are the five columns of Fig.~\ref{fig:panel} in the same order, and scene 4 is the one shown in Fig.~\ref{fig:stair_strip}. Best per row in bold.}
\label{tab:stair_cases}
\begin{tabular}{@{}lr cccccc@{}}
\toprule
Scene & $N$ & Holistic (ours) & Standard & + vel.\ loss & + spec.\ loss & Adapt.\ compr. & Direct world-coord. \\
\midrule
1 & 5,181 & \textbf{0.972} & 0.728 & 0.850 & 0.781 & 0.418 & \textbf{0.972} \\
2 & 5,913 & \textbf{0.969} & 0.589 & 0.717 & 0.587 & 0.583 & 0.928 \\
3 & 6,005 & \textbf{0.986} & 0.478 & 0.668 & 0.479 & 0.488 & 0.943 \\
4 & 2,245 & \textbf{0.974} & 0.822 & 0.761 & 0.769 & 0.551 & 0.917 \\
5 & 1,526 & \textbf{0.970} & 0.640 & 0.616 & 0.684 & 0.684 & 0.824 \\
\bottomrule
\end{tabular}
\end{table}

The symptom is wrong timing of rapid motion. Where the ground truth changes quickly, for instance at impact, the reconstruction holds its shape for several frames and then jumps to catch up, so the motion arrives late and all at once. We diagnose it in three ways. Fig.~\ref{fig:panel} plots the deformation-change curve on five scenes. The curves of the standard baseline, its two variants, and adaptive compression stay below ground truth for several frames and then overshoot it in one frame. The holistic ST-VAE follows the ground truth, and so does the direct world-coordinate baseline, which uses the same holistic temporal compression. Fig.~\ref{fig:stair_strip} shows the same behavior directly on the silhouettes of one scene. The silhouettes of the standard baseline and its variants change less than the ground truth for three frames and then more than it in the last one or two frames. Adaptive compression changes less than the ground truth over all five frame pairs and releases the motion even later. The direct world-coordinate baseline follows the ground truth until the last pair. Table~\ref{tab:stair_cases} scores the five scenes with DCC. The holistic ST-VAE is the best or tied on every scene.

\subsection{Cause}
In the standard baseline, the decoded frames at the latent-frame positions are accurate, and the frames between them carry the error. The most likely cause is that standard compression summarizes every $k$ frames into one latent frame and loses the high-frequency temporal information that says when, within those $k$ frames, the rapid change happens. One might think that the problem is a mismatch between neighboring blocks. This is not the case. Two observations show that the error builds up inside each block. First, the standard baseline moves too little during the early frames of every block and then catches up in the last frames, whereas the ground truth moves at a steady pace throughout the block. Second, smoothing the reconstruction across block boundaries does not remove the artifact.

\subsection{Why the variants and adaptive compression do not help}
The two variants of the standard baseline and the adaptive-compression baseline all keep the approach of standard temporal compression: they compress the trajectory into a fixed number of latent frames and then reconstruct the frames by local temporal upsampling, so they leave the artifact in place. The velocity loss and the spectral loss change the training signal only. The standard baseline still cannot place the rapid change at the right instant between two latent frames. Adaptive compression moves the latent frames toward the impact, which makes the steps fewer but steeper. A likely reason is that the rebound and settling phases are left with one or two latent frames, so their motion is released in a late snap. The direct world-coordinate baseline uses the holistic temporal compression and shows almost no staircase. Its high DCC in Table~\ref{tab:stair_cases} together with its lower precision in Table~\ref{tab:s2main} shows that its weakness is precision rather than timing.

\FloatBarrier
\section{Chaos in Traditional GPU Simulators}
\label{app:chaos}
This section gives the details of the chaos experiment in Sec.~\ref{sec:chaos} (Chaos analysis). A common assumption is that a physics simulator produces a deterministic trajectory once the initial state and parameters are fully specified. In practice, this is not guaranteed for modern GPU solvers. Because of their speed, GPU solvers are increasingly used to generate physical simulation data at scale, so the consequences of this non-determinism for learning deserve a closer look. Two separate facts are involved. First, a GPU solver is not bit-reproducible: its parallel reductions add floating-point numbers in a different order on every run, which perturbs the state at the level of rounding error. On its own this would be harmless. Second, the complex dynamics of deformable bodies are chaotic: it can amplify such a perturbation into a macroscopic difference in the outcome. The experiment below separates the two.

\subsection{Setup}
We take one held-out scene, a soft tennis racket dropped onto the ground, and run the GPU solver five times on the same GPU from \emph{byte-identical} inputs: the same tetrahedral mesh, initial state, material parameters, and solver configuration. To measure how far two runs drift apart at a given frame, we compute the displacement of every vertex between the two runs and report the largest one as a fraction of the object's extent.

\subsection{The perturbation is tiny}
The non-determinism of the solver is real but minute. After the first simulation step the five runs differ by at most $0.02$\,mm, two hundred-thousandths of the object size. Before the racket reaches the ground, the runs still differ by at most $3\%$ of the object size, mostly in the strings of the net, and their centres of mass by less than $1$\,mm. Until the racket reaches the ground, the five runs are therefore practically identical.

\subsection{Amplification after impact}
\begin{figure}[t]
\centering
\includegraphics[width=\linewidth]{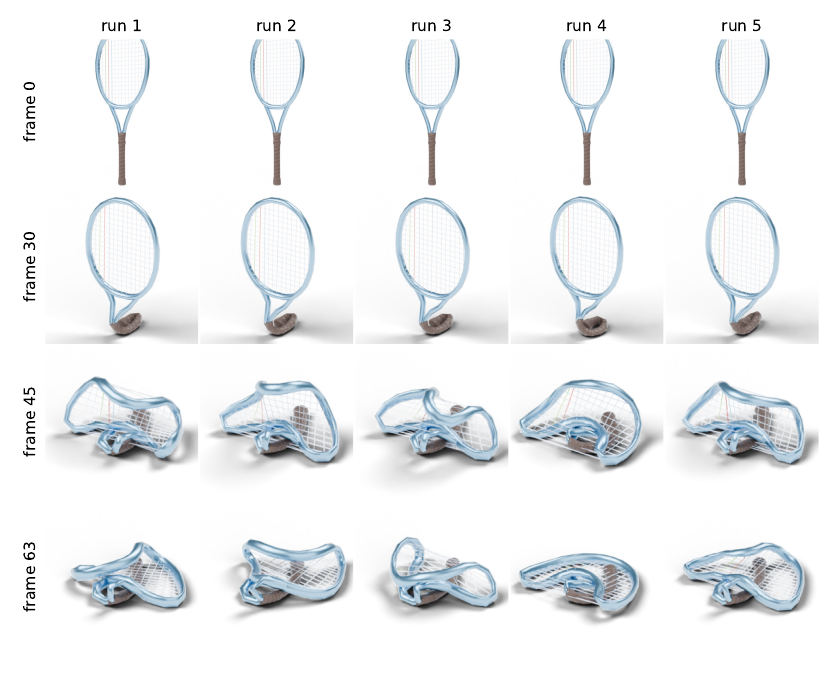}
\caption{Chaos in a traditional GPU simulator. Five reruns (columns) of the \emph{same} soft-racket scene from \emph{byte-identical} inputs (same mesh, initial state, material, and solver configuration), shown at four sampled frames (rows): before the drop, just after impact, mid-bounce, and the last frame of the trajectory. The runs are indistinguishable until impact and then settle into visibly different poses. The complex dynamics are chaotic, so the post-impact outcome is not a deterministic function of the initial state.}
\label{fig:chaos}
\end{figure}

The behavior changes qualitatively at impact. After the racket hits the ground, the runs diverge rapidly and settle into visibly different poses (Fig.~\ref{fig:chaos}). Over the ten pairs of runs, the final divergence ranges from $10\%$ to $51\%$ of the object size, with a median of $21\%$. The same vertex can land up to $442$\,mm apart in two runs.

To check that this is chaos and not a peculiarity of the reduction order, we replace the rounding noise by a controlled perturbation: a deterministic initial rotation of the racket between $10^{-8}$ and $10^{-2}$\,rad. The final divergence is of the same order for every magnitude and does not grow with it. A perturbation a million times larger does not produce a larger difference. This is the signature of chaotic dynamics. The racket has several possible resting poses. Even the smallest perturbation is already enough to change which one it ends in, so making the perturbation larger does not make the difference larger. It only changes which resting pose the racket lands in.

\subsection{Where the chaos comes from}
The sensitivity comes from complex dynamics involving contact. When the racket hits the ground, the solver makes a discrete decision about which edge or corner touches first, and a microscopic perturbation can flip this decision. From then on the bounce and the final pose follow different paths. In our experiments, soft objects that are thin, branchy, or non-convex are more prone to amplifying this chaotic effect.

\subsection{How common is chaos across scenes?}
The racket above is a single scene. To measure how often two runs of the solver diverge, we run the GPU solver a second time from byte-identical inputs on all $300$ Objaverse held-out scenes of Table~\ref{tab:s2main} and compare the two runs. We measure their difference relative to the object size, defined as the diagonal of the bounding box at frame $0$. Table~\ref{tab:resim} counts the scenes whose difference exceeds $1\%$, $5\%$, and $10\%$ of the object size, using either the mean per-vertex error over the whole trajectory or the per-vertex error at the last frame.

Most scenes are reproduced closely. The median difference between the two runs is $0.5$\,mm over the trajectory and $1.0$\,mm at the last frame. A substantial minority of scenes, however, diverge. At the last frame, about a quarter of the scenes end more than $1\%$ of the object size apart, and the most divergent scene ends $35\%$ apart. In all of these scenes, the two runs agree to sub-millimeter precision until ground contact and separate afterwards, as in the racket example. A difference of $1\%$ of the object size means that the two runs end more than $1$\,cm apart on a meter-scale object, which is already a large error for high-precision physical simulation. A regression model can only learn the mean of such diverging outcomes, so high precision is difficult to reach on these scenes.

\begin{table}[!htb]\centering\small
\caption{Divergence between two runs of the GPU solver from byte-identical inputs on the $300$ Objaverse held-out scenes of Table~\ref{tab:s2main}. Each entry gives the number and the percentage of scenes whose difference between the two runs exceeds the given fraction of the object size, measured (a) as the mean per-vertex error over the whole trajectory and (b) as the per-vertex error at the last frame.}
\label{tab:resim}
\begin{tabular}{@{}l ccc@{}}
\toprule
Difference between the two runs & $>1\%$ of size & $>5\%$ of size & $>10\%$ of size \\
\midrule
(a) mean error, whole trajectory & $35$ ($11.7\%$) & $3$ ($1.0\%$) & $2$ ($0.7\%$) \\
(b) final-frame error & $71$ ($23.7\%$) & $23$ ($7.7\%$) & $4$ ($1.3\%$) \\
\bottomrule
\end{tabular}
\end{table}

\subsection{Regression versus diffusion on chaotic and non-chaotic scenes}
\label{app:chaos71}
In a chaotic scene, the same initial state can lead to several different outcomes. A regression model gives one prediction per initial state, while a diffusion model can draw several. This section tests whether, on chaotic scenes, the samples of the LDM actually cover the different outcomes that the solver produces. We split the $300$ held-out scenes of Table~\ref{tab:resim} by how far the two solver runs end apart: $71$ chaotic scenes, whose runs end more than $1\%$ of the object size apart, and $229$ non-chaotic scenes. For every scene we draw eight samples from the LDM and report two numbers: the error of a single sample, and the error of the best of the eight samples. Choosing the best sample needs the ground truth, so the best-of-eight result is an upper bound on sample selection, not a usable predictor. All other settings follow Table~\ref{tab:s3phys}.

\begin{table}[!htb]\centering\small
\caption{Regression versus diffusion on the $229$ non-chaotic and the $71$ chaotic held-out scenes (two solver runs ending less or more than $1\%$ of the object size apart). Medians over scenes. Vol.: volume violation rate (\%), KE: $KE_{\mathrm{def}}/KE_{\mathrm{rig}}$ relative to the ground truth, Err: world-coordinate error (mm) (Appendix~\ref{app:metrics_s3}). Best of $8$ picks, per scene, the sample closest to the ground truth.}
\label{tab:chaos71}
\setlength{\tabcolsep}{3.5pt}
\begin{tabular}{@{}l ccc ccc@{}}
\toprule
 & \multicolumn{3}{c}{Non-chaotic ($229$ scenes)} & \multicolumn{3}{c}{Chaotic ($71$ scenes)} \\
\cmidrule(lr){2-4}\cmidrule(lr){5-7}
 & Vol.\,$\downarrow$ & KE\,$\to$1 & Err\,$\downarrow$ & Vol.\,$\downarrow$ & KE\,$\to$1 & Err\,$\downarrow$ \\
\midrule
Regression & $12.6$ & $0.72$ & $25.0$ & $38.4$ & $0.69$ & $63.2$ \\
Diffusion, single sample & $3.8$ & $0.82$ & $18.0$ & $25.4$ & $0.75$ & $54.0$ \\
Diffusion, best of $8$ & $4.4$ & $0.83$ & $14.6$ & $22.8$ & $0.79$ & $44.7$ \\
\bottomrule
\end{tabular}
\end{table}

\textbf{Diffusion outperforms regression on both subsets.} Table~\ref{tab:chaos71} shows that a single diffusion sample already improves on the regression model in every metric, on both subsets. It has a lower error than the regression model in $79\%$ of the non-chaotic scenes and $65\%$ of the chaotic scenes. The advantage of diffusion is therefore not limited to chaotic scenes.

\textbf{Multiple samples matter more under chaos.} What differs between the two subsets is how much the eight samples add over a single one. Picking the best of the eight lowers the median error by $3.4$\,mm on the non-chaotic scenes and by $9.3$\,mm on the chaotic scenes. The samples also spread further apart under chaos: the gap between the best and the worst sample of a scene is $7$\,mm on the non-chaotic scenes and $25$\,mm on the chaotic scenes (medians over scenes). On the chaotic scenes, this spread follows the physics. It grows with the distance between the two solver runs (Spearman $\rho=0.38$, $p=0.001$), and the second solver run lies inside the spread of the eight samples in $99\%$ of the scenes. The benefit does not depend on picking the best sample: averaged over the eight samples, the median final-frame error on the chaotic scenes is $138$\,mm, below the $162$\,mm of the regression model. The diffusion model is thus more uncertain where the physics is less predictable, and this uncertainty covers outcomes that the solver itself produces.

\begin{figure}[!htb]
\centering
\includegraphics[width=\linewidth]{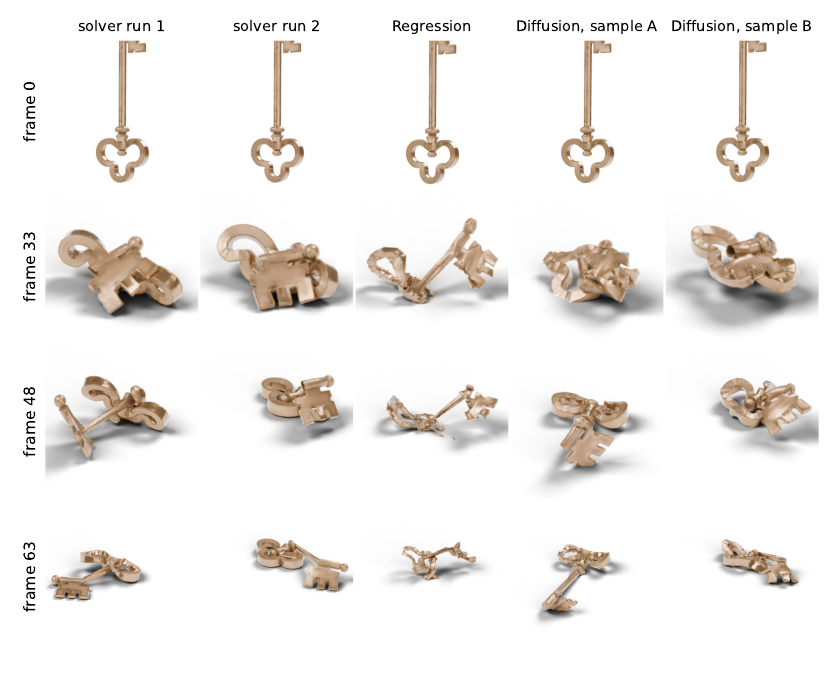}
\caption{The key scene of Appendix~\ref{app:chaos71} (a held-out scene) at four sampled frames (rows; ground contact at frame $31$). Columns: the two runs of the GPU solver from byte-identical inputs, which fall over in different directions; the regression prediction, which settles in between and resembles neither run; and the two diffusion samples closest to run $1$ and to run $2$ among the eight samples.}
\label{fig:keys}
\end{figure}

\textbf{An example: a falling soft key.} In a scene with several possible outcomes, diffusion can learn these outcomes, whereas regression can only learn a non-physical average that matches none of them. Fig.~\ref{fig:keys} shows such a scene, a soft key. When the solver is run twice, the key falls over in two different directions. The regression model predicts a single pose that lies between the two runs and resembles neither. The eight diffusion samples instead fall on both sides, and the closest of them come close to each of the two runs. Other scenes behave similarly. Among the six scenes whose two runs end more than $30^\circ$ apart, at least one diffusion sample ends within $30^\circ$ of a solver run in five scenes, while the regression model does so in only one.

\textbf{An advantage of the paradigm, not of a stronger model.} One may worry that the diffusion model is simply the stronger model (the LDM uses $50$ sampling steps, the regression model a single pass), and that the results above reflect this difference. Three observations show that on chaotic scenes the advantage comes from the paradigm. First, the diffusion samples differ from each other in a way that follows the physics: their spread grows with the distance between the two solver runs and contains the second run in $99\%$ of the chaotic scenes. A deterministic model, however strong, gives a single prediction per initial state and has no such spread. Second, where the solver itself produces clearly different outcomes, the diffusion samples reach them, while the regression prediction is a non-physical average between them, as in the key example. Third, the benefit of drawing several samples depends on the scenes, not on the model. With the same model on both subsets, it is almost three times larger on the chaotic scenes ($9.3$ against $3.4$\,mm), whereas a difference in model strength would affect both subsets alike.

In summary, diffusion predicts more accurately than regression on both chaotic and non-chaotic scenes. On chaotic scenes it also produces several plausible outcomes, while regression gives a single non-physical average result. These results show that diffusion is better suited to chaotic deformable dynamics.

\FloatBarrier
\section{Baselines}
\label{app:baselines}
This section describes the baselines of Table~\ref{tab:s2main} and Table~\ref{tab:s3phys}, and reports the full reconstruction and prediction results (Tables~\ref{tab:s2full} and~\ref{tab:s3full}). The reconstruction baselines are alternative learned representations of the same trajectories, and the prediction baselines are alternative dynamics models started from the same initial state.

\subsection{\texorpdfstring{Reconstruction baselines (Tables~\ref{tab:s2main} and~\ref{tab:s2full})}{Reconstruction baselines}}
\label{app:recon-baselines}
All reconstruction baselines use the $8\times$ temporal compression rate and the $T{=}32$ window of ST-VAE Small, start from the same pretrained spatial VAE (except the direct world-coordinate baseline), and are trained on the same data with comparable budgets.
\begin{itemize}
\item \textbf{Standard}: standard temporal compression of video VAEs. It stacks per-frame latents $\{\mathbf{z}_t\}_{t=1}^{T}$ and shares the factorized spatiotemporal attention of our ST-VAE, but downsamples time by strided convolution to $T_{\mathrm{lat}}=T/k$ latent frames and reconstructs the full frame rate from this time-ordered sequence of latent frames with a local temporal upsampler.
\item \textbf{+ velocity loss}: the strongest variant of the standard baseline we found. It adds a velocity-matching loss on the consecutive-frame differences of the reconstructed deformation and an unfrozen spatial VAE.
\item \textbf{+ spectral loss}: the standard baseline with an additional loss on the deformation-change curve of Appendix~\ref{app:metrics-t1}. The curve is transformed along the time axis with a Fourier transform, and the energy at each frequency is compared with that of the ground-truth curve. Low frequencies correspond to slow changes of the deformation and high frequencies to fast ones, so the loss asks the reconstruction to contain as much fast change as the ground truth instead of smoothing it away. It fails possibly because a frequency spectrum records how much fast change there is but not when it happens, so the standard baseline still cannot place the rapid change at the right instant between two latent frames.
\item \textbf{Adaptive compression}: a separate baseline that keeps the number of latent frames of the standard baseline but places them non-uniformly along time according to the temporal curvature of the ground-truth deformation. Its lower DCC in Table~\ref{tab:s2main} may have the following cause. With only $M{=}4$ latent frames per $32$-frame window, the impact takes most of them and the rebound and settling frames are left with one or two, so their motion is likely reconstructed flat and released in a late snap.
\item \textbf{Direct world-coord.} (direct world-coordinate reconstruction): removes the factoring of Eq.~\eqref{eq:factoring}. Its spatial VAE is trained on world-coordinate frames without canonicalization, and the same holistic temporal compression is applied on top without a rigid token, so rigid motion and deformation share one latent.
\end{itemize}

\subsection{\texorpdfstring{Full reconstruction results (Table~\ref{tab:s2full})}{Full reconstruction results}}
\label{app:recon-full}
Table~\ref{tab:s2full} reports the full ST-VAE reconstruction comparison with all six ST-VAEs, including the two variants of the standard baseline and the adaptive-compression baseline, which are omitted from Table~\ref{tab:s2main}.

\begin{table}[!htb]\centering\footnotesize
\setlength{\tabcolsep}{1.8pt}\renewcommand{\arraystretch}{0.85}\setlength{\aboverulesep}{0.15ex}\setlength{\belowrulesep}{0.25ex}
\caption{Full version of Table~\ref{tab:s2main} with all six ST-VAEs, including the two variants of the standard baseline and the adaptive-compression baseline. ST-VAE reconstruction comparison across the three evaluation sets (row blocks: held-out in-distribution / GSO zero-shot / Toys4K zero-shot). Left six columns: scenes within the training vertex range ($500\le N\le5{,}000$; $300$ scenes per set). Right three columns: large-mesh subsets beyond the $5{,}000$-vertex training cap ($N>5{,}000$; $300$ scenes per set), reported on mean deformation and rigid errors only. DCC is the deformation-change correlation with ground truth and directly reflects the staircase artifact (higher is better); Recon is the reconstruction error of the deformation field, i.e.\ of the geometry with the rigid motion removed (mm); rot and trans are the reconstruction errors of the rigid rotation (deg) and translation (mm). Full definitions in Appendix~\ref{app:metrics}. \tightbox{tableBest}{\textbf{best}} and \underline{second-best} per column within each block. $^\dagger$Direct world-coordinate baseline (no factoring). Its errors are computed after removing the rigid motion from its output.}
\label{tab:s2full}
\begin{tabular*}{\linewidth}{@{\extracolsep{\fill}}c l cccccc ccc@{}}
\toprule
 & & \multicolumn{6}{c}{$500\le N\le 5{,}000$} & \multicolumn{3}{c}{\makebox[0pt][c]{$5{,}000<N\le10{,}000$}} \\
\cmidrule(lr){3-8}\cmidrule(lr){9-11}
 & & DCC$\uparrow$ & \multicolumn{3}{c}{Recon$\downarrow$} & \multicolumn{2}{c}{Rigid$\downarrow$} & Recon$\downarrow$ & \multicolumn{2}{c}{Rigid$\downarrow$} \\
\cmidrule(lr){3-3}\cmidrule(lr){4-6}\cmidrule(lr){7-8}\cmidrule(lr){9-9}\cmidrule(lr){10-11}
 & ST-VAE & mean & mean & med & P90 & rot & trans & mean & rot & trans \\
\midrule
\multirow{6}{*}{\rotatebox[origin=c]{90}{\shortstack{Objaverse\\Held-out}}} & Standard & 0.665 & 5.32 & 3.47 & 12.01 & \bestcell{0.270} & \secondcell{0.57} & 6.14 & \secondcell{0.185} & \secondcell{0.50} \\
 & \;+ velocity loss & 0.842 & 5.34 & \secondcell{3.15} & 11.60 & 0.831 & 3.33 & 6.25 & 0.638 & 3.88 \\
 & \;+ spectral loss & 0.697 & \secondcell{5.17} & 3.42 & \secondcell{11.40} & 0.290 & 0.73 & \secondcell{5.96} & 0.210 & 0.66 \\
 & Adaptive compression & 0.511 & 5.86 & 3.65 & 13.72 & 0.440 & 5.09 & 6.59 & 0.314 & 4.30 \\
 & Direct world-coord.$^\dagger$ & \secondcell{0.880} & 6.50 & 4.21 & 15.60 & 0.951 & 3.89 & 7.99 & 0.846 & 4.14 \\
 & \textbf{Holistic (ours)} & \bestcell{0.949} & \bestcell{2.48} & \bestcell{1.62} & \bestcell{8.08} & \secondcell{0.276} & \bestcell{0.53} & \bestcell{4.71} & \bestcell{0.169} & \bestcell{0.47} \\
\midrule
\multirow{6}{*}{\rotatebox[origin=c]{90}{GSO}} & Standard & 0.641 & 3.46 & 1.64 & 8.83 & \secondcell{0.161} & \secondcell{0.48} & 4.30 & \secondcell{0.149} & \secondcell{0.42} \\
 & \;+ velocity loss & 0.834 & 3.51 & 1.77 & \secondcell{8.25} & 0.572 & 3.49 & 4.51 & 0.536 & 3.07 \\
 & \;+ spectral loss & 0.672 & \secondcell{3.39} & \secondcell{1.63} & 8.52 & 0.181 & 0.62 & \secondcell{4.13} & 0.176 & 0.57 \\
 & Adaptive compression & 0.482 & 4.01 & 1.83 & 10.63 & 0.262 & 5.37 & 4.73 & 0.252 & 5.08 \\
 & Direct world-coord.$^\dagger$ & \secondcell{0.836} & 5.56 & 2.98 & 13.35 & 0.640 & 3.65 & 5.17 & 0.678 & 3.12 \\
 & \textbf{Holistic (ours)} & \bestcell{0.941} & \bestcell{1.91} & \bestcell{1.02} & \bestcell{5.88} & \bestcell{0.158} & \bestcell{0.45} & \bestcell{3.39} & \bestcell{0.147} & \bestcell{0.37} \\
\midrule
\multirow{6}{*}{\rotatebox[origin=c]{90}{Toys4K}} & Standard & 0.663 & 5.49 & 3.83 & 11.11 & \secondcell{0.310} & \secondcell{0.62} & 6.86 & \secondcell{0.189} & \secondcell{0.51} \\
 & \;+ velocity loss & 0.836 & 5.81 & 3.95 & 11.34 & 0.927 & 4.65 & 7.18 & 0.686 & 3.78 \\
 & \;+ spectral loss & 0.691 & \secondcell{5.47} & \secondcell{3.82} & \secondcell{11.03} & 0.328 & 0.80 & \secondcell{6.70} & 0.213 & 0.68 \\
 & Adaptive compression & 0.504 & 6.03 & 4.27 & 12.53 & 0.501 & 5.14 & 7.48 & 0.328 & 5.92 \\
 & Direct world-coord.$^\dagger$ & \secondcell{0.858} & 6.96 & \secondcell{3.82} & 17.05 & 1.191 & 4.47 & 8.16 & 0.901 & 4.55 \\
 & \textbf{Holistic (ours)} & \bestcell{0.930} & \bestcell{3.12} & \bestcell{2.05} & \bestcell{8.29} & \bestcell{0.301} & \bestcell{0.61} & \bestcell{5.44} & \bestcell{0.174} & \bestcell{0.49} \\
\bottomrule
\end{tabular*}
\end{table}

\subsection{\texorpdfstring{Prediction baselines (Table~\ref{tab:s3phys})}{Prediction baselines}}
\label{app:predbaselines}
The prediction baselines receive the same initial state as PhysLDM.
\begin{itemize}
\item \textbf{Regression}: a deterministic regression model that is capacity-matched to the LDM and shares its ST-VAE and conditioning. It is a transformer that emits the full latent trajectory in one pass and is trained with the token-weighted mean-squared error of Appendix~\ref{app:arch}, so that only the predictive paradigm differs between the two.
\item \textbf{MGN-style AR} (MeshGraphNets-style autoregressive model): a graph network \citep{DBLP:conf/iclr/PfaffFSB21} over the tetrahedral mesh graph that predicts one frame at a time from the previous frames, with the noise-injected one-step training standard in that family \citep{DBLP:conf/icml/Sanchez-Gonzalez20}, retrained on our scenes. Predicting world coordinates step by step, it accumulates error over time and drifts far from the ground truth (Table~\ref{tab:s3phys}). It is compared with our model under a fair setting.
\end{itemize}

\subsection{\texorpdfstring{Full prediction results (Table~\ref{tab:s3full})}{Full prediction results}}
\label{app:pred-full}
Table~\ref{tab:s3full} reports the prediction comparison with all six physics checks, including the velocity and acceleration violation rates that are omitted from Table~\ref{tab:s3phys}, together with the ST-VAE reconstruction of the ground truth as a reference row, which gives the autoencoder's own contribution to each metric.

\begin{table}[!htb]\centering\footnotesize
\setlength{\tabcolsep}{2pt}\renewcommand{\arraystretch}{0.85}\setlength{\aboverulesep}{0.15ex}\setlength{\belowrulesep}{0.25ex}
\caption{Full version of Table~\ref{tab:s3phys} with all six physics checks and the ST-VAE reconstruction reference row. Prediction comparison across the three evaluation sets, on the same $300$ scenes per set within the training vertex range ($500\le N\le5{,}000$) as Table~\ref{tab:s2main}. Viol.\ \%: percentage of vertices or tetrahedra, over all frames, whose predicted volume, speed, or acceleration deviates from the ground truth beyond a threshold; Self-int.\ is the fraction of surface triangles in a self-intersection; $KE_{\mathrm{def}}/KE_{\mathrm{rig}}$ is the deformation-to-rigid kinetic-energy ratio relative to ground truth (closer to $1$ is better); World err is the per-vertex error in world coordinates (mm); full definitions in Appendix~\ref{app:metrics}. \emph{ST-VAE reconstruction} is the ground truth encoded and decoded by the ST-VAE behind the LDM, shown as a reference. \tightbox{tableBest}{\textbf{best}} and \underline{second-best} among the three models per column within each block. MGN-style AR is a MeshGraphNets-style (MGN) autoregressive graph network retrained on our scenes.}
\label{tab:s3full}
\begin{tabular*}{\linewidth}{@{\extracolsep{\fill}}c l cccccc@{}}
\toprule
 &  & Volume & Self-int. & Vel. & Acc. & $\frac{KE_{\mathrm{def}}}{KE_{\mathrm{rig}}}$ & World err \\
 & Prediction Model & viol.\ \%$\downarrow$ & tris \%$\downarrow$ & viol.\ \%$\downarrow$ & viol.\ \%$\downarrow$ & $\times$GT$\rightarrow$1 & median (mm)$\downarrow$ \\
\midrule
\multirow{4}{*}{\rotatebox[origin=c]{90}{\shortstack{Objaverse\\Held-out}}} & \emph{ST-VAE reconstruction} & 8.5 & 0.12 & 0.8 & 16.9 & 0.91 & 1.97 \\
 & \textbf{Diffusion (ours)} & \bestcell{7.9} & \bestcell{0.24} & \bestcell{7.0} & \bestcell{25.8} & \bestcell{0.80} & \bestcell{28.2} \\
 & Regression & \secondcell{22.7} & \secondcell{0.79} & \secondcell{8.6} & \secondcell{29.2} & 0.72 & \secondcell{37.5} \\
 & MGN-style AR & 55.1 & 3.53 & 21.4 & 69.0 & \secondcell{0.77} & 80.6 \\
\midrule
\multirow{4}{*}{\rotatebox[origin=c]{90}{GSO}} & \emph{ST-VAE reconstruction} & 2.1 & 0.00 & 0.5 & 11.8 & 0.92 & 1.21 \\
 & \textbf{Diffusion (ours)} & \bestcell{2.0} & \bestcell{0.00} & \bestcell{3.8} & \bestcell{17.4} & \bestcell{0.86} & \bestcell{14.0} \\
 & Regression & \secondcell{5.1} & \secondcell{0.01} & \secondcell{6.4} & \secondcell{22.7} & 0.72 & \secondcell{19.9} \\
 & MGN-style AR & 47.1 & 0.68 & 20.9 & 66.6 & \secondcell{0.76} & 78.4 \\
\midrule
\multirow{4}{*}{\rotatebox[origin=c]{90}{Toys4K}} & \emph{ST-VAE reconstruction} & 10.6 & 0.19 & 1.1 & 17.6 & 0.91 & 2.15 \\
 & \textbf{Diffusion (ours)} & \bestcell{10.3} & \bestcell{0.32} & \bestcell{8.6} & \bestcell{28.5} & \bestcell{0.80} & \bestcell{35.6} \\
 & Regression & \secondcell{24.3} & \secondcell{1.07} & \secondcell{9.0} & \secondcell{29.6} & 0.69 & \secondcell{42.6} \\
 & MGN-style AR & 57.5 & 4.73 & 20.2 & 72.2 & \secondcell{0.77} & 79.1 \\
\bottomrule
\end{tabular*}
\end{table}

\FloatBarrier
\section{More Ablations}
\label{app:moreabl}
This section collects the ablations that support design choices of the model: the use of the ST-VAE, the dedicated rigid pathway, the spatial-VAE pretraining, and the decoder GNN.

\subsection{The ST-VAE}
This section gives the measurements behind the ST-VAE ablation of Sec.~\ref{sec:ablations}, which asks what happens to the training cost when the ST-VAE is removed and the LDM denoises the raw vertex$\times$frame tokens directly. We compare two options on the same GPU at increasing vertex counts $N$. The first is our LDM architecture applied to the raw tokens, without any latent. The second is our pipeline with the ST-VAE, where the LDM works on a fixed set of $4{,}104$ latent tokens and the ST-VAE is the stage with the largest memory footprint. Table~\ref{tab:latent} lists the peak training memory of both.

Without the ST-VAE the number of tokens is $T\times N$, so memory grows with $N$. The raw-token model fits at batch~$1$ up to $2{,}500$ vertices and runs out of memory at $2{,}600$, below the median resolution of our dataset (${\sim}3{,}400$ vertices). More than half of the training scenes therefore cannot be processed at all. With the ST-VAE, the LDM costs about $4.4$\,GiB per trajectory at any $N$. The ST-VAE stage grows linearly with $N$, by about $8$\,GiB per $1{,}000$ vertices, and ST-VAE Base trains up to about $14{,}000$ vertices on the same card. The latent therefore does not remove the vertex-dependent cost, but moves it to a stage that scales far more gently.

\begin{table}[t]\centering\small
\setlength{\tabcolsep}{6pt}
\caption{Training memory versus vertex count $N$ on a $140$\,GiB H200 ($T{=}64$, \texttt{bf16}, batch~$1$). No ST-VAE: our LDM architecture applied directly to the $T{\times}N$ raw vertex tokens, peak memory (GiB) of one training step. With ST-VAE: peak memory (GiB) of one ST-VAE Base training step. The LDM stage with the ST-VAE costs about $4.4$\,GiB per trajectory at any $N$ and is not listed. OOM: out of memory.}
\label{tab:latent}
\begin{tabular}{@{}rcc@{}}
\toprule
 & no ST-VAE & with ST-VAE \\
$N$ & LDM on raw tokens (GiB) & ST-VAE Base step (GiB) \\
\midrule
500 & 28.9 & 33.5 \\
1{,}000 & 55.4 & 37.4 \\
2{,}000 & 108.3 & 45.8 \\
2{,}500 & 134.5 & 50.7 \\
2{,}600 & OOM & 50.7 \\
3{,}500 & OOM & 56.0 \\
5{,}000 & OOM & 70.4 \\
7{,}500 & OOM & 92.9 \\
10{,}000 & OOM & 116.0 \\
14{,}000 & OOM & 135.3 \\
14{,}500 & OOM & OOM \\
\bottomrule
\end{tabular}
\end{table}

\subsection{The rigid pathway}
Sec.~\ref{sec:ablations} shows that routing the rigid token through the shared holistic temporal compression makes the translation error explode and starves the deformation track: on a small held-out diagnostic set, the translation error grows from $1.8$ to $887$\,mm, and the reconstructed deformation shrinks to $28\%$ of its true magnitude ($97\%$ with the dedicated pathway). In addition, the failure is not a matter of numerical scale. Normalizing the rigid signal to unit variance before the shared compression barely helps ($52.8\!\rightarrow\!49.6$\,mm world error), whereas the dedicated linear pathway reaches $8.3$\,mm. The rigid and the deformation signals compete for the capacity of the shared latent, and the dedicated pathway removes this competition.

\subsection{Spatial VAE pretraining}
This section explains why the spatial VAE needs to be pretrained first rather than trained directly together with the ST-VAE. We compare two training schemes at a matched budget. The first pretrains the spatial VAE and then trains the ST-VAE on top of it with an additional latent-anchor loss (Sec.~\ref{sec:training}), which ties the per-frame latents reconstructed by the ST-VAE to those of the pretrained spatial VAE. The second trains both stages jointly from scratch, with a randomly initialized spatial VAE.

We find that joint training works poorly. At the standard learning rate, training diverges within a few thousand steps. At a reduced learning rate, it converges to a much worse model: the geometric reconstruction error is about $5\times$ that of the pretrained scheme, the deformation is much weaker than the ground truth, and volume is lost. The likely reason is that the two stages chase each other. The latent that the ST-VAE learns to reproduce keeps moving while the spatial VAE is still changing, so the temporal model has no fixed target. Pretraining fixes the target first. Once pretrained, the spatial VAE can either be kept frozen during ST-VAE training or fine-tuned with a smaller learning rate, and the two options give similar results. We keep it frozen for ST-VAE Small and fine-tune it for ST-VAE Base.

\subsection{The spatial decoder GNN}
We use a GNN component in the spatial VAE decoder because the field decoder places every vertex independently, so nothing ties neighboring vertices together, and the GNN over the tetrahedral mesh adds this coupling. We compare three placements of the GNN: in the decoder, as in our model; removed; and moved to the encoder instead. Without the GNN, reconstruction degrades and many mesh edges end up stretched far from their rest length. With the GNN in the encoder, the result is worse on every metric, because smoothing the mesh before encoding removes detail before it reaches the latent, whereas smoothing after decoding only regularizes the output.

\FloatBarrier
\section{Inference Runtime}
\label{app:runtime}

This section reports the inference speed of PhysLDM and how it depends on the mesh resolution. We measure batch-1 inference on a single H200 for $64$-frame trajectories on five held-out scenes with $980$ to $9{,}851$ vertices, the last three beyond the $5{,}000$-vertex training cap. One inference pass consists of three steps: the spatial VAE encodes frame $0$, the latent model predicts the latent trajectory (one forward pass for the regression model, $50$ or $20$ Euler steps for the LDM), and the ST-VAE decodes all $64$ frames. Timings are medians over repeated runs with the GPU synchronized and no I/O in the timed region, in \texttt{bf16} and in \texttt{fp32}. Table~\ref{tab:runtime} lists the decoding time, the resulting frame rates, and the peak GPU memory.

Two observations follow. First, the cost of the latent model is independent of the mesh resolution, because it always works on the same $4{,}104$ latent tokens, so its latency is the same at every $N$ (\texttt{bf16}: $9.8$\,ms for the regression model, $460$\,ms for the LDM at $20$ steps, and $1{,}150$\,ms at $50$). The decoder is the only stage whose cost grows with $N$, and in \texttt{bf16} it is nearly flat, from $318$\,ms at $980$ vertices to $384$\,ms at $9{,}851$, with no loss of accuracy. As a result, the LDM runs at about $40$ to $80$ frames per second in \texttt{bf16} and the regression model at up to $195$, even on meshes twice the training cap. Second, the whole pass needs only about $5$\,GiB of GPU memory across the range.

\begin{table}[t]\centering\small
\setlength{\tabcolsep}{4pt}
\caption{Batch-1 inference of PhysLDM on one H200 for $64$-frame trajectories at different vertex counts $N$ (held-out scenes; the last three rows exceed the $5{,}000$-vertex training cap). Decode: time of the ST-VAE decoder. fps $=64/(\text{condition build}+\text{latent model}+\text{decode})$. Peak: maximum GPU memory allocated over the whole pass.}
\label{tab:runtime}
\begin{tabular}{@{}r cc ccc ccc c@{}}
\toprule
 & \multicolumn{2}{c}{Decode (ms)} & \multicolumn{3}{c}{\texttt{bf16} fps} & \multicolumn{3}{c}{\texttt{fp32} fps} & Peak \\
\cmidrule(lr){2-3}\cmidrule(lr){4-6}\cmidrule(lr){7-9}
$N$ & \texttt{bf16} & \texttt{fp32} & Regression & LDM-20 & LDM-50 & Regression & LDM-20 & LDM-50 & (GiB) \\
\midrule
980 & 318 & 288 & 195 & 82 & 44 & 192 & 30 & 13 & 4.8 \\
2,493 & 327 & 425 & 189 & 79 & 42 & 136 & 28 & 13 & 4.8 \\
5,007 & 328 & 597 & 188 & 81 & 43 & 100 & 26 & 12 & 4.8 \\
7,534 & 332 & 778 & 186 & 81 & 43 & 78 & 25 & 12 & 5.2 \\
9,851 & 384 & 960 & 162 & 76 & 42 & 64 & 23 & 12 & 5.4 \\
\bottomrule
\end{tabular}
\end{table}

\FloatBarrier
\section{Additional Qualitative Results}
\label{app:qual}
This section extends the qualitative figures of the main text with more scenes, lists the parameters of every scene shown, and presents the extreme case of Appendix~\ref{app:extreme}.

\subsection{More reconstruction and prediction results}
\begin{figure*}[!t]
\centering
\setlength{\tabcolsep}{2pt}\renewcommand{\arraystretch}{0.6}
\resizebox{\textwidth}{!}{\begin{tabular}{@{}c@{\,}c@{\hspace{5pt}}c@{}}
\rotatebox{90}{\scriptsize GT} & \makebox[0pt][l]{\includegraphics[height=\qualh]{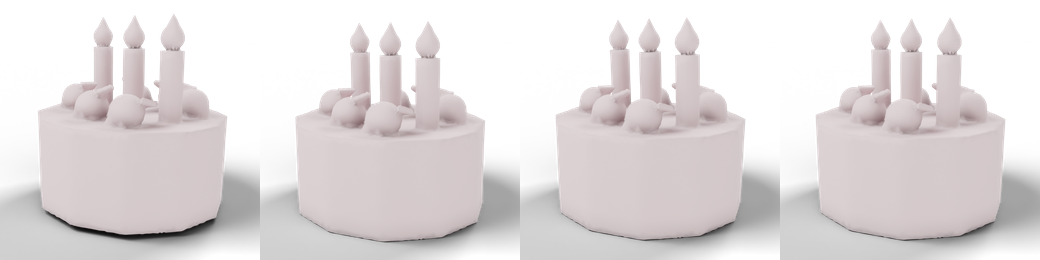}}\makebox[0pt][l]{\raisebox{1.5pt}{\hspace{1.5pt}\scriptsize $N{=}6640$}}\hspace{4.0000\qualh} & \makebox[0pt][l]{\includegraphics[height=\qualh]{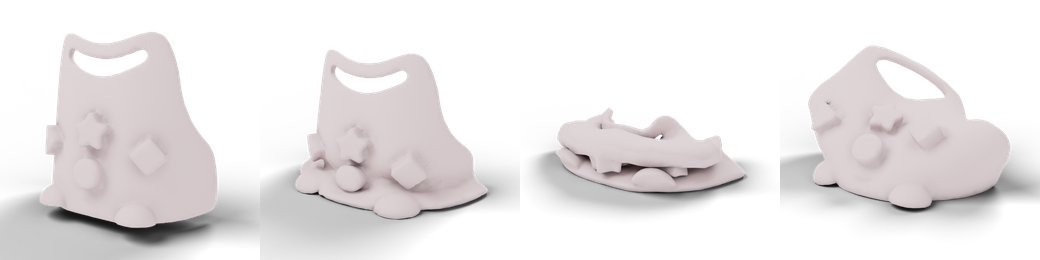}}\makebox[0pt][l]{\raisebox{1.5pt}{\hspace{1.5pt}\scriptsize $N{=}6426$}}\hspace{4.0000\qualh} \\
\rotatebox{90}{\scriptsize\shortstack{\textbf{Holistic}\\\textbf{(ours)}}} & \includegraphics[height=\qualh]{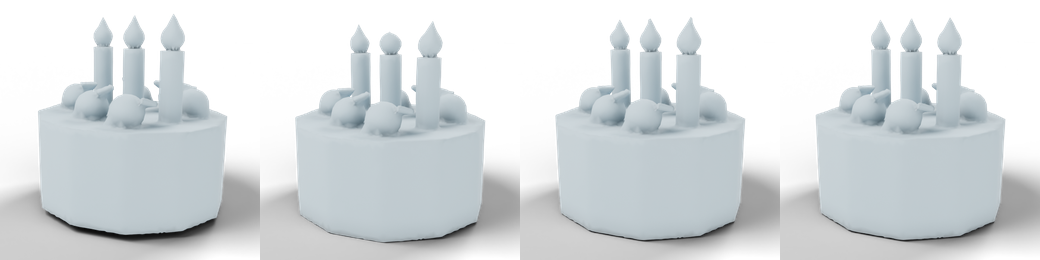} & \includegraphics[height=\qualh]{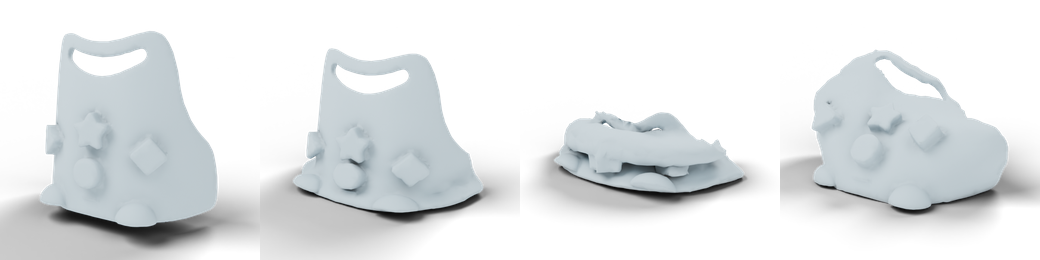} \\
\rotatebox{90}{\scriptsize Standard} & \includegraphics[height=\qualh]{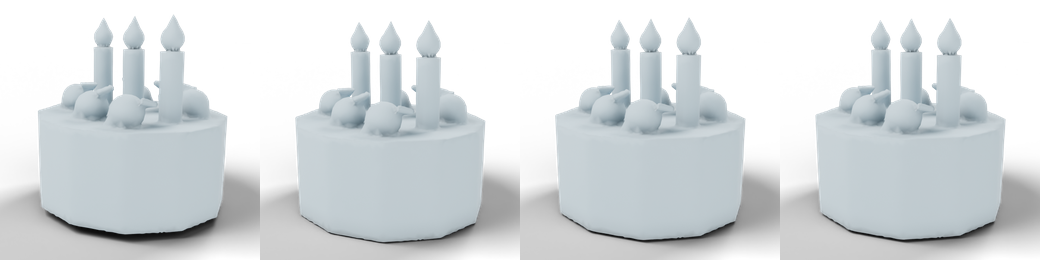} & \includegraphics[height=\qualh]{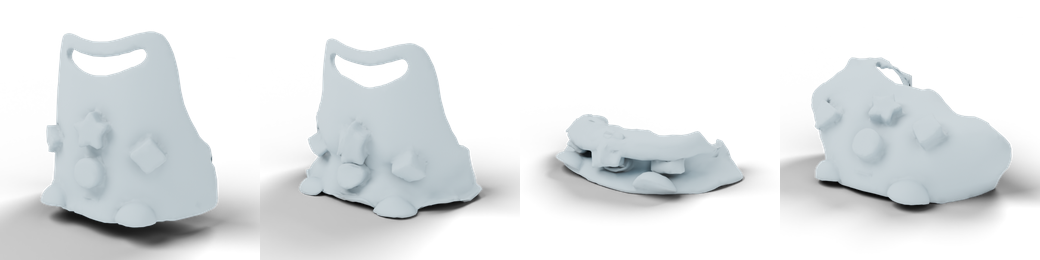} \\
\rotatebox{90}{\scriptsize\shortstack{+ velocity\\loss}} & \includegraphics[height=\qualh]{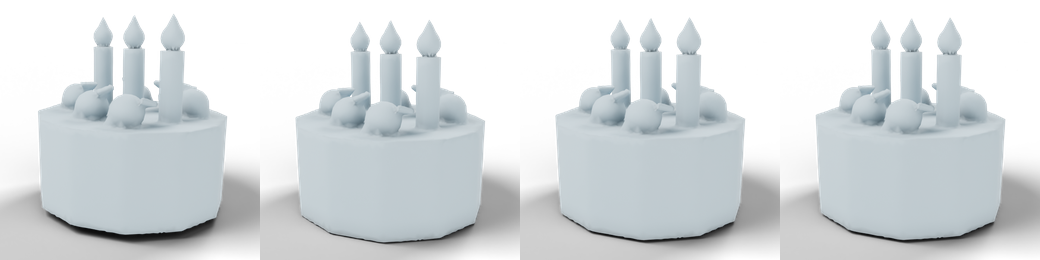} & \includegraphics[height=\qualh]{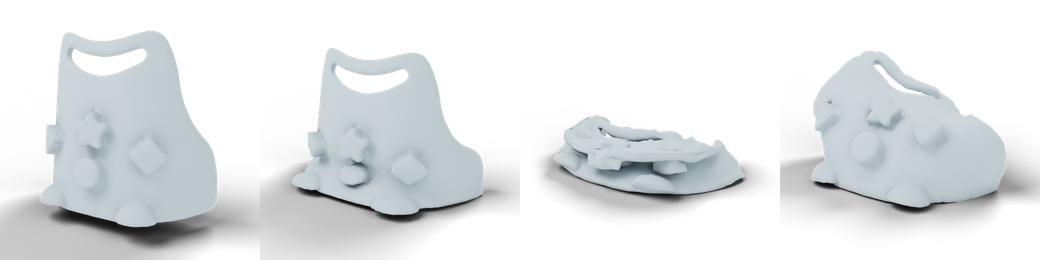} \\
\rotatebox{90}{\scriptsize\shortstack{+ spectral\\loss}} & \includegraphics[height=\qualh]{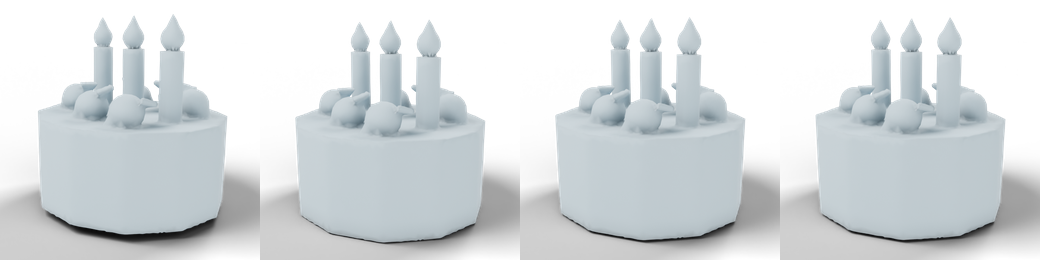} & \includegraphics[height=\qualh]{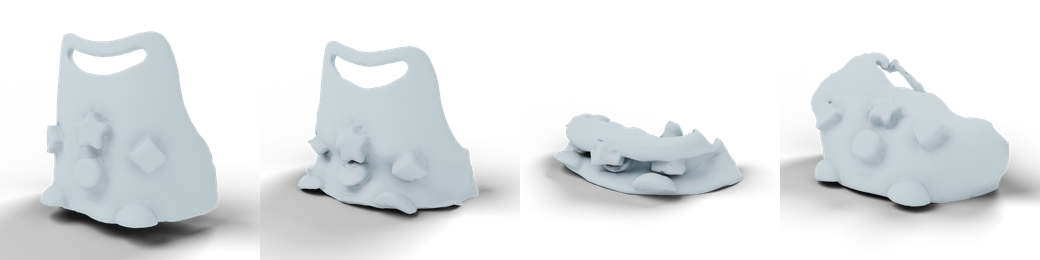} \\
\rotatebox{90}{\scriptsize\shortstack{Adaptive\\compression}} & \includegraphics[height=\qualh]{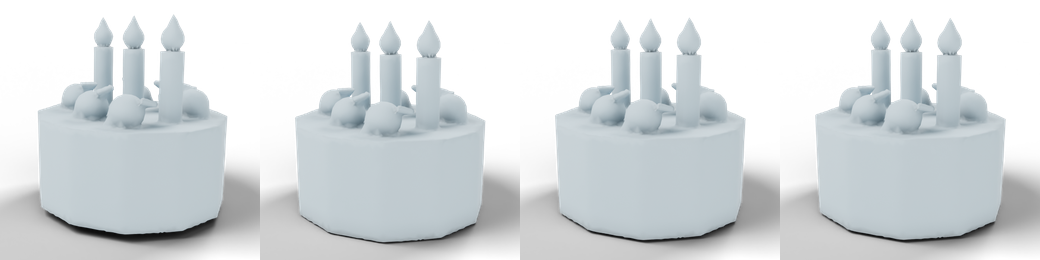} & \includegraphics[height=\qualh]{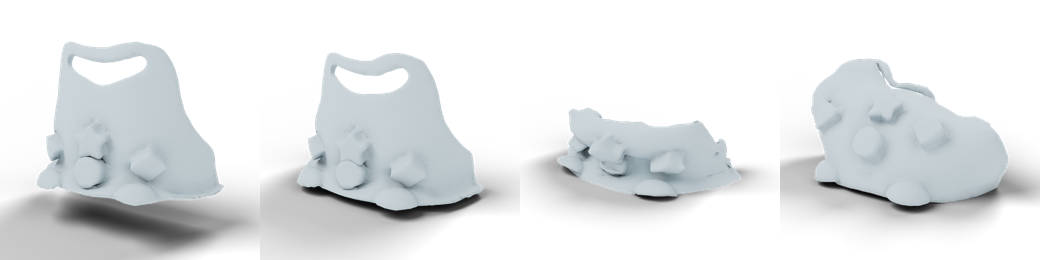} \\
\rotatebox{90}{\scriptsize\shortstack{Direct\\world-coord.}} & \makebox[0pt][l]{\includegraphics[height=\qualh]{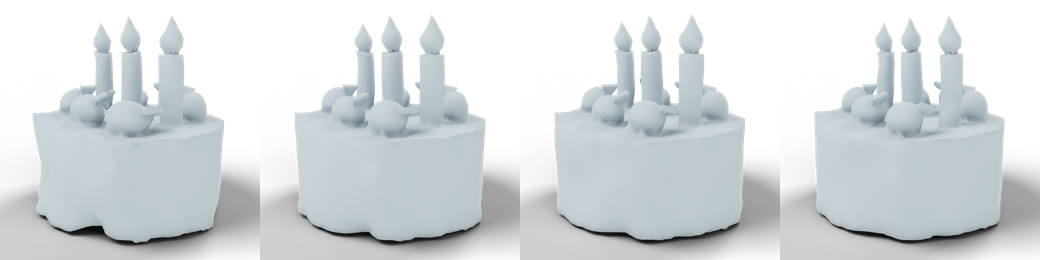}}\makebox[0pt][l]{\hspace{3pt}\tikz[baseline=0pt]{\node[inner sep=0pt,inner xsep=3pt,font=\small\itshape,text=black!65] (t) at ($(2.0000\qualh,0)+(-1.5pt,0)$) {time};\draw[black!65,line width=0.8pt,line cap=round] (0,0) -- (t.west);\draw[black!65,line width=0.8pt,line cap=round,-{Straight Barb[length=5.5pt,width=6.5pt,angle'=60]}] (t.east) -- ($(4.0000\qualh,0)+(-3pt,0)$);}}\hspace{4.0000\qualh} & \includegraphics[height=\qualh]{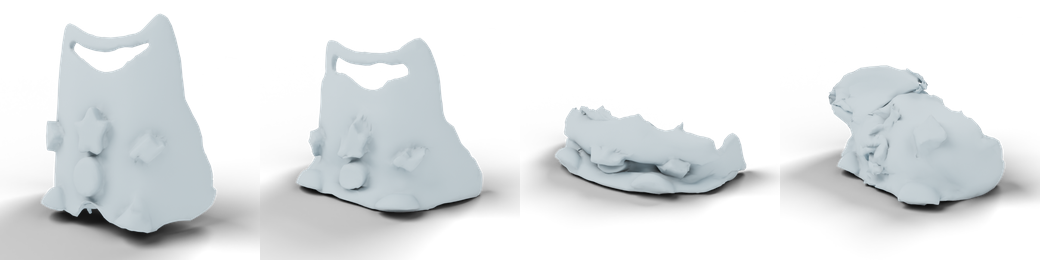} \\
\end{tabular}}

\caption{ST-VAE reconstruction on two more scenes (a Toys4K cake and a GSO shape-sorter toy) at four frames sampled along the trajectory. Rows: ground truth (rose), \textbf{Holistic (ours)}, the standard baseline, its two variants (velocity loss and spectral loss), the adaptive-compression baseline, and the direct world-coordinate baseline. Scene details are listed in Table~\ref{tab:qual_recon_cases}.}
\label{fig:qual_recon_app}
\end{figure*}

\begin{figure*}[!t]
\centering
\setlength{\tabcolsep}{2pt}\renewcommand{\arraystretch}{0.6}
\resizebox{\textwidth}{!}{\begin{tabular}{@{}c@{\,}c@{\hspace{5pt}}c@{}}
\rotatebox{90}{\scriptsize GT} & \makebox[0pt][l]{\includegraphics[height=\qualh]{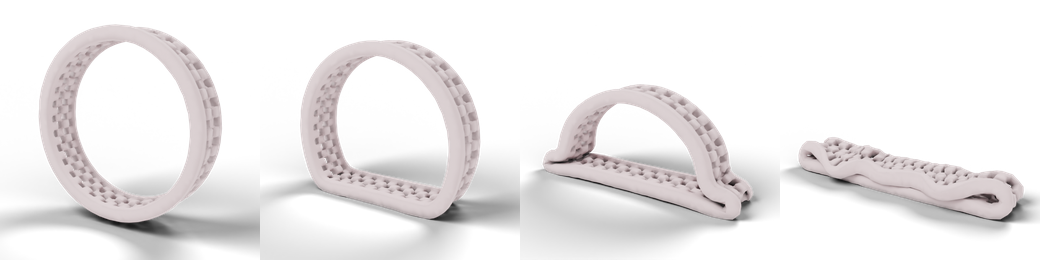}}\makebox[0pt][l]{\raisebox{1.5pt}{\hspace{1.5pt}\scriptsize $N{=}9772$}}\hspace{4.0000\qualh} & \makebox[0pt][l]{\includegraphics[height=\qualh]{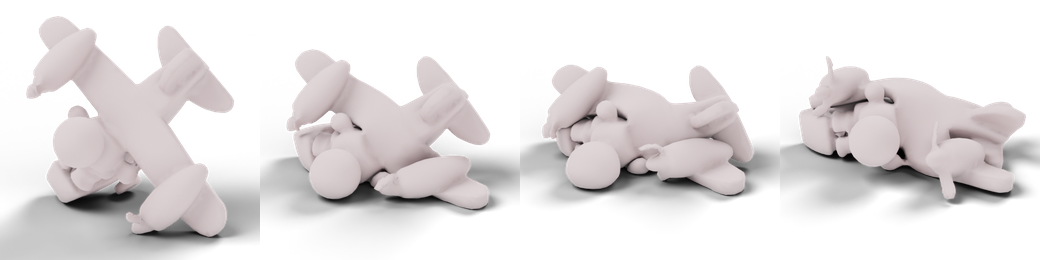}}\makebox[0pt][l]{\raisebox{1.5pt}{\hspace{1.5pt}\scriptsize $N{=}6891$}}\hspace{4.0000\qualh} \\
\rotatebox{90}{\scriptsize\shortstack{\textbf{Holistic}\\\textbf{(ours)}}} & \includegraphics[height=\qualh]{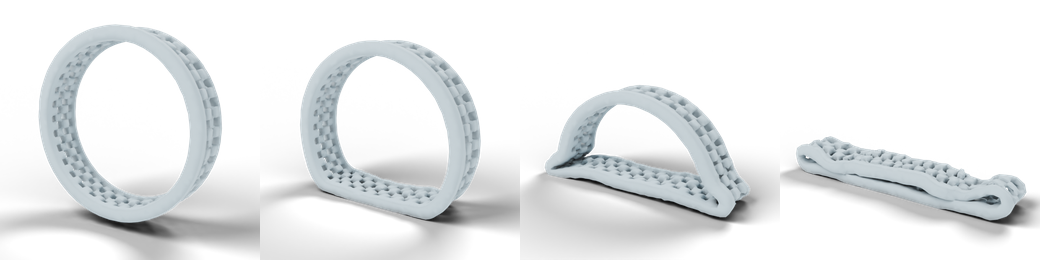} & \includegraphics[height=\qualh]{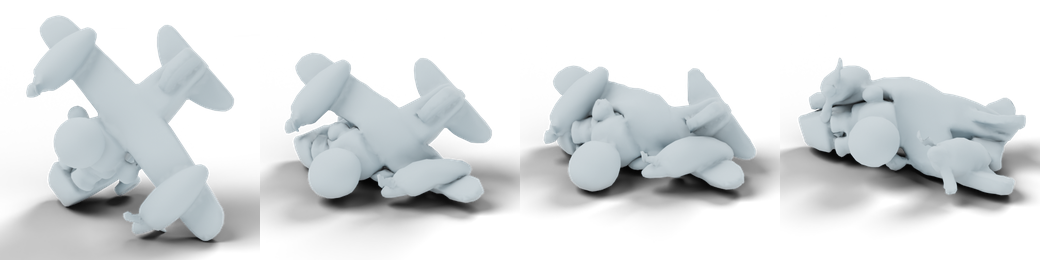} \\
\rotatebox{90}{\scriptsize Standard} & \includegraphics[height=\qualh]{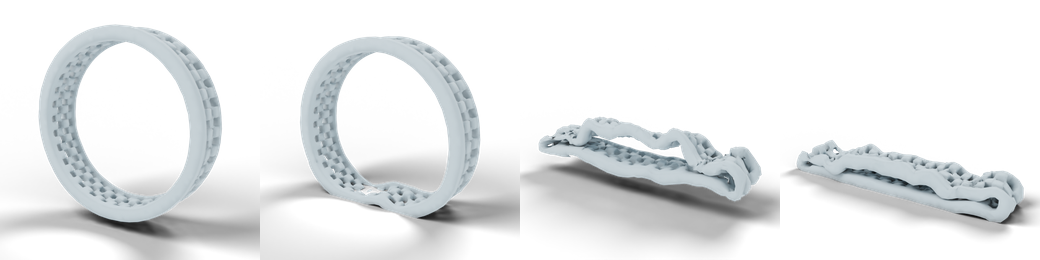} & \includegraphics[height=\qualh]{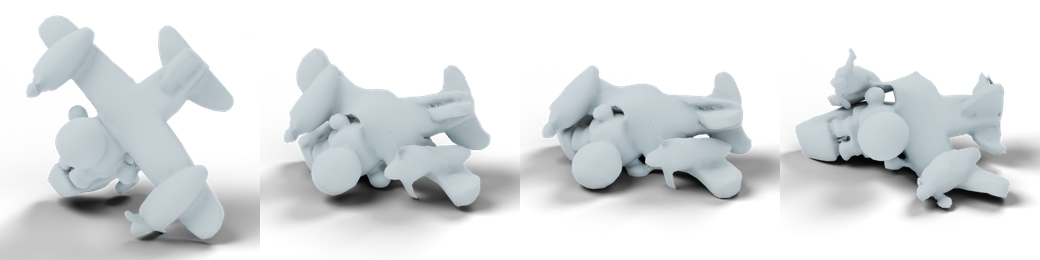} \\
\rotatebox{90}{\scriptsize\shortstack{+ velocity\\loss}} & \includegraphics[height=\qualh]{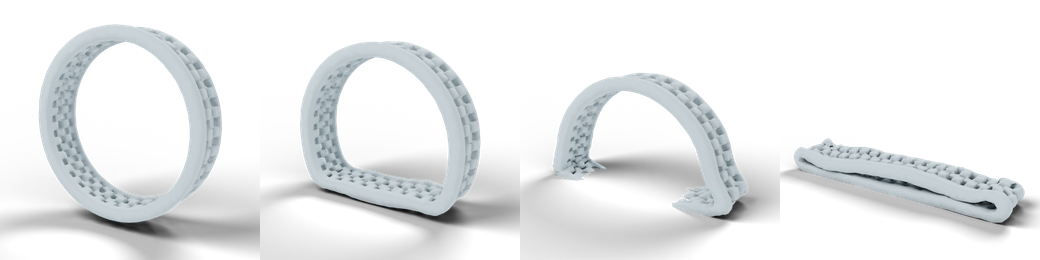} & \includegraphics[height=\qualh]{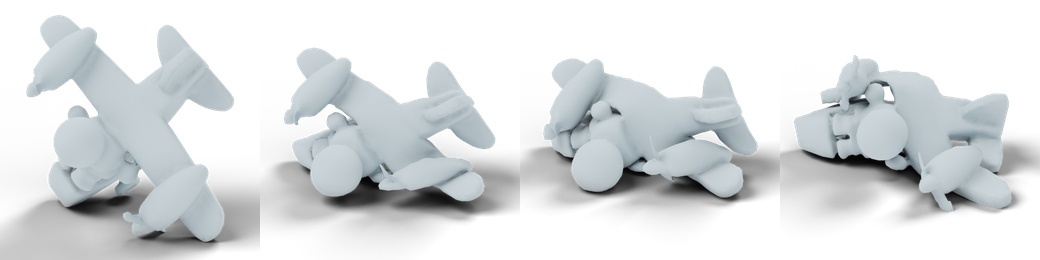} \\
\rotatebox{90}{\scriptsize\shortstack{+ spectral\\loss}} & \includegraphics[height=\qualh]{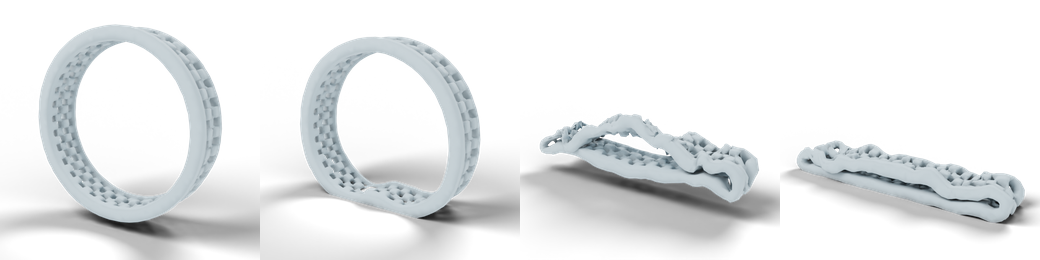} & \includegraphics[height=\qualh]{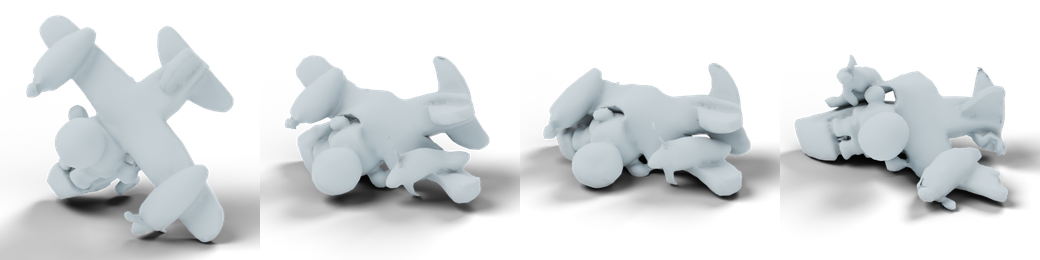} \\
\rotatebox{90}{\scriptsize\shortstack{Adaptive\\compression}} & \includegraphics[height=\qualh]{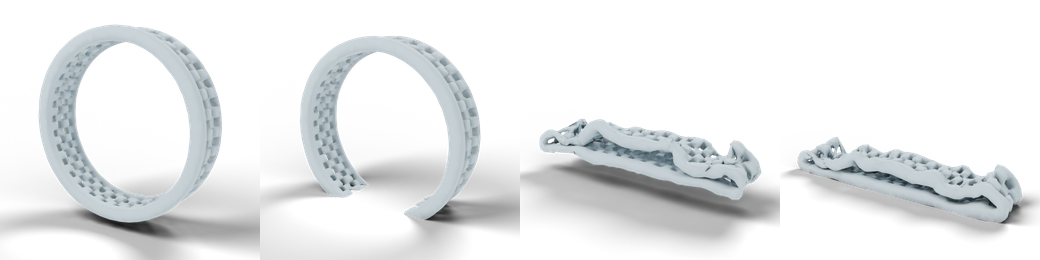} & \includegraphics[height=\qualh]{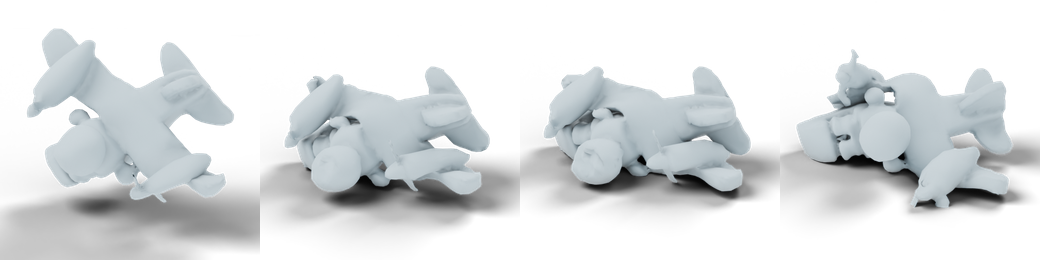} \\
\rotatebox{90}{\scriptsize\shortstack{Direct\\world-coord.}} & \includegraphics[height=\qualh]{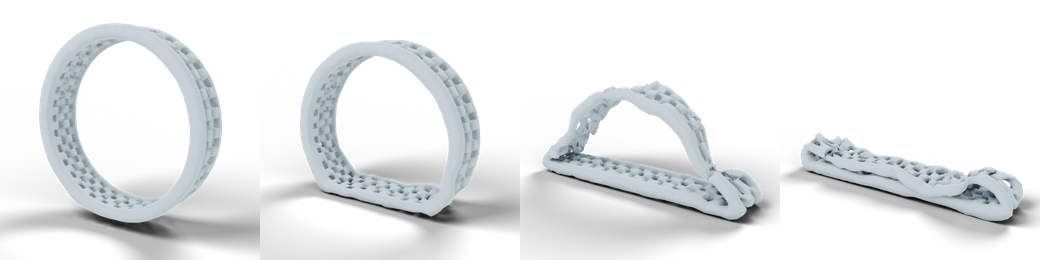} & \includegraphics[height=\qualh]{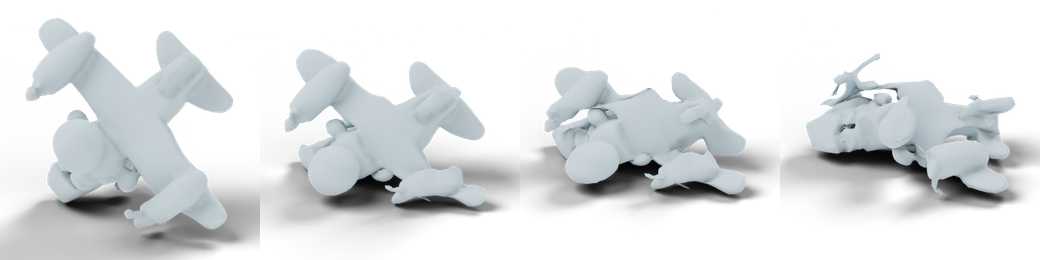} \\
\end{tabular}}

\caption{ST-VAE reconstruction on two more scenes (a held-out Objaverse object and a GSO turboprop), with the same rows, sampled frames, and rendering as Fig.~\ref{fig:qual_recon_app}. Scene details are listed in Table~\ref{tab:qual_recon_cases}.}
\label{fig:qual_recon_app2}
\end{figure*}

Figs.~\ref{fig:qual_recon_app}, \ref{fig:qual_recon_app2}, and~\ref{fig:qual_pred_app} extend Figs.~\ref{fig:qual_recon} and~\ref{fig:qual_pred} with further scenes, rendered and laid out in the same way. Figs.~\ref{fig:qual_recon_app} and~\ref{fig:qual_recon_app2} show ST-VAE reconstruction on four more scenes against every ST-VAE baseline of Table~\ref{tab:s2full}. Fig.~\ref{fig:qual_pred_app} shows our prediction on four more scenes against the prediction baselines of Table~\ref{tab:s3phys}. The observations of the main text hold on these scenes as well. The results of the standard baseline, its variants, and adaptive compression look reasonable in these static frames, but their DCC values (Table~\ref{tab:s2full}) show that they have the staircase artifact. The direct world-coordinate baseline produces noisy geometric detail. The regression model produces wrong physics and jitter, and the MGN-style AR baseline drifts over time. Tables~\ref{tab:qual_recon_cases} and~\ref{tab:qual_pred_cases} list the source dataset, node count, and simulation parameters of every scene in these figures and in the main-text figures.

The bottom-left block of Fig.~\ref{fig:teaser} shows one-shot prediction on a larger set of objects in one scene. Twenty-six objects from GSO, Toys4K, and the held-out Objaverse scenes, with $1{,}517$ to $9{,}821$ nodes, are thrown onto a round stage with initial speeds between $0.57$ and $1.87$\,m/s and Young's moduli between $3\times10^{4}$ and $3\times10^{6}$\,Pa, so that soft plush toys, a rubber boot, and a bunch of grapes share the scene with a helmet, a drill, and a cardboard box. The inset shows the initial state: every object in its undeformed rest shape, placed above the position where it will land, seen from the same camera as the main view. The main view shows the predicted state near the end of the sequence, when almost all objects have landed, rendered with the original textures of the assets. Each object is predicted independently as a $64$-frame trajectory from its initial state, and the soft objects deform visibly on landing while the stiff ones keep their shape.

\begin{table}[!t]
\centering\small
\caption{Scenes of Figs.~\ref{fig:qual_recon}, \ref{fig:qual_recon_app}, and~\ref{fig:qual_recon_app2}: source dataset, tetrahedral node count $N$, Young's modulus $E$, and initial velocity $v_0$ (m/s, $y$ up). The Young's moduli were chosen to show both soft bodies with large deformation and stiff, nearly rigid ones, so some values do not match the real material of the object.}
\label{tab:qual_recon_cases}
\setlength{\tabcolsep}{4pt}
\begin{tabular*}{\linewidth}{@{\extracolsep{\fill}}llllll@{}}
\toprule
Figure & Object & Dataset & $N$ & $E$ (Pa) & $v_0$ (m/s) \\
\midrule
Fig.~\ref{fig:qual_recon} & phone & Objaverse (held-out) & 6,106 & $2.76\times10^{6}$ & $(+0.59,\,+0.00,\,+0.94)$, $|v_0|=1.11$ \\
Fig.~\ref{fig:qual_recon} & boot & GSO & 6,569 & $5\times10^{4}$ & $0$ (released from rest) \\
Fig.~\ref{fig:qual_recon} & chess piece & Toys4K & 6,322 & $5\times10^{4}$ & $0$ (released from rest) \\
Fig.~\ref{fig:qual_recon} & toy & GSO & 9,403 & $5\times10^{4}$ & $0$ (released from rest) \\
Fig.~\ref{fig:qual_recon_app} & cake & Toys4K & 6,640 & $1.45\times10^{6}$ & $(-0.21,\,+0.00,\,+0.58)$, $|v_0|=0.62$ \\
Fig.~\ref{fig:qual_recon_app} & shape-sorter toy & GSO & 6,426 & $5\times10^{4}$ & $0$ (released from rest) \\
Fig.~\ref{fig:qual_recon_app2} & object & Objaverse (held-out) & 9,772 & $5\times10^{4}$ & $0$ (released from rest) \\
Fig.~\ref{fig:qual_recon_app2} & turboprop & GSO & 6,891 & $5\times10^{4}$ & $0$ (released from rest) \\
\bottomrule
\end{tabular*}
\end{table}

\begin{figure*}[!t]
\centering
\setlength{\tabcolsep}{2pt}\renewcommand{\arraystretch}{0.6}
\resizebox{\textwidth}{!}{\begin{tabular}{@{}c@{\,}c@{\hspace{5pt}}c@{}}
\rotatebox{90}{\scriptsize GT} & \makebox[0pt][l]{\includegraphics[height=\qualh]{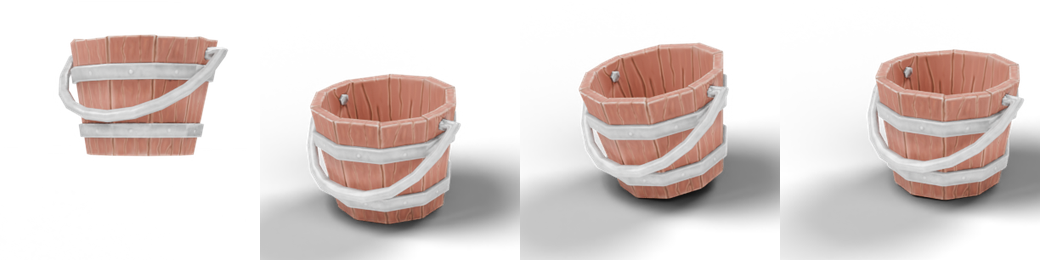}}\makebox[0pt][l]{\raisebox{1.5pt}{\hspace{1.5pt}\scriptsize $N{=}9853$}}\hspace{4.0000\qualh} & \makebox[0pt][l]{\includegraphics[height=\qualh]{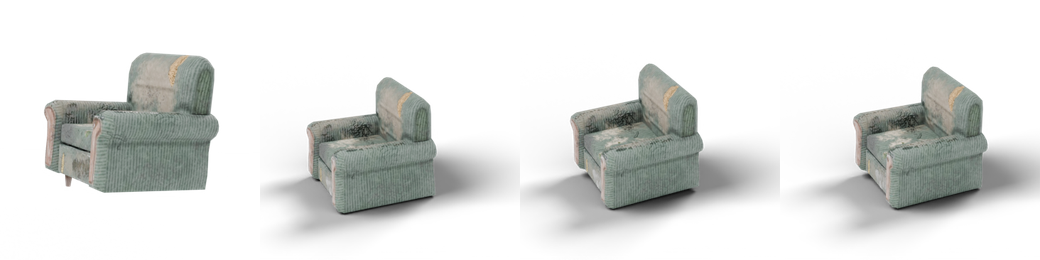}}\makebox[0pt][l]{\raisebox{1.5pt}{\hspace{1.5pt}\scriptsize $N{=}4228$}}\hspace{4.0000\qualh} \\
\rotatebox{90}{\scriptsize\shortstack{\textbf{Diffusion}\\\textbf{(ours)}}} & \includegraphics[height=\qualh]{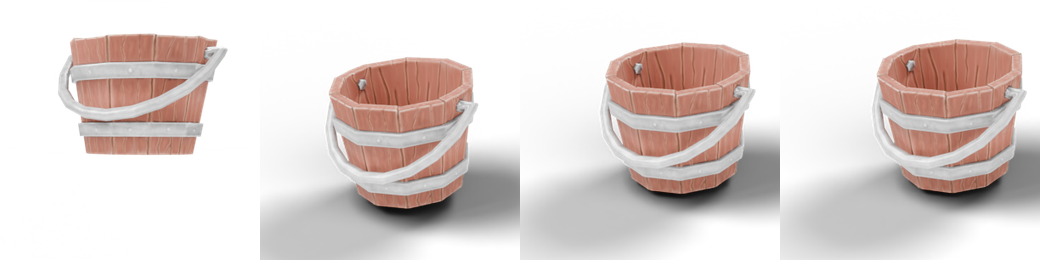} & \includegraphics[height=\qualh]{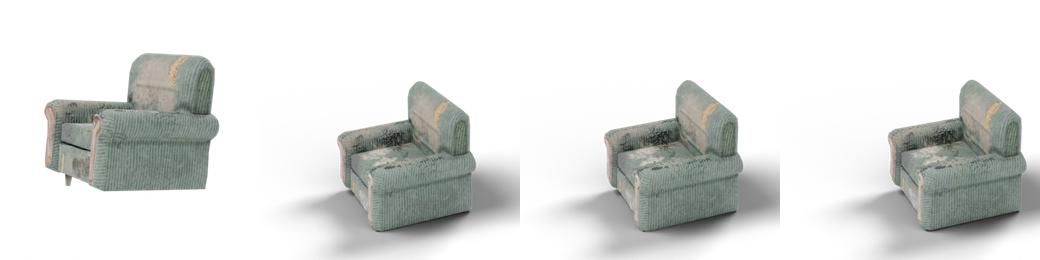} \\
\rotatebox{90}{\scriptsize Regression} & \includegraphics[height=\qualh]{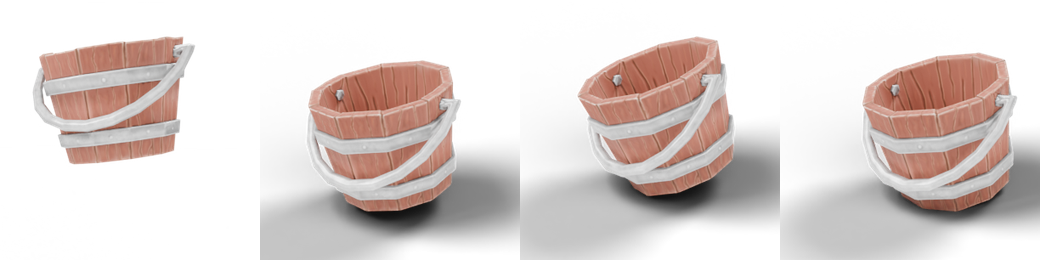} & \includegraphics[height=\qualh]{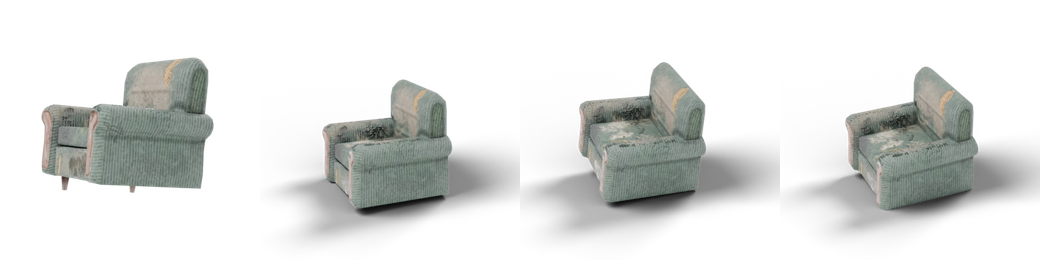} \\
\rotatebox{90}{\scriptsize\shortstack{MGN-style\\AR}} & \makebox[0pt][l]{\includegraphics[height=\qualh]{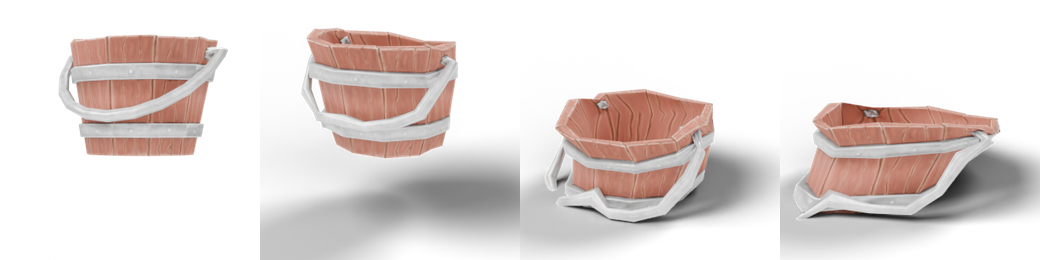}}\makebox[0pt][l]{\hspace{3pt}\tikz[baseline=0pt]{\node[inner sep=0pt,inner xsep=3pt,font=\small\itshape,text=black!65] (t) at ($(2.0000\qualh,0)+(-1.5pt,0)$) {time};\draw[black!65,line width=0.8pt,line cap=round] (0,0) -- (t.west);\draw[black!65,line width=0.8pt,line cap=round,-{Straight Barb[length=5.5pt,width=6.5pt,angle'=60]}] (t.east) -- ($(4.0000\qualh,0)+(-3pt,0)$);}}\hspace{4.0000\qualh} & \includegraphics[height=\qualh]{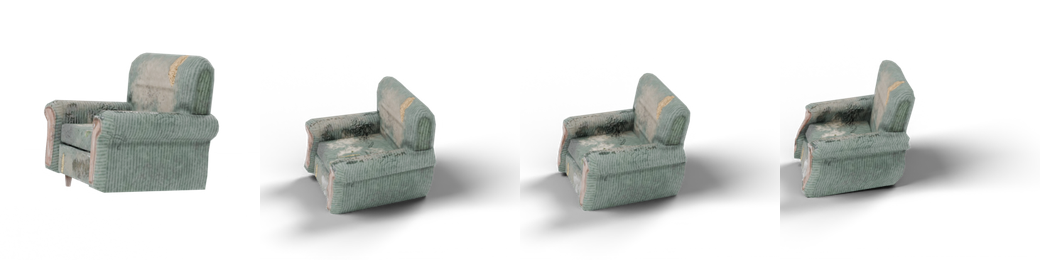} \\
\noalign{\vskip 4pt}
\rotatebox{90}{\scriptsize GT} & \makebox[0pt][l]{\includegraphics[height=\qualh]{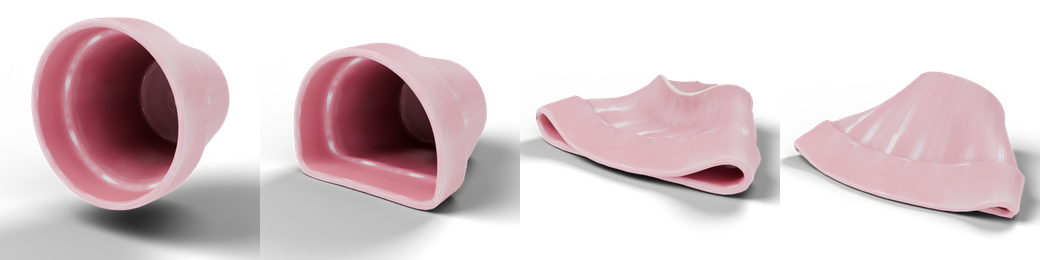}}\makebox[0pt][l]{\raisebox{1.5pt}{\hspace{1.5pt}\scriptsize $N{=}3487$}}\hspace{4.0000\qualh} & \makebox[0pt][l]{\includegraphics[height=\qualh]{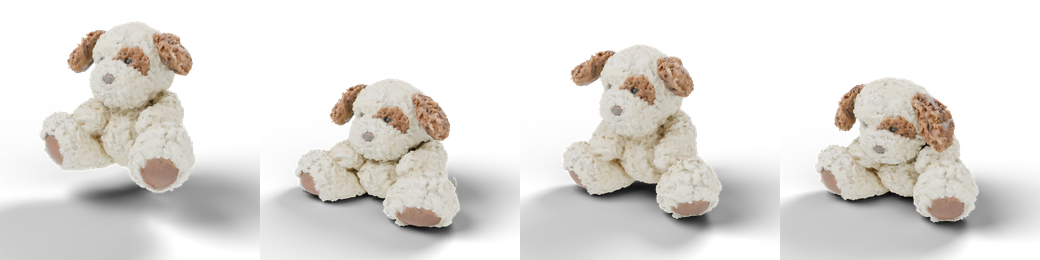}}\makebox[0pt][l]{\raisebox{1.5pt}{\hspace{1.5pt}\scriptsize $N{=}8496$}}\hspace{4.0000\qualh} \\
\rotatebox{90}{\scriptsize\shortstack{\textbf{Diffusion}\\\textbf{(ours)}}} & \includegraphics[height=\qualh]{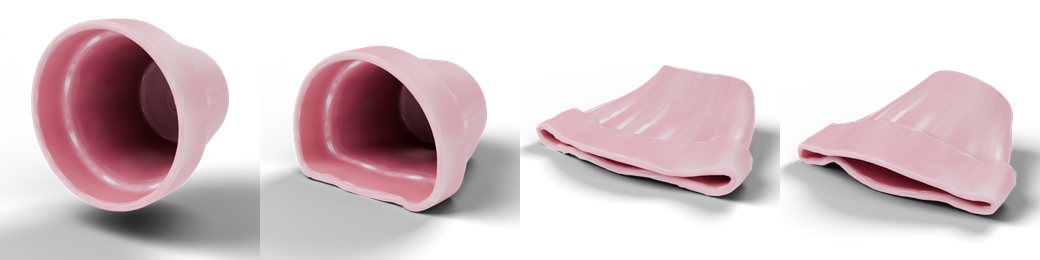} & \includegraphics[height=\qualh]{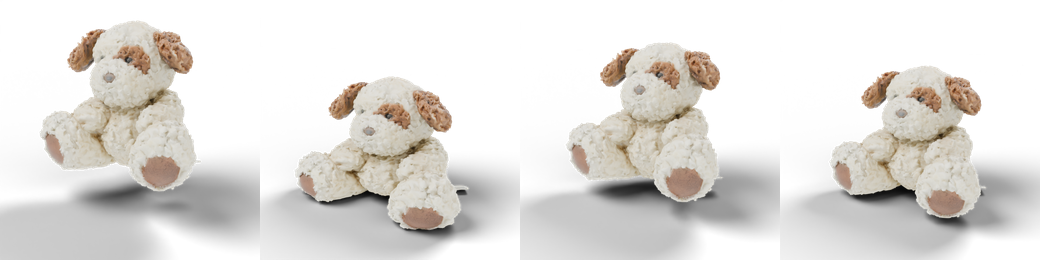} \\
\rotatebox{90}{\scriptsize Regression} & \includegraphics[height=\qualh]{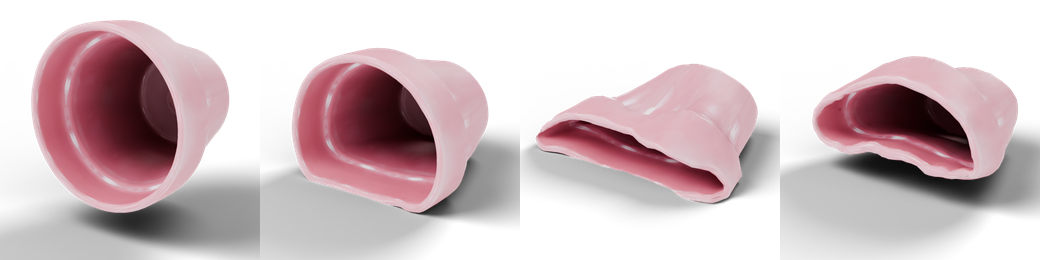} & \includegraphics[height=\qualh]{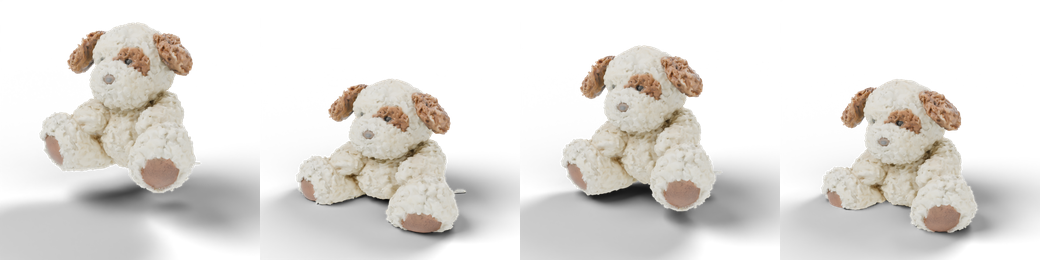} \\
\rotatebox{90}{\scriptsize\shortstack{MGN-style\\AR}} & \includegraphics[height=\qualh]{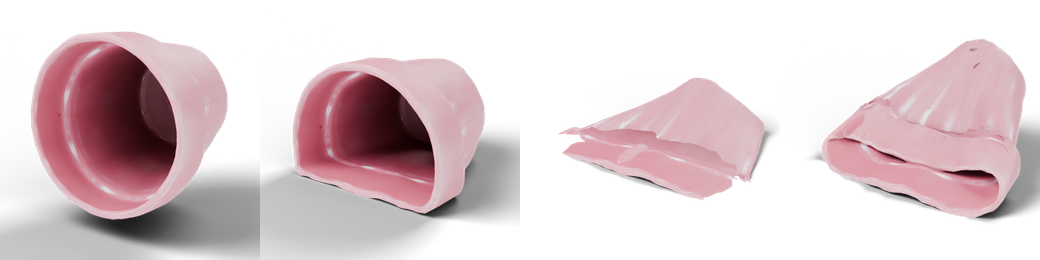} & \includegraphics[height=\qualh]{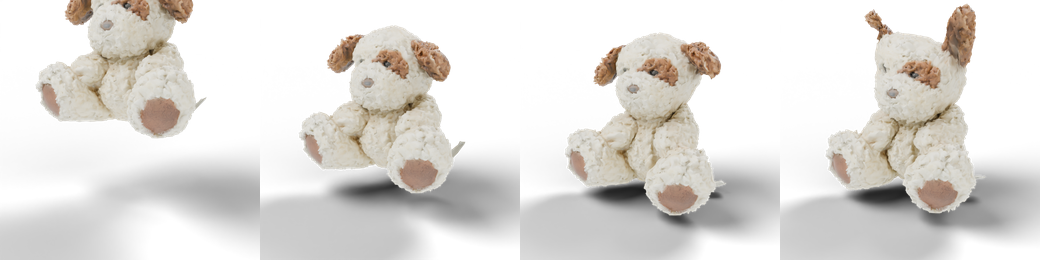} \\
\end{tabular}}

\caption{Prediction on four more scenes (a held-out Objaverse bucket and sofa, a GSO flower pot, and a GSO plush dog) at four frames sampled along the trajectory. Rows: ground truth, \textbf{Diffusion (ours)}, Regression, and the MGN-style AR baseline. Scene details are listed in Table~\ref{tab:qual_pred_cases}.}
\label{fig:qual_pred_app}
\end{figure*}

\begin{table}[!t]
\centering\small
\caption{Scenes of Figs.~\ref{fig:qual_pred} and~\ref{fig:qual_pred_app}: source dataset, tetrahedral node count $N$, Young's modulus $E$, and initial velocity $v_0$ (m/s, $y$ up). The Young's moduli were chosen to show both soft bodies with large deformation and stiff, nearly rigid ones, so some values do not match the real material of the object.}
\label{tab:qual_pred_cases}
\setlength{\tabcolsep}{4pt}
\begin{tabular*}{\linewidth}{@{\extracolsep{\fill}}llllll@{}}
\toprule
Figure & Object & Dataset & $N$ & $E$ (Pa) & $v_0$ (m/s) \\
\midrule
Fig.~\ref{fig:qual_pred} & punch-drop toy & GSO & 7,767 & $2\times10^{6}$ & $(-0.48,\,+0.00,\,-1.78)$, $|v_0|=1.85$ \\
Fig.~\ref{fig:qual_pred} & cap & Objaverse (held-out) & 2,435 & $5\times10^{4}$ & $0$ (released from rest) \\
Fig.~\ref{fig:qual_pred} & bowl & Objaverse (held-out) & 5,877 & $5\times10^{4}$ & $0$ (released from rest) \\
Fig.~\ref{fig:qual_pred} & grapes & Toys4K & 8,767 & $5\times10^{4}$ & $0$ (released from rest) \\
Fig.~\ref{fig:qual_pred_app} & bucket & Objaverse (held-out) & 9,853 & $2.6\times10^{6}$ & $(-0.94,\,+0.00,\,+1.33)$, $|v_0|=1.63$ \\
Fig.~\ref{fig:qual_pred_app} & sofa & Objaverse (held-out) & 4,228 & $1.05\times10^{6}$ & $(-1.34,\,+0.00,\,-0.20)$, $|v_0|=1.36$ \\
Fig.~\ref{fig:qual_pred_app} & flower pot & GSO & 3,487 & $5\times10^{4}$ & $0$ (released from rest) \\
Fig.~\ref{fig:qual_pred_app} & plush dog & GSO & 8,496 & $5\times10^{4}$ & $0$ (released from rest) \\
\bottomrule
\end{tabular*}
\end{table}

\FloatBarrier
\section{Extreme Cases Analysis}
\label{app:extreme}
This section analyzes extreme cases. The accuracy of the LDM is fundamentally bounded by the reconstruction performance of the ST-VAE, meaning that the most challenging cases for the ST-VAE set the limits of the entire pipeline. Across the held-out scenes, the largest reconstruction errors mostly occur on geometries with many long, thin features.

\begin{figure*}[!t]
\centering
\setlength{\tabcolsep}{2pt}\renewcommand{\arraystretch}{0.6}\setlength{\qualh}{3.3cm}
\resizebox{\textwidth}{!}{\begin{tabular}{@{}c@{\,}c@{\hspace{5pt}}c@{}}
\rotatebox{90}{\scriptsize GT} & \makebox[0pt][l]{\includegraphics[height=\qualh]{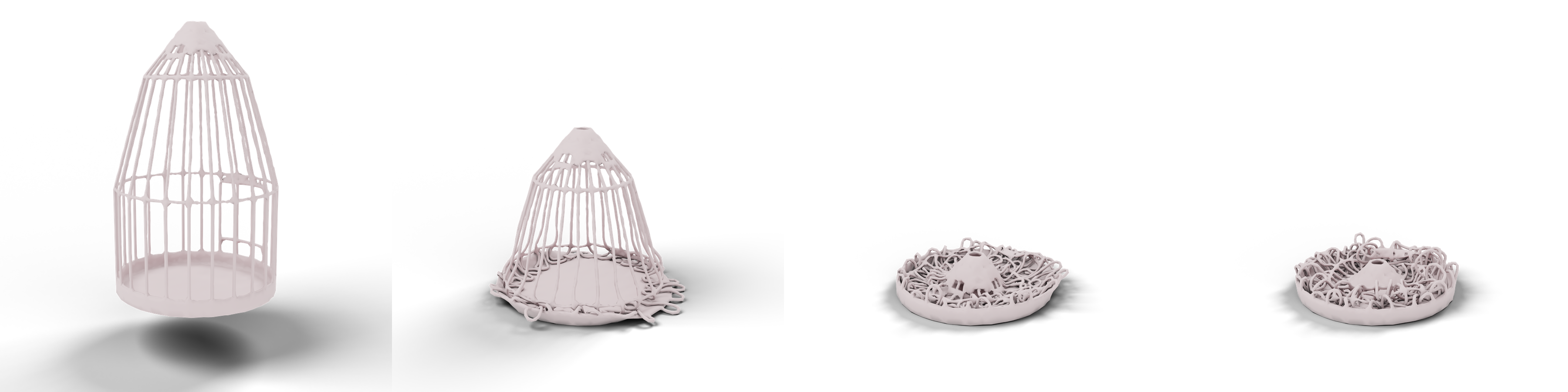}}\makebox[0pt][l]{\raisebox{1.5pt}{\hspace{1.5pt}\scriptsize $N{=}9021$}}\hspace{4.0000\qualh} \\
\rotatebox{90}{\scriptsize\shortstack{\textbf{Holistic}\\\textbf{(ours)}}} & \makebox[0pt][l]{\includegraphics[height=\qualh]{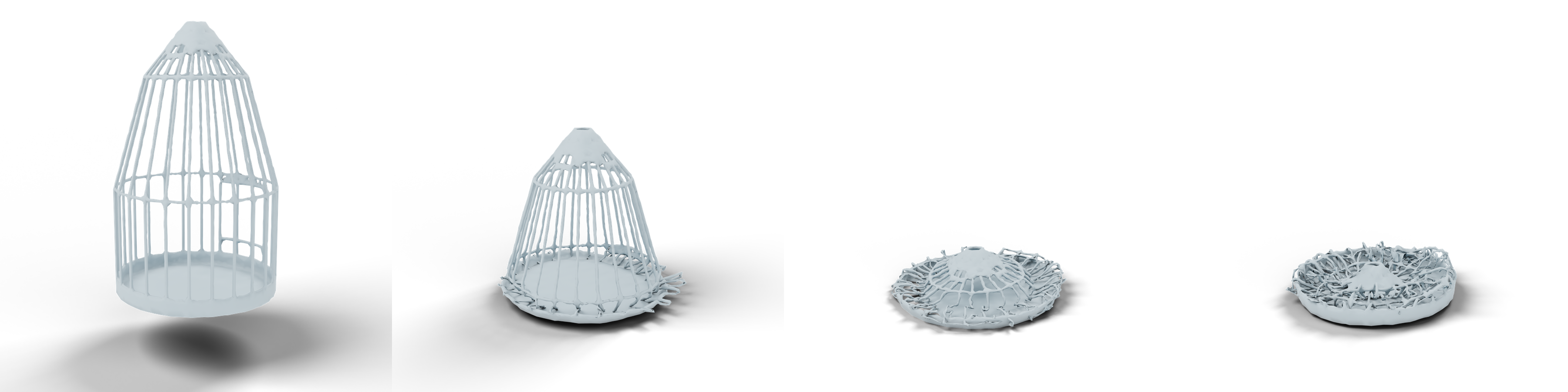}}\makebox[0pt][l]{\hspace{3pt}\tikz[baseline=0pt]{\node[inner sep=0pt,inner xsep=3pt,font=\small\itshape,text=black!65] (t) at ($(2.0000\qualh,0)+(-1.5pt,0)$) {time};\draw[black!65,line width=0.8pt,line cap=round] (0,0) -- (t.west);\draw[black!65,line width=0.8pt,line cap=round,-{Straight Barb[length=5.5pt,width=6.5pt,angle'=60]}] (t.east) -- ($(4.0000\qualh,0)+(-3pt,0)$);}}\hspace{4.0000\qualh} \\
\end{tabular}}
\\[4pt]
\includegraphics[width=0.72\textwidth]{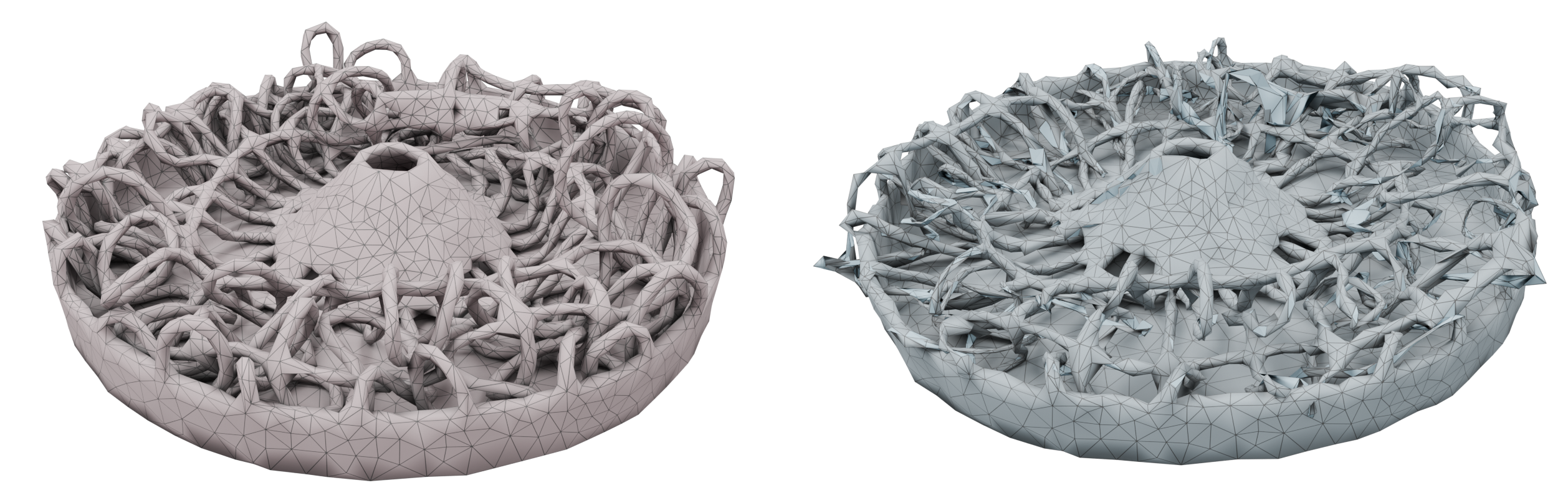}
\caption{ST-VAE reconstruction of the extreme case, a soft birdcage (held-out Objaverse, $N{=}9{,}021$ nodes, $1.8\times$ the $5{,}000$-vertex training cap, $E=5\times10^{4}$\,Pa) with dozens of slender bars. Top: four frames sampled along the trajectory, ground truth (rose) and \textbf{Holistic (ours)}. Bottom: the zoom-in view of the last frame, ground truth (left) and reconstruction (right), with the surface wireframe shown. The overall shape of the collapsed cage is recovered, while the thin bars are reconstructed distorted and tangled. Mean reconstruction error over the trajectory: $9.9$\,mm, against $5.3$\,mm for the compact teaser mug of $11{,}621$ nodes.}
\label{fig:extreme}
\end{figure*}

Fig.~\ref{fig:extreme} illustrates a representative extreme case, a soft birdcage comprising $9{,}021$ nodes, which is $1.8$ times the $5{,}000$-vertex training cap and contains dozens of thin bars only a few tetrahedra thick. Its mean reconstruction error over the trajectory is $9.9$\,mm, about twice the held-out mean for meshes above the training cap (Table~\ref{tab:s2main}). This error is localized to the thin bars, which exhibit distorted and tangled reconstructions. Nevertheless, the overall shape of the collapsed cage, its pose, and its motion over the trajectory are all recovered.

Furthermore, node count alone is not the primary performance bottleneck. For instance, the teaser mug shown in Fig.~\ref{fig:teaser} represents a compact solid with $11{,}621$ nodes ($2.3$ times the training cap) yet achieves a lower reconstruction error of $5.3$\,mm. What distinguishes the birdcage is the density of thin features that deform independently of each other. The folding of every bar is a high-frequency deformation, and the fixed-size latent has to carry it for dozens of bars at once. The latent has enough capacity for the global shape, but not for the fine detail of every bar.

\FloatBarrier
\section{Multi-Body Extension}
\label{app:multibody}
This section describes the multi-body extension of our method and presents its qualitative results. Our approach generalizes effectively to multi-object scenes. The factorization defined in Eq.~\eqref{eq:factoring} is applied at the scene level: a single global rigid transform $(\mathbf{R}_t, \mathbf{c}_t)$ is fitted across the vertices of all bodies, while the relative motion among bodies is captured by the canonical deformation field $\mathbf{d}_t$. Because the spatial VAE encodes all vertices within a scene through a shared set of $K$ query tokens, multi-body scenes are encoded, compressed, and predicted in the same manner as single-body objects, requiring no architectural modifications to the ST-VAE or the LDM.

We train both the ST-VAE and the LDM on scenes containing up to three bodies ($n \le 3$), where randomly selected objects are thrown toward each other to collide in mid-air before hitting the ground. Each body is assigned an independent random Young's modulus and initial velocity.

\begin{figure*}[!t]
\centering
\setlength{\tabcolsep}{2pt}\renewcommand{\arraystretch}{0.6}
\resizebox{\textwidth}{!}{\begin{tabular}{@{}c@{\,}c@{\hspace{5pt}}c@{}}
\rotatebox{90}{\scriptsize GT} & \makebox[0pt][l]{\includegraphics[height=\qualh]{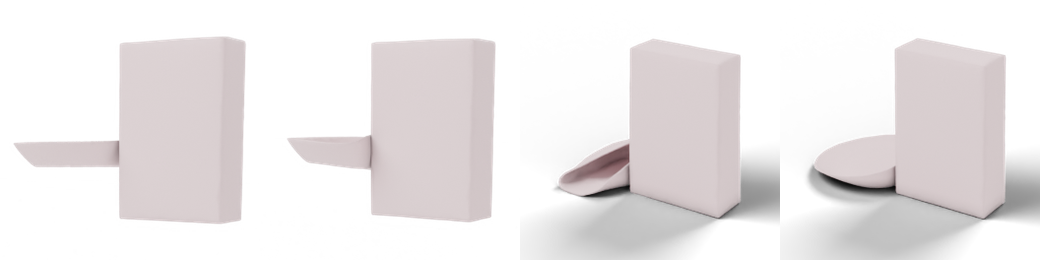}}\makebox[0pt][l]{\raisebox{1.5pt}{\hspace{1.5pt}\scriptsize $N{=}5575$}}\hspace{4.0000\qualh} & \makebox[0pt][l]{\includegraphics[height=\qualh]{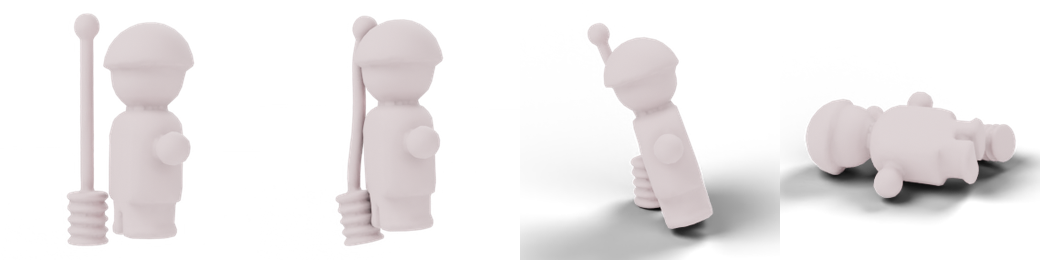}}\makebox[0pt][l]{\raisebox{1.5pt}{\hspace{1.5pt}\scriptsize $N{=}3698$}}\hspace{4.0000\qualh} \\
\rotatebox{90}{\scriptsize Ours} & \makebox[0pt][l]{\includegraphics[height=\qualh]{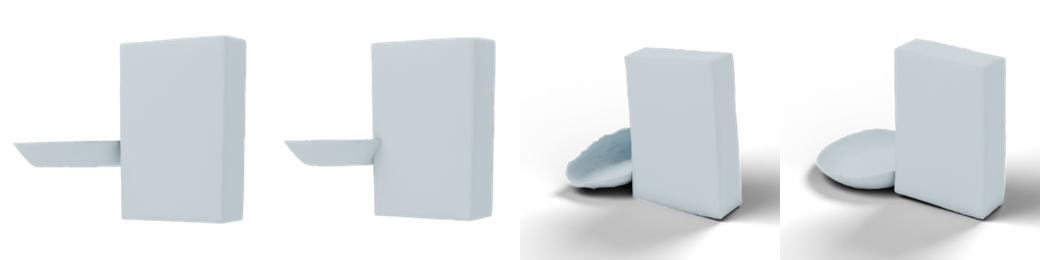}}\makebox[0pt][l]{\hspace{3pt}\tikz[baseline=0pt]{\node[inner sep=0pt,inner xsep=3pt,font=\small\itshape,text=black!65] (t) at ($(2.0000\qualh,0)+(-1.5pt,0)$) {time};\draw[black!65,line width=0.8pt,line cap=round] (0,0) -- (t.west);\draw[black!65,line width=0.8pt,line cap=round,-{Straight Barb[length=5.5pt,width=6.5pt,angle'=60]}] (t.east) -- ($(4.0000\qualh,0)+(-3pt,0)$);}}\hspace{4.0000\qualh} & \includegraphics[height=\qualh]{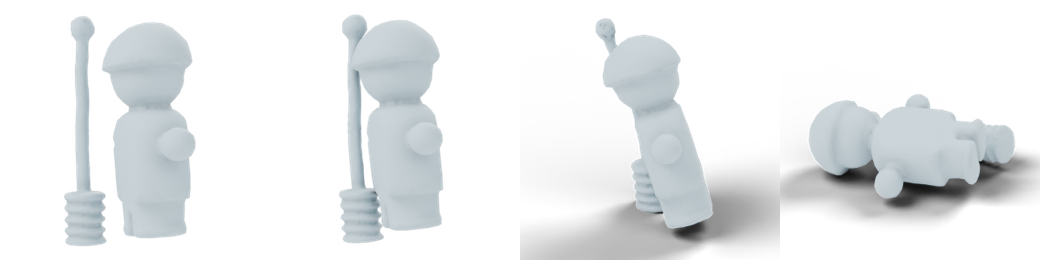} \\
\noalign{\vskip 4pt}
\rotatebox{90}{\scriptsize GT} & \makebox[0pt][l]{\includegraphics[height=\qualh]{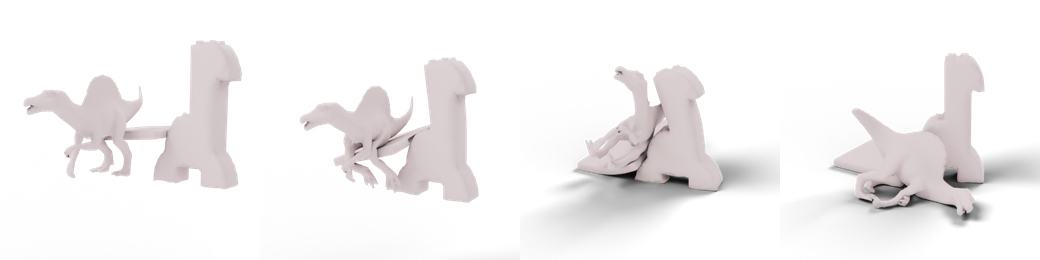}}\makebox[0pt][l]{\raisebox{1.5pt}{\hspace{1.5pt}\scriptsize $N{=}5589$}}\hspace{4.0000\qualh} & \makebox[0pt][l]{\includegraphics[height=\qualh]{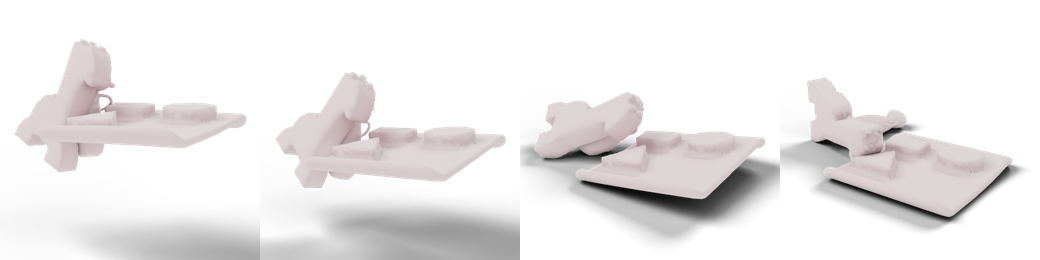}}\makebox[0pt][l]{\raisebox{1.5pt}{\hspace{1.5pt}\scriptsize $N{=}5738$}}\hspace{4.0000\qualh} \\
\rotatebox{90}{\scriptsize Ours} & \includegraphics[height=\qualh]{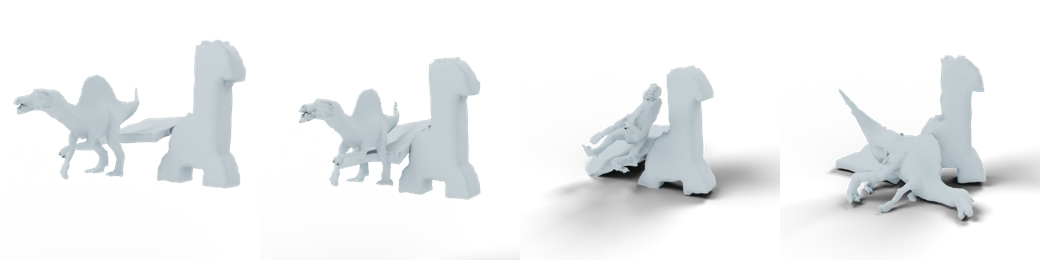} & \includegraphics[height=\qualh]{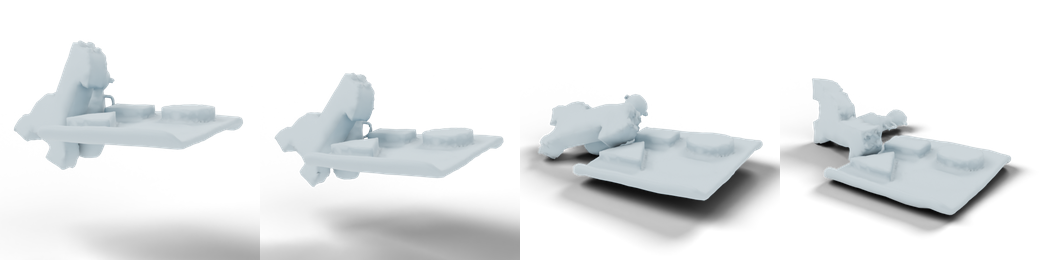} \\
\end{tabular}}

\caption{Scene-level ST-VAE reconstruction on held-out multi-body scenes (two two-body and two three-body scenes, each body with a random Young's modulus and initial velocity) at four sampled frames: just before and just after the mid-air collision, the first frame after ground impact, and a settled frame. Rows: ground truth (rose) and reconstruction.}
\label{fig:multibody-recon}
\end{figure*}

\begin{figure*}[!t]
\centering
\setlength{\tabcolsep}{2pt}\renewcommand{\arraystretch}{0.6}
\resizebox{\textwidth}{!}{\begin{tabular}{@{}c@{\,}c@{\hspace{5pt}}c@{}}
\rotatebox{90}{\scriptsize GT} & \makebox[0pt][l]{\includegraphics[height=\qualh]{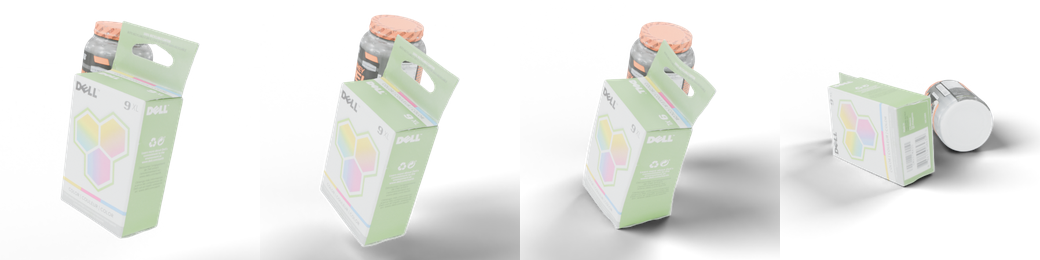}}\makebox[0pt][l]{\raisebox{1.5pt}{\hspace{1.5pt}\scriptsize $N{=}6078$}}\hspace{4.0000\qualh} & \makebox[0pt][l]{\includegraphics[height=\qualh]{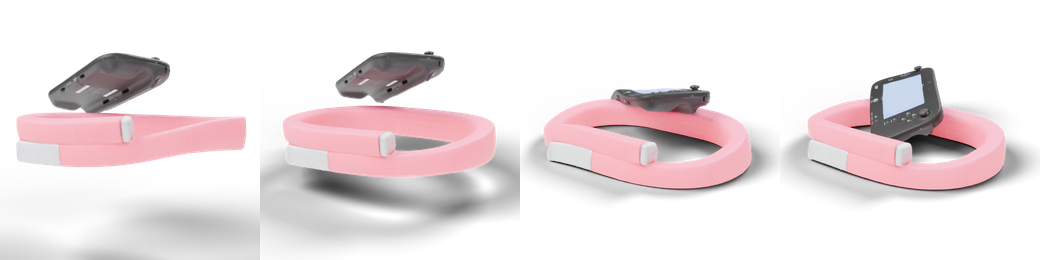}}\makebox[0pt][l]{\raisebox{1.5pt}{\hspace{1.5pt}\scriptsize $N{=}5792$}}\hspace{4.0000\qualh} \\
\rotatebox{90}{\scriptsize Prediction} & \makebox[0pt][l]{\includegraphics[height=\qualh]{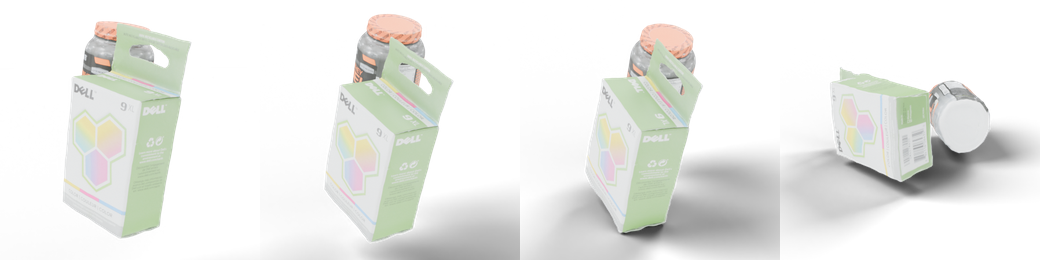}}\makebox[0pt][l]{\hspace{3pt}\tikz[baseline=0pt]{\node[inner sep=0pt,inner xsep=3pt,font=\small\itshape,text=black!65] (t) at ($(2.0000\qualh,0)+(-1.5pt,0)$) {time};\draw[black!65,line width=0.8pt,line cap=round] (0,0) -- (t.west);\draw[black!65,line width=0.8pt,line cap=round,-{Straight Barb[length=5.5pt,width=6.5pt,angle'=60]}] (t.east) -- ($(4.0000\qualh,0)+(-3pt,0)$);}}\hspace{4.0000\qualh} & \includegraphics[height=\qualh]{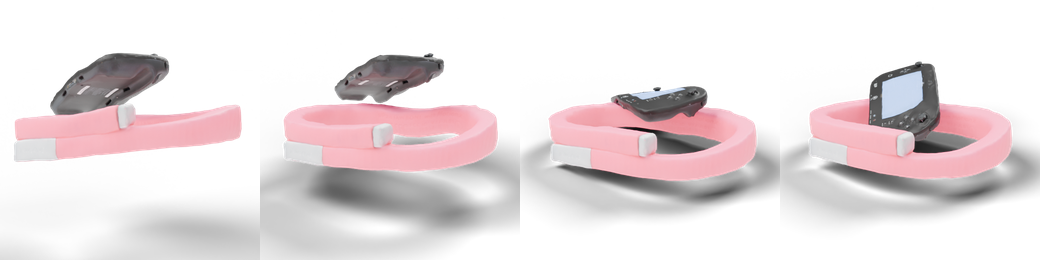} \\
\noalign{\vskip 4pt}
\rotatebox{90}{\scriptsize GT} & \makebox[0pt][l]{\includegraphics[height=\qualh]{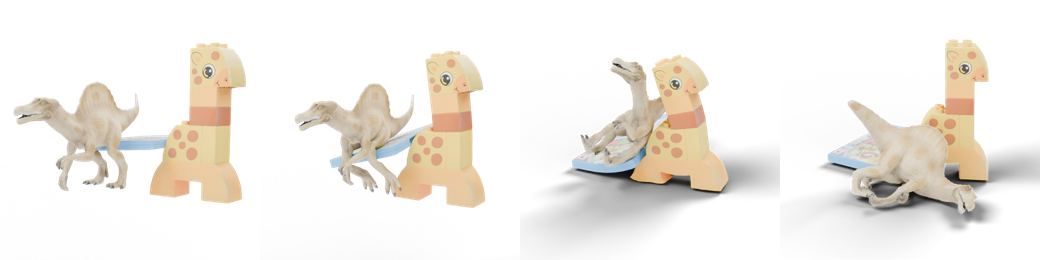}}\makebox[0pt][l]{\raisebox{1.5pt}{\hspace{1.5pt}\scriptsize $N{=}5589$}}\hspace{4.0000\qualh} & \makebox[0pt][l]{\includegraphics[height=\qualh]{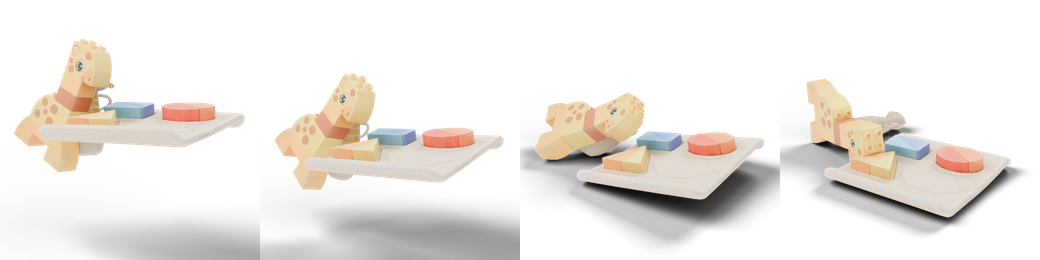}}\makebox[0pt][l]{\raisebox{1.5pt}{\hspace{1.5pt}\scriptsize $N{=}5738$}}\hspace{4.0000\qualh} \\
\rotatebox{90}{\scriptsize Prediction} & \includegraphics[height=\qualh]{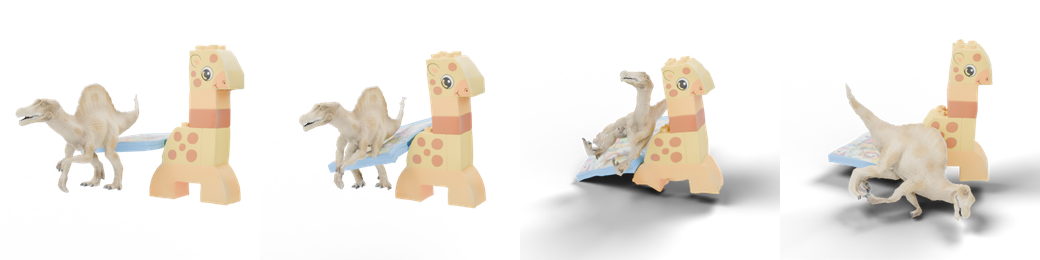} & \includegraphics[height=\qualh]{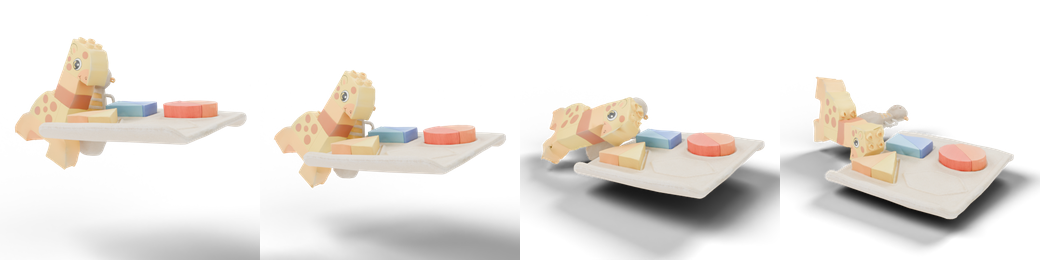} \\
\end{tabular}}

\caption{One-shot prediction on held-out multi-body scenes (two two-body and two three-body scenes, each body with a random Young's modulus and initial velocity), sampled frames as in Fig.~\ref{fig:multibody-recon}. Rows: ground truth and our prediction conditioned on the initial state, rendered by driving each body's original textured asset with the predicted tetrahedral trajectory.}
\label{fig:multibody-pred}
\end{figure*}

Fig.~\ref{fig:multibody-recon} illustrates the ST-VAE reconstruction on four multi-body scenes across four representative frames: immediately before and after the mid-air collision, the first frame following ground impact, and the final settled state. All bodies within each scene, including their collisions and ground contacts, are successfully reconstructed through the single global transform. Fig.~\ref{fig:multibody-pred} demonstrates one-shot prediction on four multi-body scenes conditioned on the initial state, with predictions rendered using the original textured assets. The entire scene, encompassing each individual body along with its collision dynamics and landing behavior, is predicted end-to-end in a single diffusion pass.

\FloatBarrier
\section{Differentiable Applications: Setup and Results}
\label{app:diffapp}
This section gives the setup and the full results of the two differentiable tasks of Sec.~\ref{sec:diffapp}. Both tasks run on the same scene and use the same model. The first, material estimation, needs a first-order gradient. The second, designing the throw, needs a second-order gradient. Table~\ref{tab:diffapp_merged} summarizes how our model and the two differentiable simulators, DiffIPC \citep{DBLP:journals/tog/HuangTGFSPZ24} and Newton/Warp, behave on them.

\begin{table}[h!]\centering\small
\setlength{\tabcolsep}{4pt}\renewcommand{\arraystretch}{1.3}
\caption{Comparison of our model with two differentiable simulators on downstream applications. Our approach supports higher-order derivatives, enabling complex optimization tasks that are either computationally infeasible or numerically unstable for existing solvers.}
\label{tab:diffapp_merged}
\begin{tabular}{@{}p{2.2cm} p{3.2cm} p{3.8cm} p{3.8cm}@{}}
\toprule
Simulator & Differentiation Support & 1st-Order Task \newline (Material Estimation) & 2nd-Order Task \newline (Throw Design) \\
\midrule
\textbf{Ours} & \textbf{Autograd} \newline (higher order) & \textbf{Success (40 seconds)} & \textbf{Success (7 minutes)} \newline Efficient 2nd-order grads. \\
\addlinespace
DiffIPC & Analytic adjoint \newline (1st order only) & \textcolor{orange!85!black}{\textbf{26 minutes}} \newline Unstable under defaults. \newline Needs manual tuning. & \textcolor{red!80!black}{\textbf{Infeasible (Multi-day)}} \newline ${\sim}635\times$ slower via finite diff. \\
\addlinespace
Newton/Warp & Recorded tape & \textcolor{red!80!black}{\textbf{Fails}} \newline Gradients violently explode. & \textcolor{red!80!black}{\textbf{Infeasible}} \newline No usable 1st-order gradients. \\
\bottomrule
\end{tabular}
\end{table}

\subsection{Material estimation}
\begin{figure}[htbp]
\centering
\includegraphics[width=\linewidth]{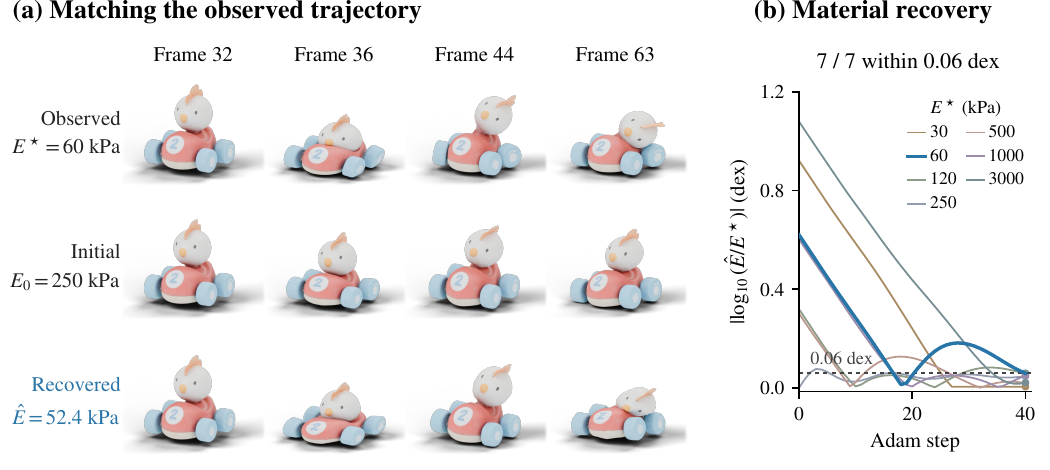}
\caption{\textbf{Basic material estimation.}
(a) Four matched time samples of the observed trajectory ($E^\star=60$\,kPa), the initial model prediction ($E_0=250$\,kPa), and the prediction after 40 Adam steps ($\hat E=52.4$\,kPa). All views share the same camera and scale.
(b) Absolute log-stiffness error for seven ground-truth moduli, with the illustrated case highlighted in blue. All seven final estimates are within $0.06$\,dex; each optimization takes approximately 40 seconds. Curves show the recorded iterates without smoothing.}
\label{fig:diff_material}
\end{figure}

The first task is to recover the Young's modulus $E^\star$ of a body from a single observed trajectory. Because $E$ enters our model through the scalar material channel, the mismatch between the predicted rollout and the observation is end-to-end differentiable with respect to $\log E$. Starting from initial guesses up to 1.1\,dex away from seven ground-truth moduli $E^\star\in[3{\cdot}10^4,3{\cdot}10^6]$, our model successfully recovers all seven within $0.06$\,dex in just 40 Adam steps ($\sim$40 seconds).
Figure~\ref{fig:diff_material} compares the observed, initial, and recovered trajectories and shows the estimation error across all seven moduli.

Existing differentiable simulators already have difficulty with this first-order task. DiffIPC does not converge under default settings and only runs stably after manual tuning of the solver parameters, and then needs about 26 minutes. Newton/Warp, meanwhile, fails on this task: it consumes massive memory (49\,GB tape for 131k substeps) before its gradients explode through the complex dynamics simulation, resulting in random gradient signs and total optimization failure.

\subsection{Beyond material estimation: designing the throw}
Building on the first task, we ask a further question: which throw makes the stiffness easiest to estimate? A throw is informative when a small change in $E$ produces a large change in the resulting trajectory.

\paragraph{Task Setup.}
Crucially, how large this trajectory change is depends on the stiffness of the body itself. Therefore, the trajectory sensitivity must be evaluated at a specific value of $E$, which we set to $E^\star$ ($6{\cdot}10^4$\,Pa). The design is thus local: it finds the most informative throw for a body of roughly this stiffness, assuming a rough estimate of the stiffness is already available. To ensure robustness against launch errors, we maximize the sensitivity of the predicted rigid motion to $\log_{10}E$, averaged over eight small perturbations of the release velocity $v_0$. Because the objective itself is a derivative, maximizing it via gradient ascent fundamentally requires a \emph{second-order gradient}.

\begin{figure}[htbp]
\centering
\includegraphics[width=\linewidth]{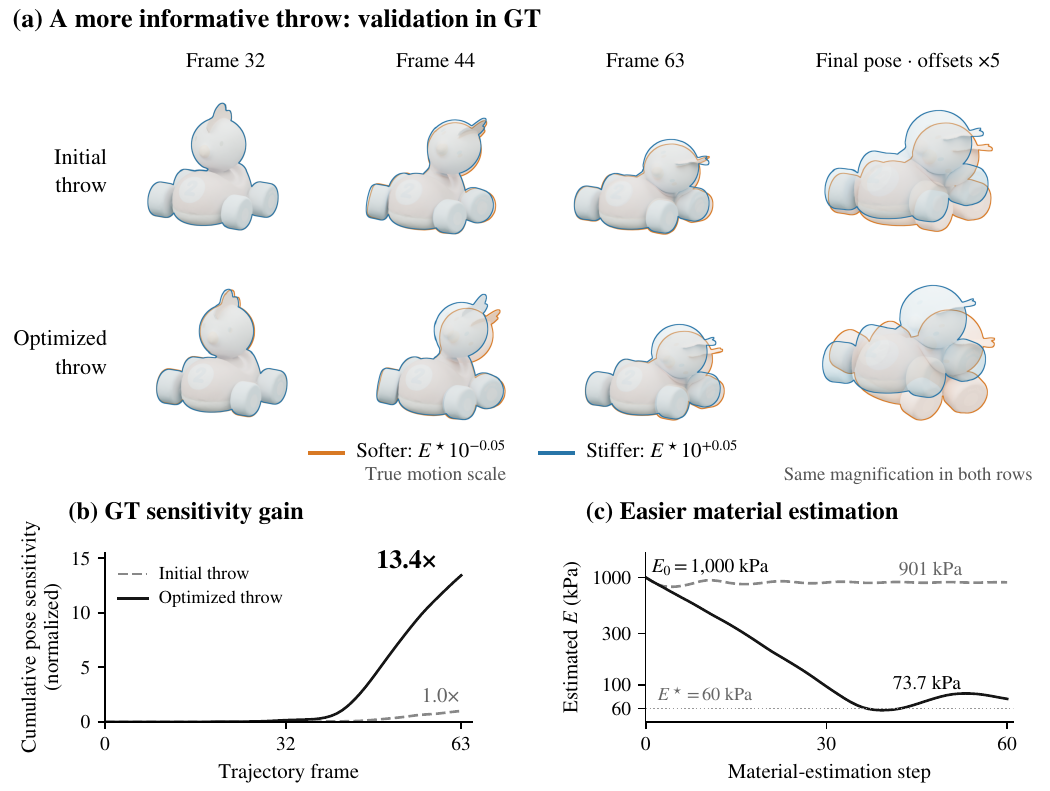}
\caption{\textbf{Beyond material estimation: designing a more informative throw.}
(a) Ground-truth replays of the initial and optimized release velocities at $E^\star=60$\,kPa. Orange and blue overlays represent stiffness perturbations of $-0.05$ and $+0.05$\,dex. The first three columns show true motion at matched times and a shared camera scale. In the final column, rigid translation and rotation offsets from the nominal trajectory are magnified by $5\times$, using equal magnification in both rows.
(b) Cumulative rigid-pose sensitivity, normalized by the initial throw's final score, increases by $13.4\times$. This score is computed from the unamplified GT trajectories.
(c) Material estimation from the same distant initial guess, $E_0=1{,}000$\,kPa. After 60 estimation steps, the initial throw remains near $901$\,kPa, whereas the optimized throw reaches $73.7$\,kPa, much closer to the true $60$\,kPa. Throw design uses 30 Adam steps and approximately 7 minutes.}
\label{fig:diff_throw}
\end{figure}

\paragraph{Optimization Results.}
Our model supports higher-order derivatives and provides this second-order gradient directly via reverse-mode automatic differentiation. In just 30 Adam steps (7 minutes), our model successfully optimizes the throw. Replaying this designed throw in the ground-truth solver confirms its effectiveness: the sensitivity of the rigid motion to $E$ increases by $13\times$, and per-vertex sensitivity by $1.85\times$. This highly informative trajectory significantly widens the basin of attraction, enabling successful $E$ recovery even from a distant initial guess ($E_0=10^6$) where the original throw stalls.
Figure~\ref{fig:diff_throw} visualizes the ground-truth trajectories of the initial and the optimized throw, the sensitivity gain, and the subsequent material-estimation runs from this distant initial guess.

\paragraph{Comparison with Differentiable Simulators.}
If the first-order task is already this challenging for existing solvers, this second-order throw design becomes infeasible for them. DiffIPC is inherently limited to first-order analytic adjoints; forcing it to approximate the necessary second-order gradient requires nested finite differences ($64$ forward solves per step). As a result, a single gradient step takes 2.4 hours, about $635\times$ slower than our model. The whole design would therefore take about three days, which is not acceptable in practice. For Newton/Warp, which already fails to produce a usable first-order gradient, such higher-order applications are insurmountable. In contrast, PhysLDM effortlessly scales to these advanced demands, acting as a fast, stable surrogate that natively supports higher-order derivatives. Table~\ref{tab:diffapp_merged} summarizes the comparison on both tasks.

\FloatBarrier
\section{Comparison with Concurrent Work}
\label{app:physiformer}
This section compares PhysLDM with the concurrent one-shot model of \citet{DBLP:journals/corr/abs-2606-27364}. That model is a diffusion transformer with factorized spatiotemporal attention that denoises the full $T$-frame trajectory directly on vertex$\times$frame tokens, without any latent, and it was trained on low-resolution surface meshes (about $356$ vertices per scene, $35$ templates). Because its cost grows with the number of vertices times the number of frames, it cannot be trained at the resolution of our dataset, so the two models cannot be compared directly on our full data. We therefore compare them in two steps. First, we measure the vertex count up to which the raw-coordinate formulation fits in memory at all. Second, we choose a resolution where it does fit, train both models on the same data there, and compare their training cost and generation quality.

\subsection{Memory and resolution limits}
Table~\ref{tab:pfwall} reports peak training memory and GPU time per sample at batch~$1$ ($T{=}64$, \texttt{bf16}, flash attention, one $140$\,GiB H200) for the published backbone of the concurrent work (DiT-L, $340$M parameters) and for its DiT-B variant ($130$M, with the same width and depth as our LDM), for vertex counts from $500$ to $5{,}000$. The published backbone runs out of memory above $1{,}000$ vertices and the DiT-B variant above $2{,}750$. Neither fits the median scene of our dataset (${\sim}3{,}400$ vertices) even at batch~$1$, and the $5{,}000$-vertex cap is further out of reach. Our LDM works on the latent and trains at ${\sim}4.4$\,GiB and ${\sim}27$\,ms per trajectory at any vertex count.

\begin{table}[t]\centering\small
\setlength{\tabcolsep}{5pt}
\caption{Memory and resolution limits of the concurrent raw-coordinate model on a $140$\,GiB H200: peak training memory (GiB) and GPU time per training step (ms) at batch~$1$, i.e., per trajectory, $T{=}64$, \texttt{bf16}, flash attention, for vertex counts $N$ from $500$ to $5{,}000$. OOM: out of memory at batch~$1$. For reference, our LDM trains at ${\sim}4.4$\,GiB (Table~\ref{tab:latent}) and ${\sim}27$\,ms per trajectory at any $N$.}
\label{tab:pfwall}
\begin{tabular}{@{}rrcc@{}}
\toprule
$N$ & tokens & Concurrent (DiT-L) & Concurrent (DiT-B) \\
 & ($T{\times}N$) & GiB / ms & GiB / ms \\
\midrule
500 & 32k & 64.4 / 669 & 24.0 / 212 \\
1{,}000 & 64k & 122.1 / 1317 & 46.0 / 412 \\
1{,}200 & 77k & OOM & 54.7 / 495 \\
2{,}000 & 128k & OOM & 89.9 / 879 \\
2{,}500 & 160k & OOM & 111.8 / 1147 \\
2{,}750 & 176k & OOM & 123 / 1289 \\
3{,}000 & 192k & OOM & OOM \\
3{,}500 & 224k & OOM & OOM \\
5{,}000 & 320k & OOM & OOM \\
\bottomrule
\end{tabular}
\end{table}

\subsection{Training setting and cost}
At a resolution where the concurrent model does fit in memory, we train both models from scratch under the same conditions. We train the concurrent model's DiT-B variant with their training procedure and our full three-stage pipeline on the same training scenes with at most $1{,}000$ vertices. The comparison is budget-matched: both models are trained on the same scenes on the same GPUs, the concurrent DiT-B has the same width and depth as our LDM, and each deploys its validation-selected checkpoint. Table~\ref{tab:pfcost} gives the training time of both. Our full three-stage pipeline trains in less time than the single-stage concurrent model, and most of our time goes to the ST-VAE rather than the LDM.

\begin{table}[t]\centering\small
\setlength{\tabcolsep}{4pt}
\caption{Training time of the two models on the same scenes ($\leq 1{,}000$ vertices) and the same GPUs. Even with three stages, our pipeline takes less time than the concurrent model.}
\label{tab:pfcost}
\begin{tabular}{@{}llr@{}}
\toprule
Model & Stage & Training time \\
\midrule
Concurrent (DiT-B) & (single) & 74.6\,h \\
\midrule
Ours & spatial VAE & 1.4\,h \\
 & ST-VAE & 40.0\,h \\
 & LDM & 19.6\,h \\
 & total & \textbf{61.1\,h} \\
\bottomrule
\end{tabular}
\end{table}

\subsection{Prediction results comparison}
\begin{figure*}[!t]
\centering
\setlength{\tabcolsep}{0pt}\renewcommand{\arraystretch}{0.6}
\resizebox{\textwidth}{!}{\begin{tabular}{@{}c@{\,}c@{\hspace{1pt}}c@{\hspace{6pt}}c@{\hspace{1pt}}c@{\hspace{6pt}}c@{\hspace{1pt}}c@{\hspace{6pt}}c@{\hspace{1pt}}c@{}}
\rotatebox{90}{\scriptsize GT} & \includegraphics[height=\qualh]{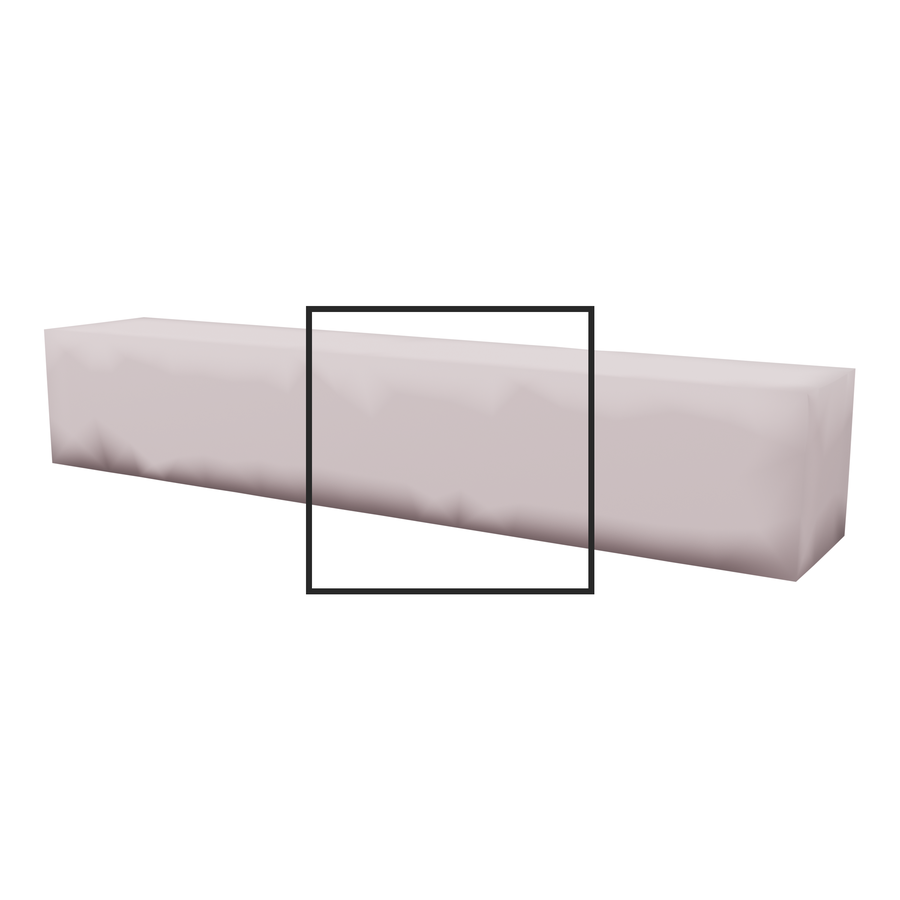} & \includegraphics[height=\qualh]{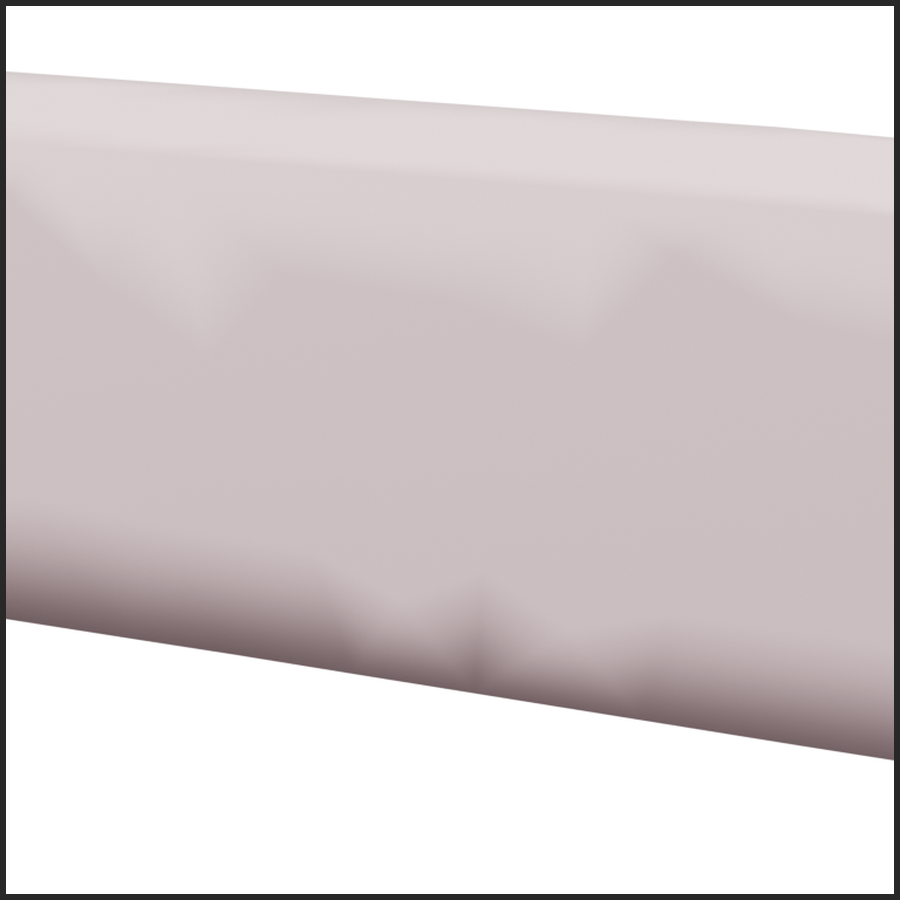} & \includegraphics[height=\qualh]{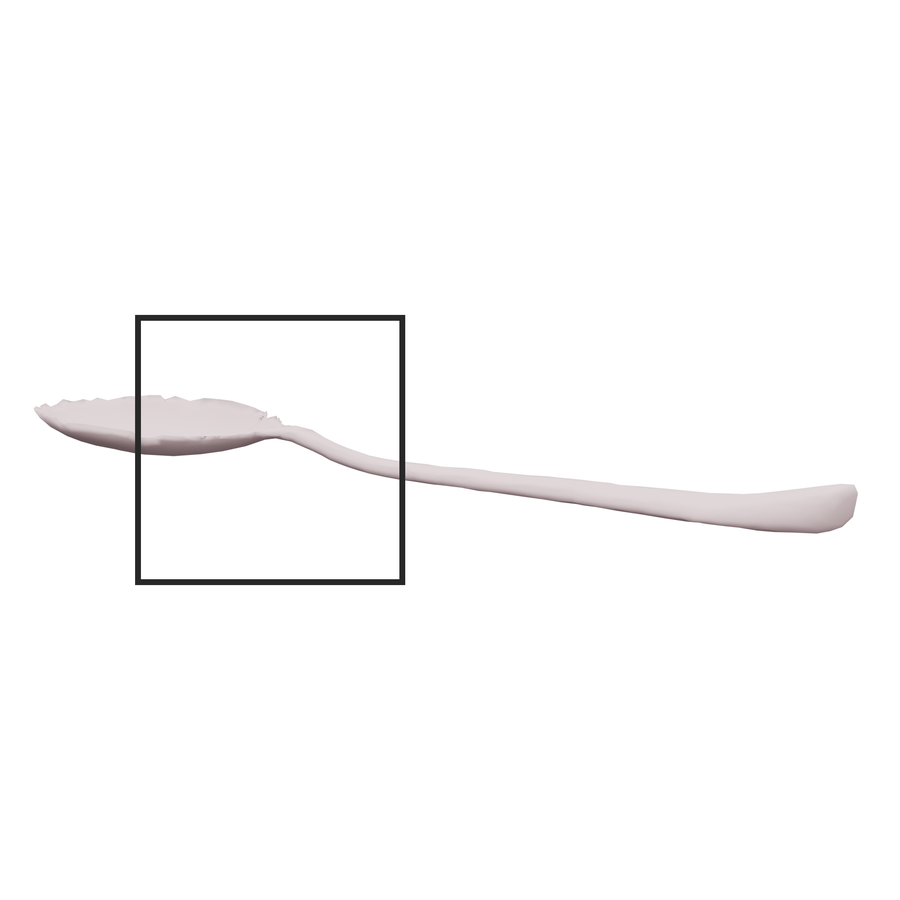} & \includegraphics[height=\qualh]{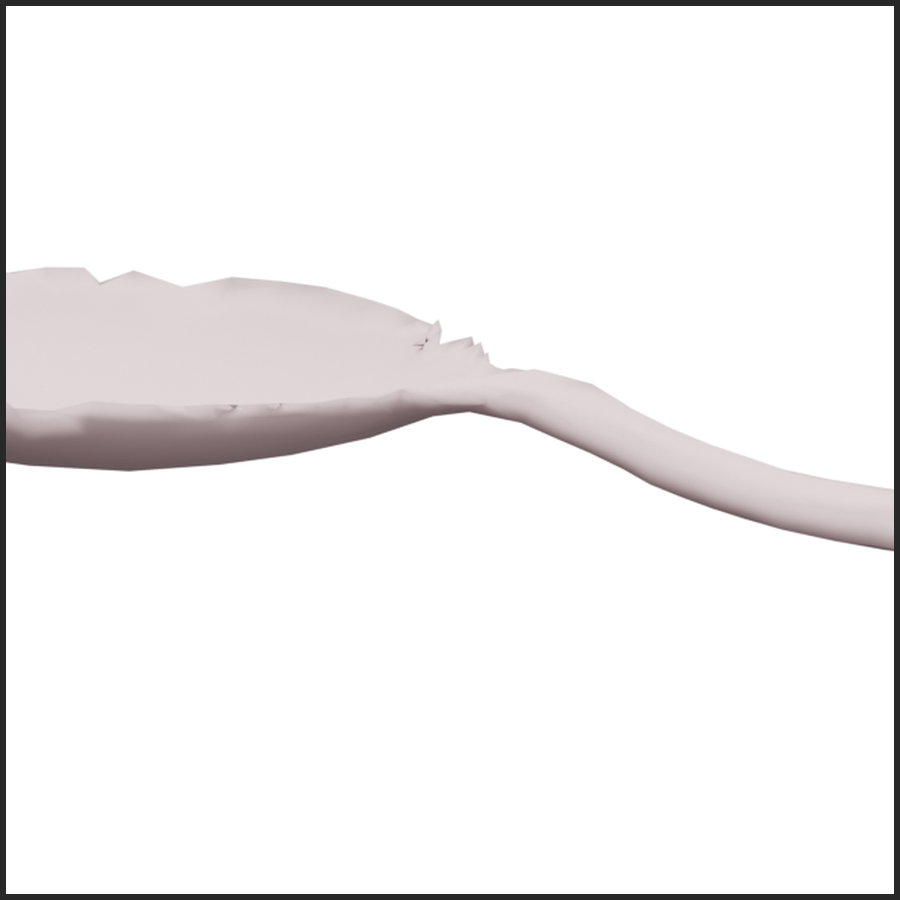} & \includegraphics[height=\qualh]{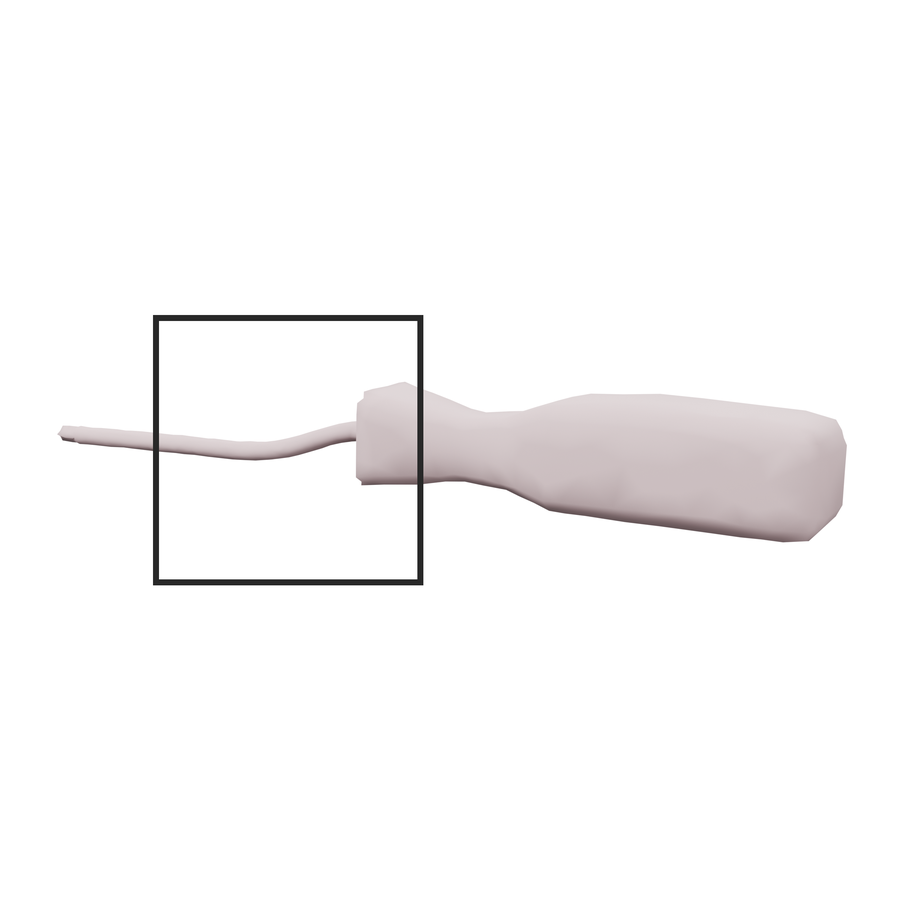} & \includegraphics[height=\qualh]{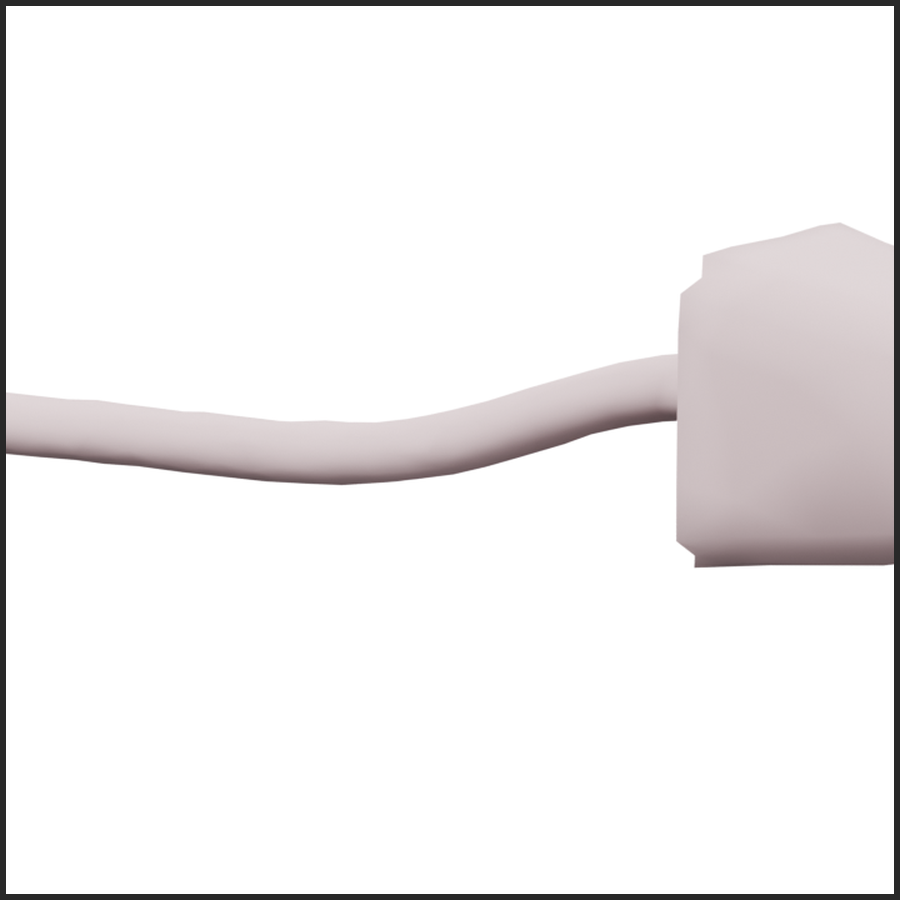} & \includegraphics[height=\qualh]{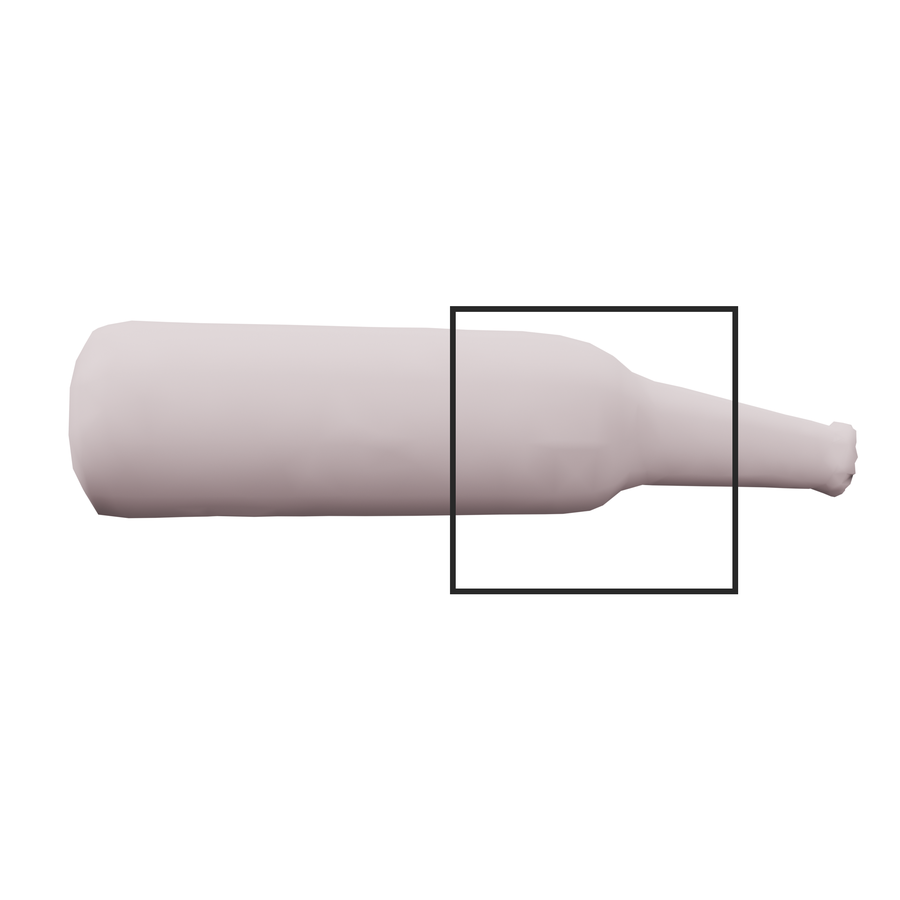} & \includegraphics[height=\qualh]{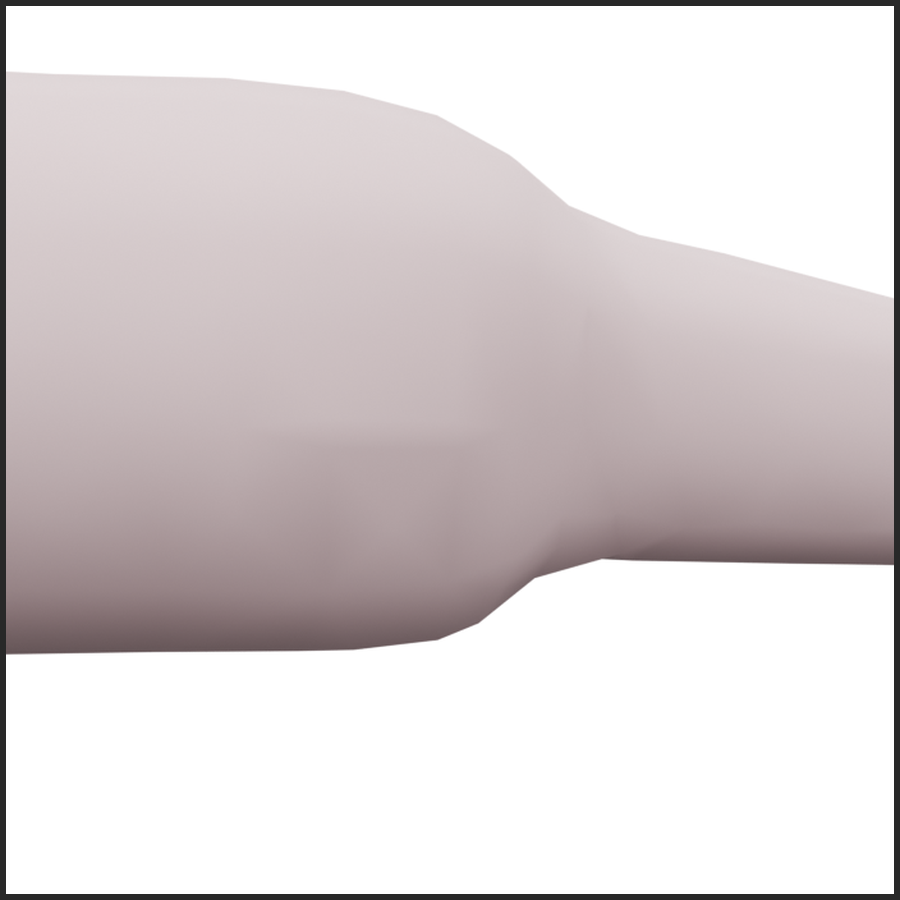} \\
\rotatebox{90}{\scriptsize \shortstack{\textbf{Ours}\\\textbf{(same budget)}}} & \includegraphics[height=\qualh]{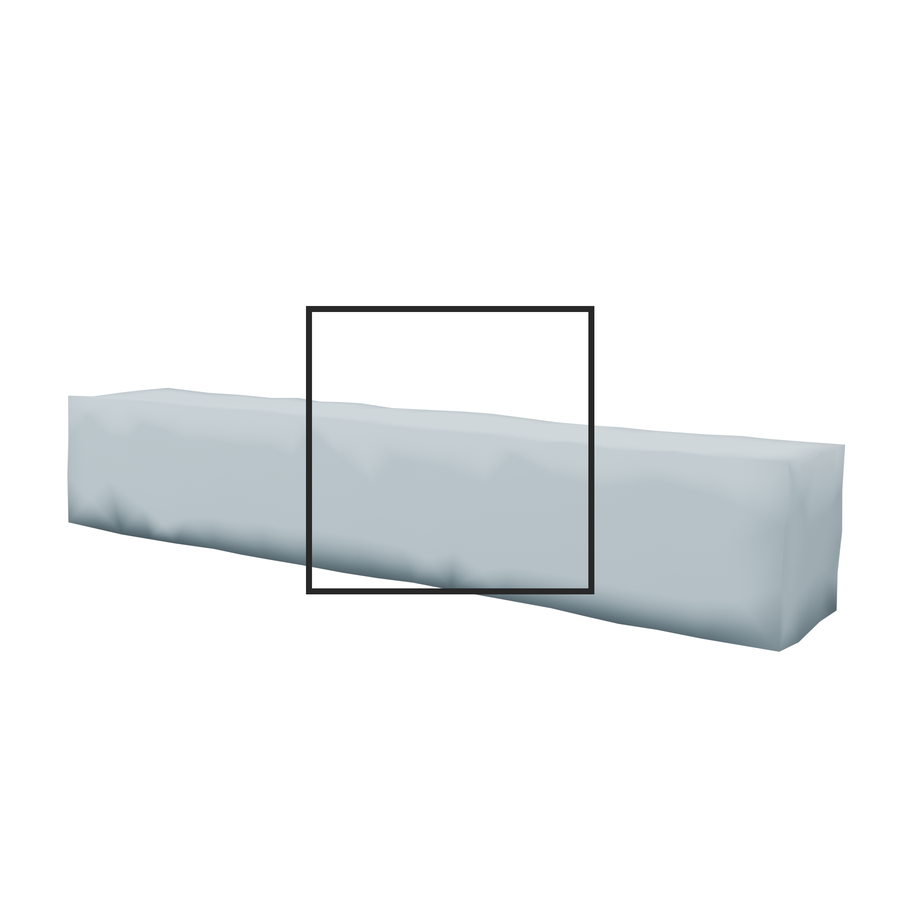} & \includegraphics[height=\qualh]{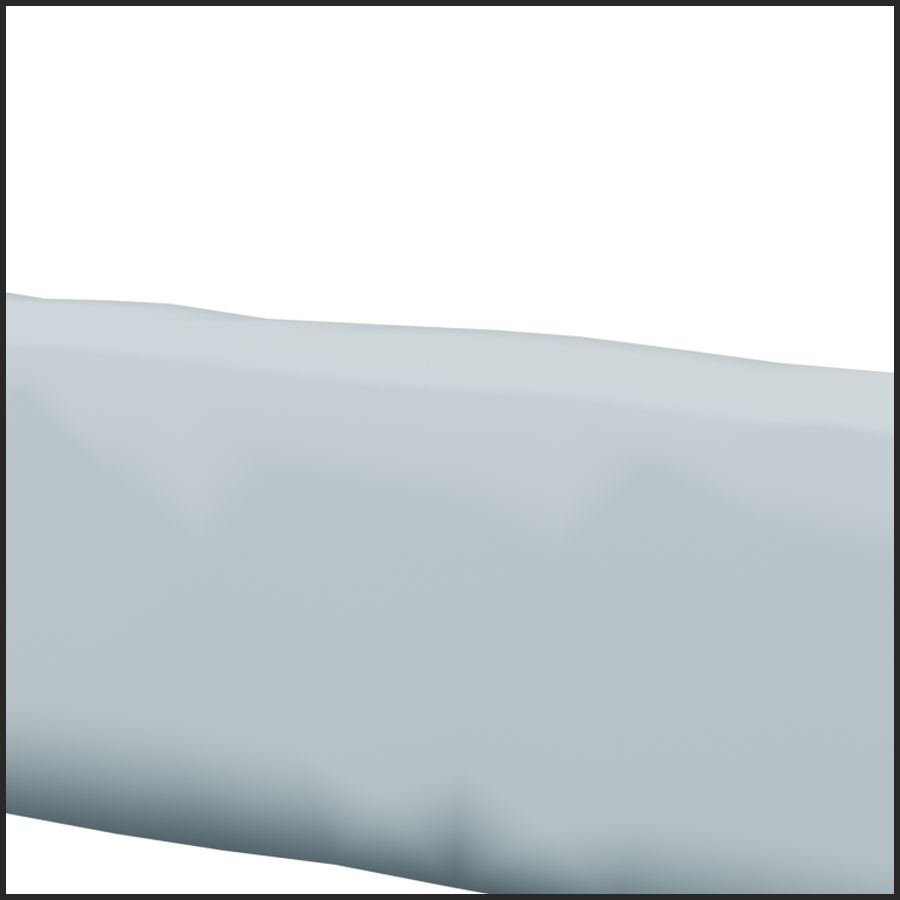} & \includegraphics[height=\qualh]{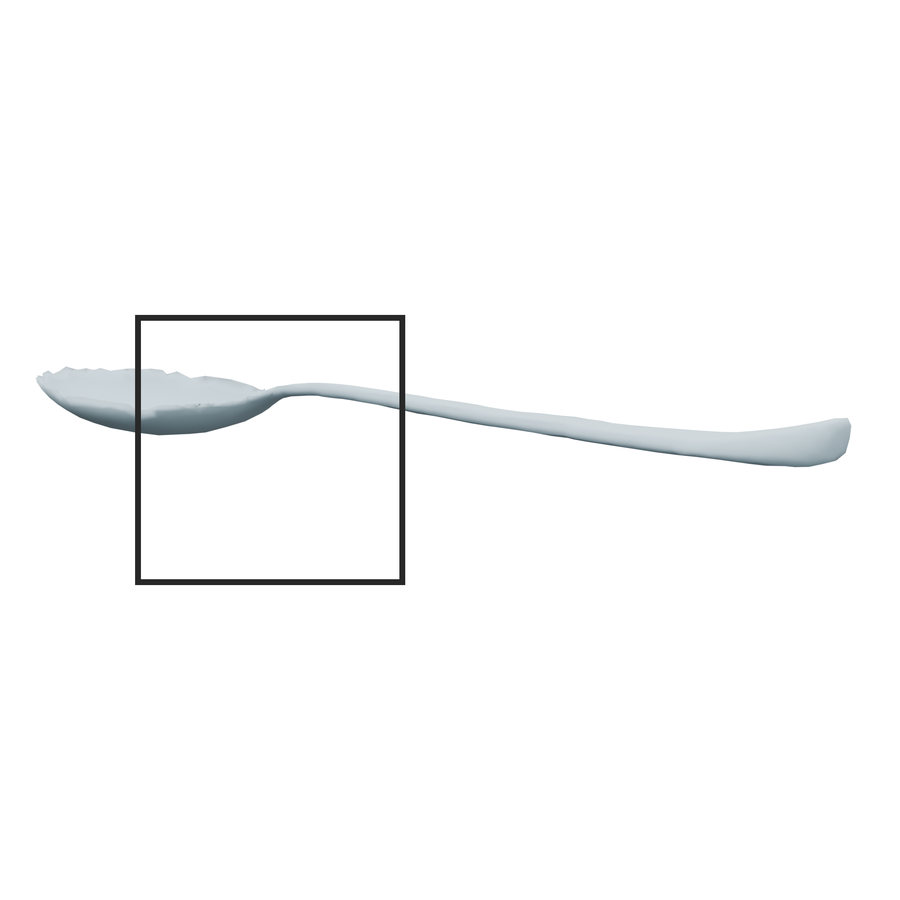} & \includegraphics[height=\qualh]{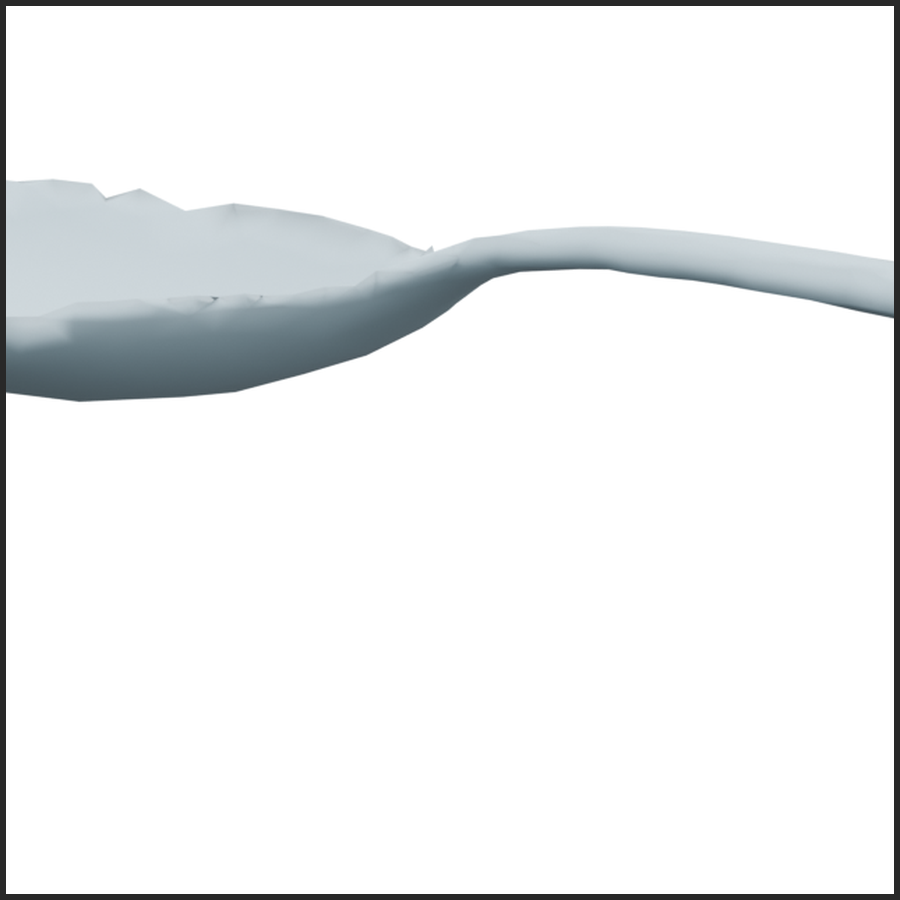} & \includegraphics[height=\qualh]{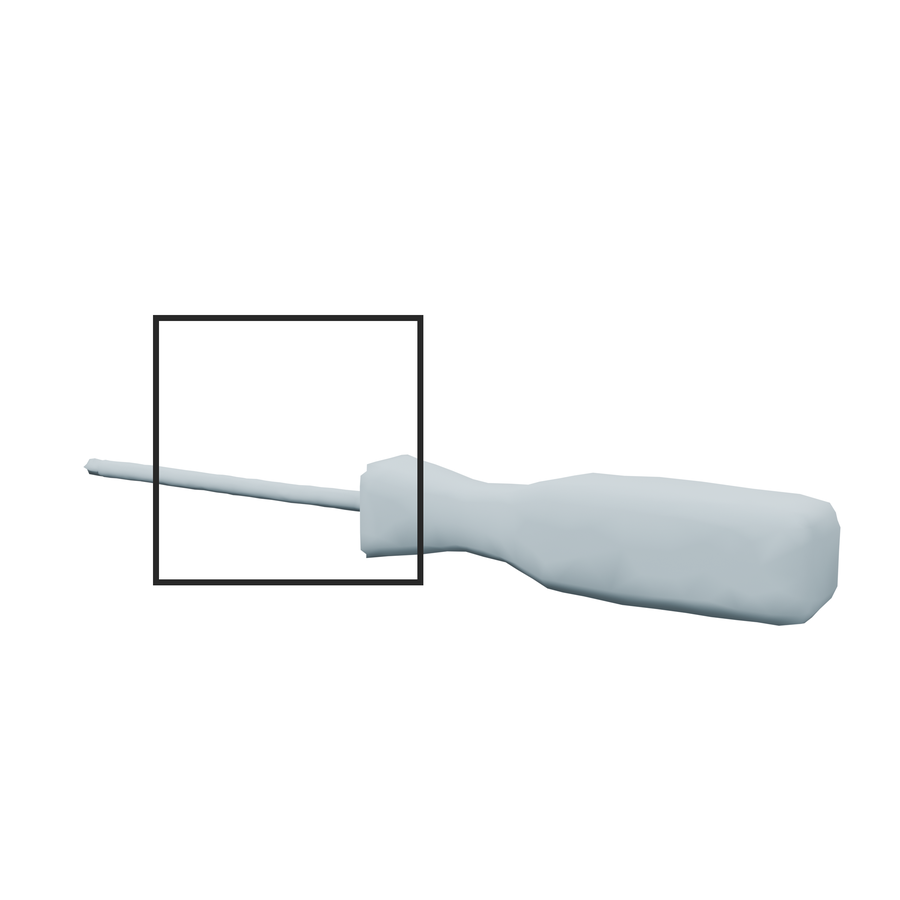} & \includegraphics[height=\qualh]{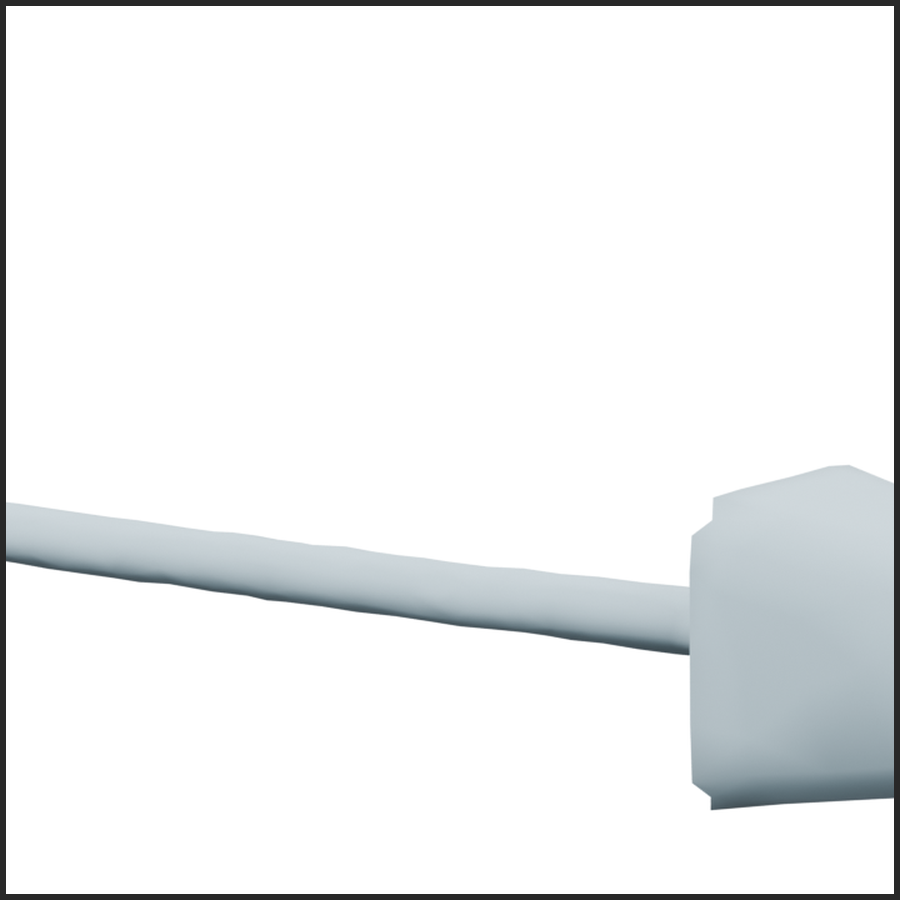} & \includegraphics[height=\qualh]{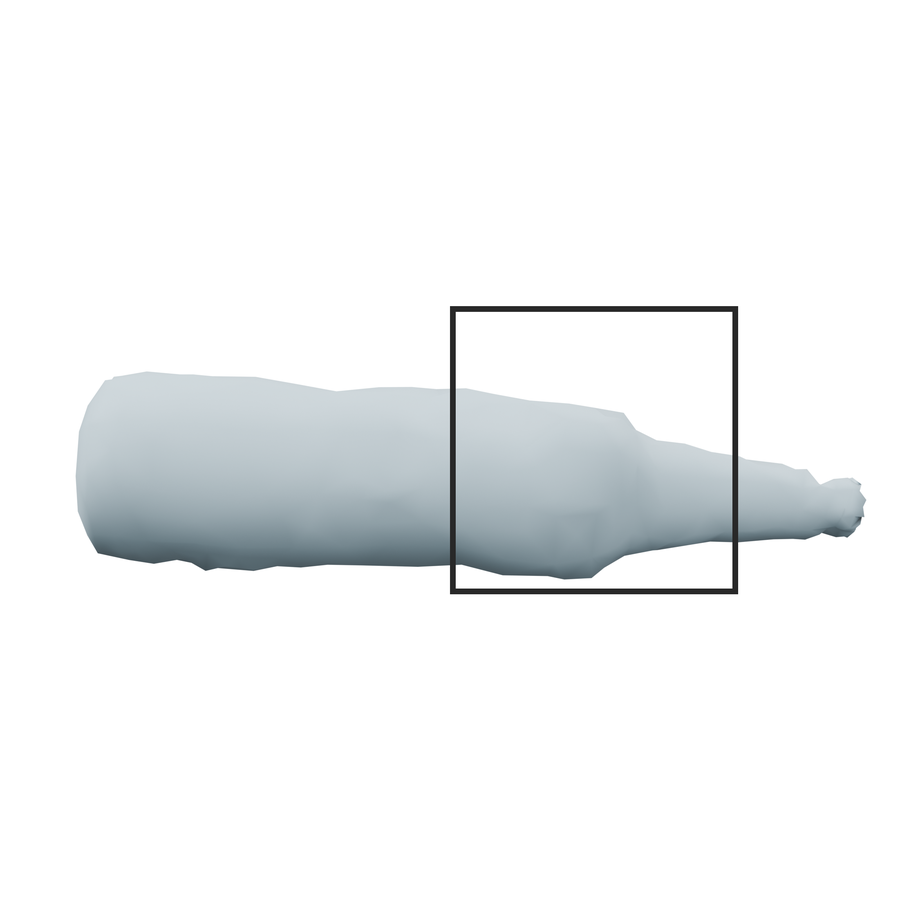} & \includegraphics[height=\qualh]{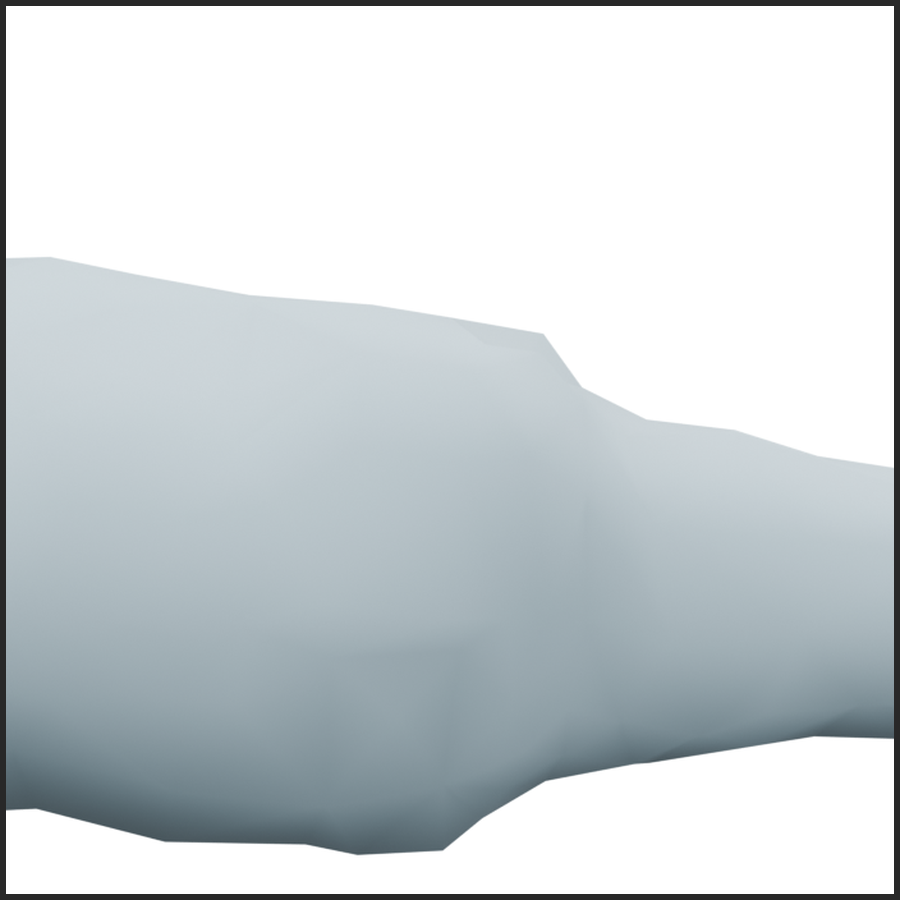} \\
\rotatebox{90}{\scriptsize \shortstack{Concurrent\\(DiT-B)}} & \includegraphics[height=\qualh]{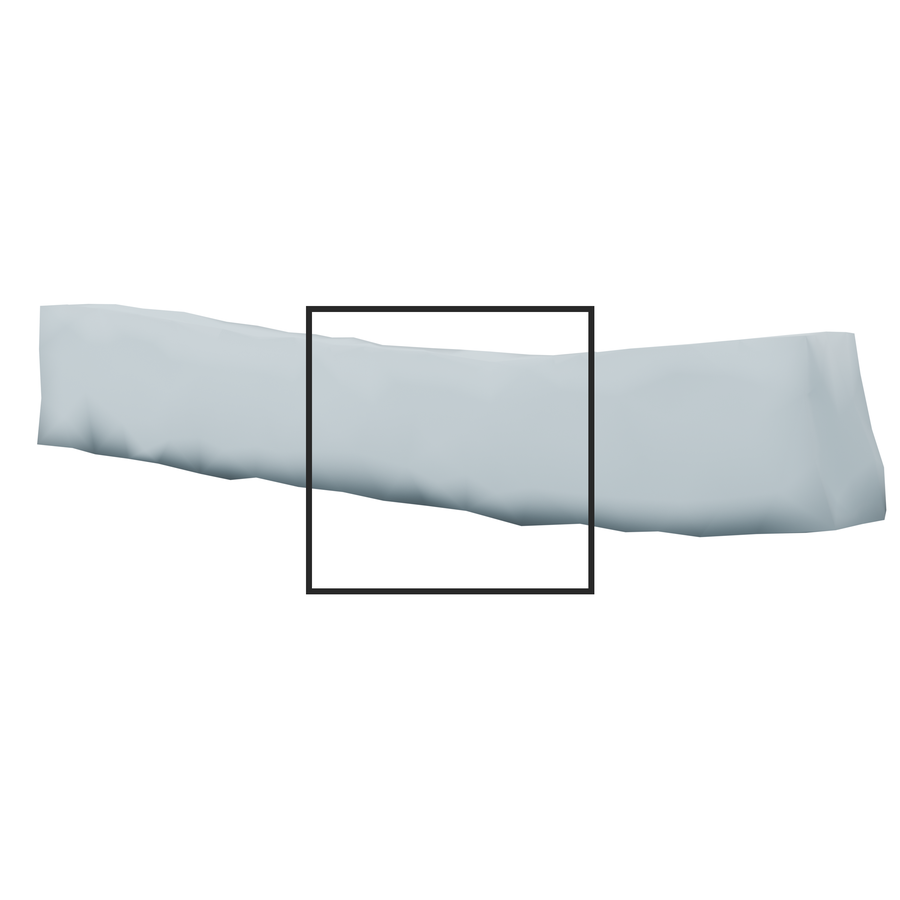} & \includegraphics[height=\qualh]{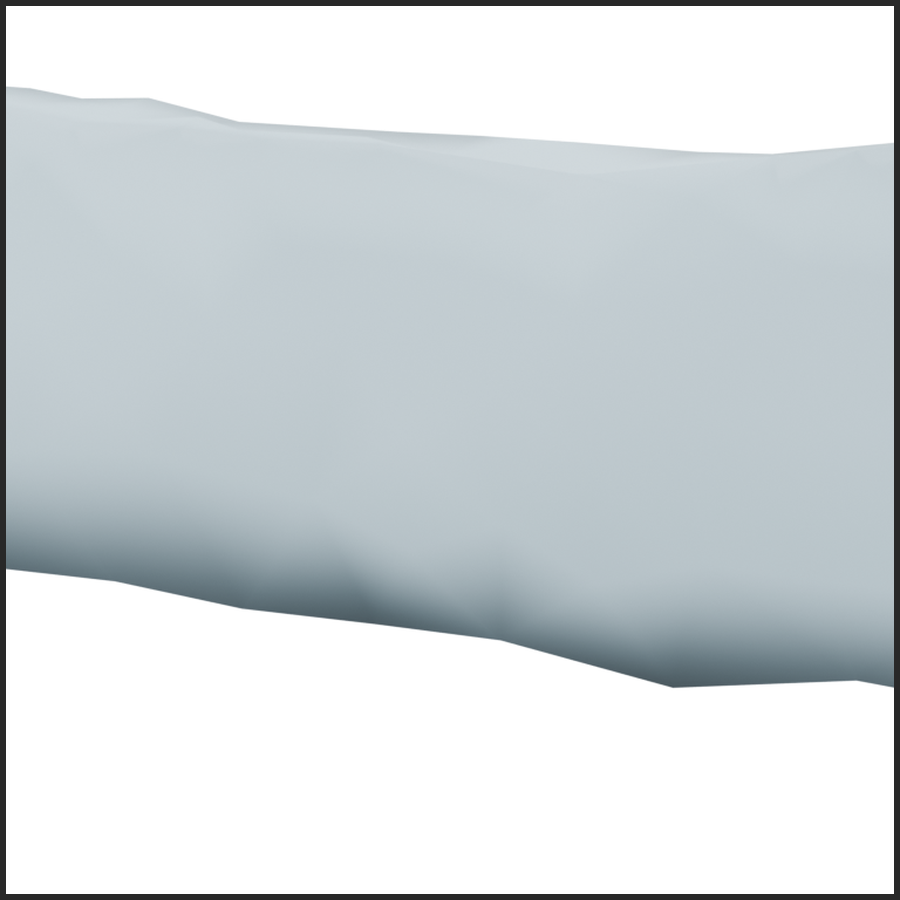} & \includegraphics[height=\qualh]{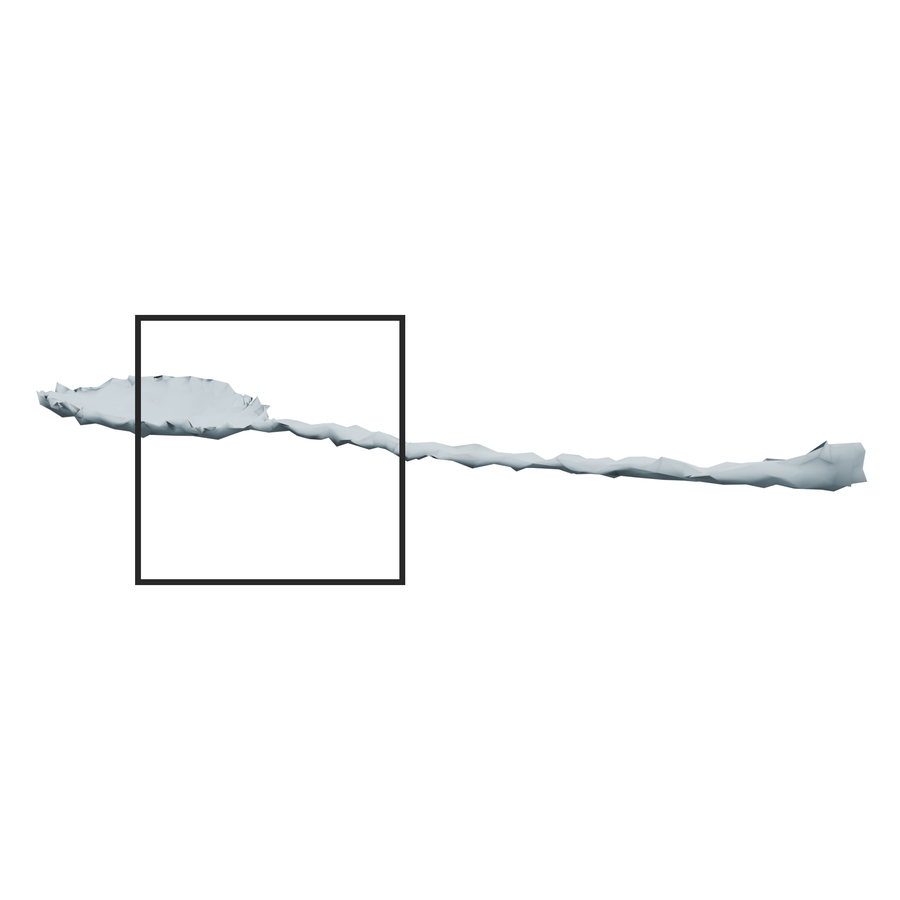} & \includegraphics[height=\qualh]{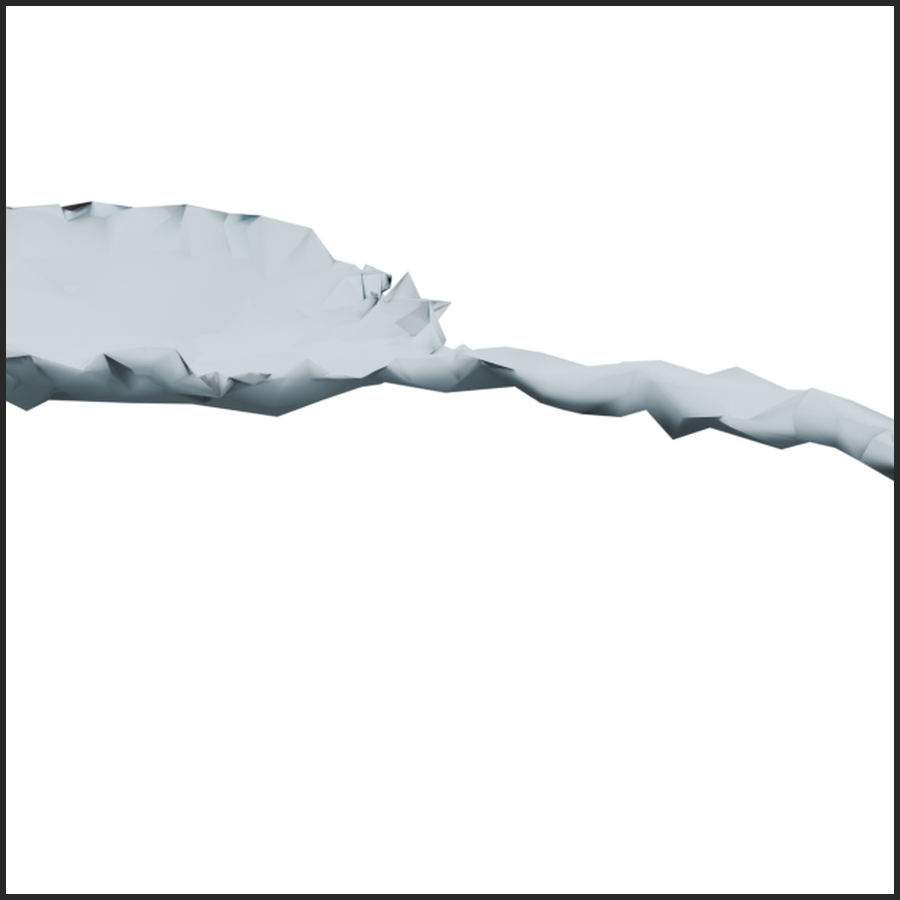} & \includegraphics[height=\qualh]{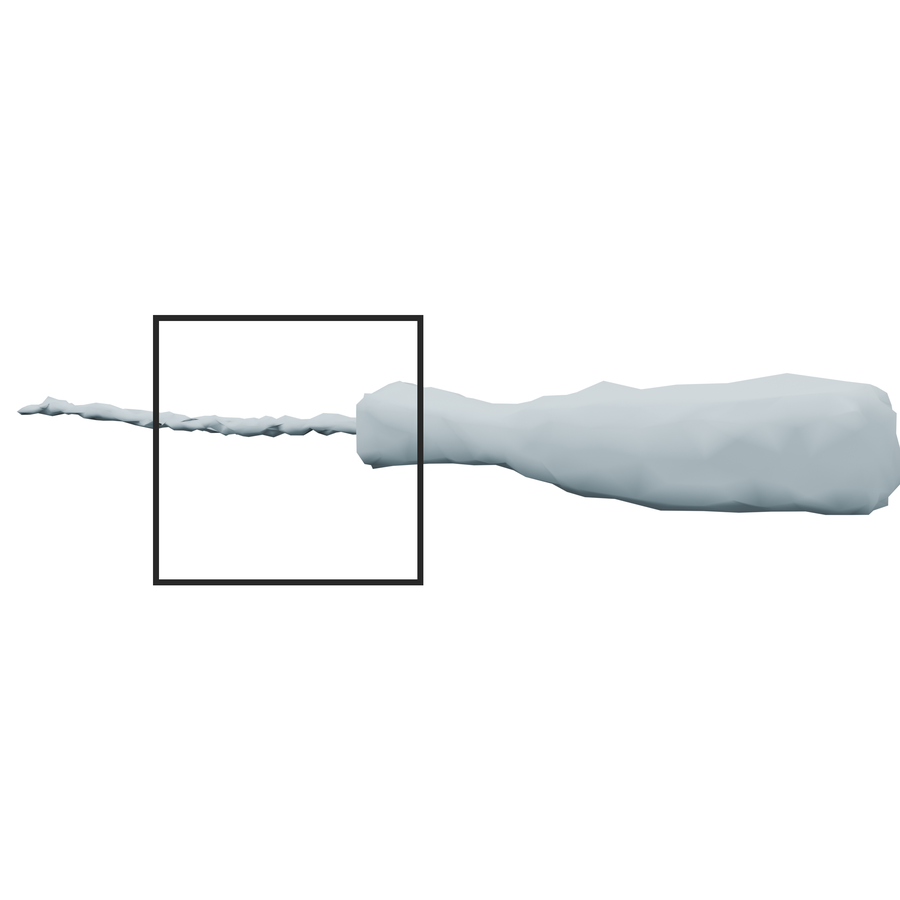} & \includegraphics[height=\qualh]{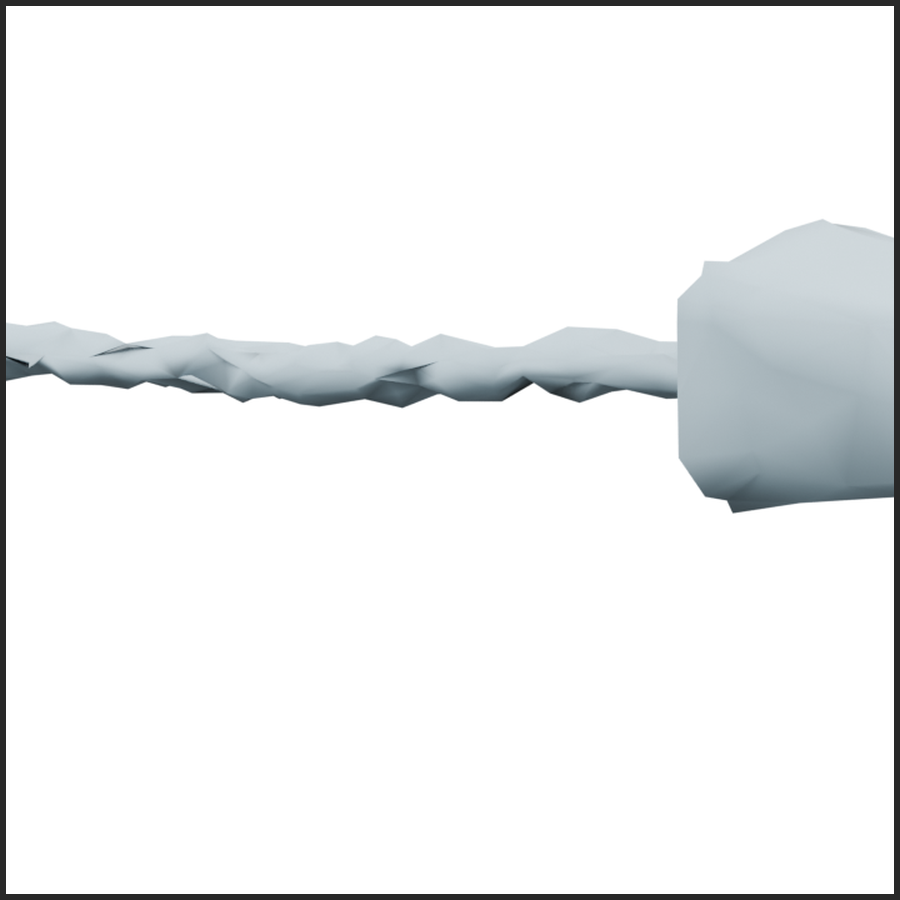} & \includegraphics[height=\qualh]{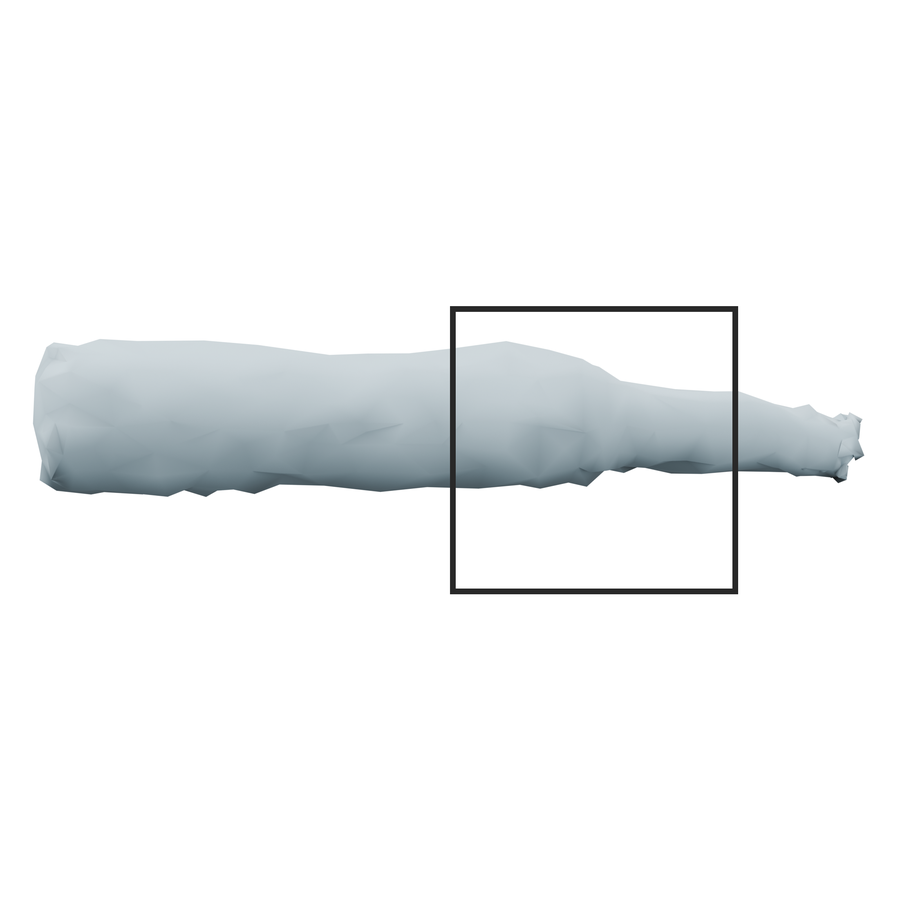} & \includegraphics[height=\qualh]{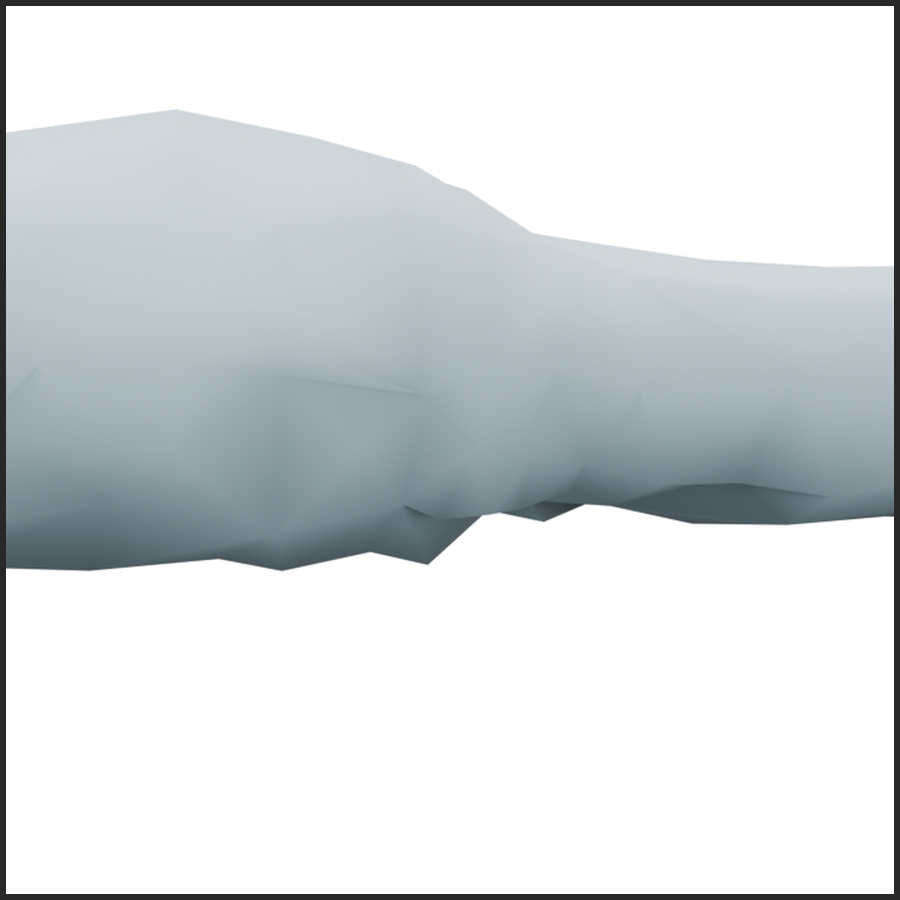} \\
\end{tabular}}

\caption{The two models on four held-out scenes (a block, a spoon, a screwdriver and a bottle; all $\leq 1{,}000$ vertices) at the last frame of the trajectory. For each scene the full object is followed by a zoom-in view of the boxed region. Rows: ground truth (rose), our three-stage pipeline trained on the same scenes with the same budget (Table~\ref{tab:pfcost}), and the concurrent DiT-B model. Our model preserves the geometric detail of every object. The concurrent model places the objects correctly, but its geometric detail is visibly noisy: the block bends and its faces ripple, the spoon and the shaft of the screwdriver are corrugated, and the bottle is lumpy. The zoom-in views show this noise on every surface.}
\label{fig:pfqual}
\end{figure*}

Fig.~\ref{fig:pfqual} compares the predictions of the two models on four held-out scenes at this resolution, shown at the last frame of the trajectory. Both models place the object roughly where the ground truth places it. The difference is in the geometric detail. The concurrent model produces surfaces that are visibly noisy: the block bends and its faces ripple, the spoon and the shaft of the screwdriver are corrugated, and the bottle is lumpy. The zoom-in views show this noise on every surface. Our model keeps the shape of every object and its surfaces stay smooth.

The reason lies in where each model generates geometry. Our ST-VAE is pretrained to reconstruct shapes, so the decoder maps any latent it receives to a clean shape on the learned manifold. A small error in the latent predicted by the LDM therefore becomes a small, coherent change of the shape, and the geometric detail is preserved. The concurrent model predicts every vertex coordinate directly, so there is nothing between its prediction and the geometry. A small error in any coordinate appears on the surface as is, and the many small errors of a diffusion model add up to noisy geometric detail.

This problem is not visible in the concurrent work's own demonstrations because meshes in their training data are simple and coarse, with at most a few hundred vertices, and the whole dataset is built from only a few dozen mesh templates. A model can learn these few simple shapes accurately with much less difficulty. On our data, with thousands of distinct objects at native resolution, learning the geometry itself becomes much harder.

\subsection{Inference cost comparison}
Our model is also much faster at inference. The concurrent model denoises $N\times 64$ vertex tokens, so its latency grows with the mesh, whereas our LDM denoises the fixed $4{,}104$ latent tokens. On a scene of about $1{,}000$ vertices, our model predicts a $64$-frame trajectory in about $1.5$\,s at $50$ sampling steps and $0.8$\,s at $20$, while the concurrent model needs about $12.9$\,s and $5.1$\,s, so ours is roughly $9\times$ and $6\times$ faster.

\FloatBarrier
\section{Training Across Energy Models and on Real Captured Trajectories}
\label{app:agnostic}
PhysLDM is trained with kinematic supervision only (Sec.~\ref{sec:training}). No energy model or material law enters the data format, the losses, or the architecture. This design gives PhysLDM three practical advantages, which we test in this section.
\begin{enumerate}
\item \textbf{Diverse energy models.} Because the constitutive model never enters training, the same pipeline can learn from simulations with any energy model.
\item \textbf{Data without an explicit energy model.} For the same reason, PhysLDM can learn from trajectories with only an implicit energy model, such as trajectories captured in the real world.
\item \textbf{Seamless support for other data formats, such as surface meshes and point clouds.} Our training data are tetrahedral meshes, but PhysLDM does not depend on the tetrahedra themselves. The spatial VAE only takes vertex positions and a graph over the vertices. For a tetrahedral mesh, the graph is given by the tetrahedral edges. A surface mesh fits this structure naturally, with its triangle edges as the graph. For a point cloud, we can use a $k$-nearest-neighbor graph or no edges at all. PhysLDM therefore runs on these formats without any change to the model or the training code. Because our input does not need to be a tetrahedral mesh, PhysLDM also supports object types beyond volumetric solids, such as cloth and rope, which are usually not represented with tetrahedra.
\end{enumerate}
We test these advantages in two steps. First, we apply the pretrained PhysLDM, trained only on Stable Neo-Hookean simulations of Objaverse meshes, to new data without any adaptation. Second, we fine-tune it on the new data with the unchanged training code.

\subsection{Data}
\label{app:agnostic_data}
\textbf{Diverse energy models.} We take $40$ held-out Objaverse scenes and simulate each scene twice with StiffGIPC from the same initial state. The first run uses the Stable Neo-Hookean energy of the training data \citep{DBLP:journals/tog/SmithGK18}. The second run uses an as-rigid-as-possible (ARAP) energy \citep{DBLP:conf/sgp/SorkineA07} with the same shear stiffness and no volume term, so the objects are softer and more compressible. The two energies move the trajectories $28$\,mm apart (median over scenes). We exclude the chaotic scenes of Appendix~\ref{app:chaos}, so this difference comes from the energy model and not from solver noise.

\textbf{Real captured point clouds.} We use $22$ real captured sequences from \citet{DBLP:conf/iccv/JiangHZYWL25}: $9$ soft volumetric objects (plush toys and a package), $8$ pieces of cloth, and $5$ ropes, each lifted, pushed, or stretched by a human hand on a table. The data are 3D points tracked over time with three RGB-D cameras, with $616$ to $7{,}826$ points per object. There is no mesh and no explicit energy model, and the forces of the human hand are not observed. The setting of these scenes is also very different from that of our training data. In our simulations, every object is released in the air with an initial velocity and falls onto the ground. In the captured sequences, the object rests on a table and is manipulated by a human hand. The real data are therefore out of domain for PhysLDM. We store the tracked points as the vertices and a $k$-nearest-neighbor graph ($k=8$) as the topology, in the same file format as our simulated data. PhysLDM is thus seamlessly compatible with these data, without any change to our code.

\textbf{Evaluation.} We use the metrics of Table~\ref{tab:s3phys} (Appendix~\ref{app:metrics_s3}): the volume violation rate, the deformation-to-rigid kinetic energy ratio, and the world-coordinate error, each reported as the median over scenes. The first two need tetrahedra and are not defined on the point clouds.

\subsection{Zero-shot results}
\label{app:agnostic_zs}
Table~\ref{tab:agnostic_zs} applies the pretrained PhysLDM to the new data without any adaptation.

\textbf{The ST-VAE generalizes zero-shot across energy models and to real point clouds.} On the simulated scenes, the reconstruction error is $4.5$\,mm with Stable Neo-Hookean and $4.6$\,mm with ARAP. On the real point clouds it is $6.0$\,mm. This is the most demanding test: the scenes show manipulation by a human hand instead of objects thrown onto the ground, the data are real captures instead of simulations, the input is a point cloud instead of a tetrahedral mesh, and cloth and rope never appear in training. Even on this out-of-domain data, the ST-VAE reconstructs the motion accurately without any adaptation. Although the ST-VAE was trained only on Stable Neo-Hookean simulations, it represents motion under an unseen energy model (ARAP) and real motion without an explicit energy model almost as accurately (Fig.~\ref{fig:phystwin}). This shows that its latent space is not tied to the energy model of the training data.

\textbf{The LDM prediction depends on the training data.} On the simulated scenes with Stable Neo-Hookean, the prediction error is $23.9$\,mm, in line with Table~\ref{tab:s3phys}. With ARAP, the error grows to $37.5$\,mm and the kinetic energy ratio drops to $0.79$: the LDM predicts the deformation of the material it was trained on, while the ARAP objects deform more. On the real sequences, the zero-shot prediction fails ($875$\,mm) because of the domain gap, not the energy model. The LDM has only learned objects thrown onto the ground, while the real object stays on the table until the human hand moves it.

\begin{table}[!htb]\centering\small
\caption{The pretrained PhysLDM on three data sources without adaptation. Medians over scenes. The volume and kinetic energy metrics need tetrahedra and are not defined on the point clouds.}
\label{tab:agnostic_zs}
\begin{tabular}{@{}l l c ccc@{}}
\toprule
Data & Output & Scenes & Vol.\ (\%)\,$\downarrow$ & $KE_{\mathrm{def}}/KE_{\mathrm{rig}}$\,$\to1$ & Err (mm)\,$\downarrow$ \\
\midrule
Stable Neo-Hookean & ST-VAE reconstruction & $40$ & $6.9$ & $0.92$ & $4.5$ \\
Stable Neo-Hookean & LDM prediction & $40$ & $8.9$ & $0.83$ & $23.9$ \\
ARAP & ST-VAE reconstruction & $40$ & $3.1$ & $0.93$ & $4.6$ \\
ARAP & LDM prediction & $40$ & $4.9$ & $0.79$ & $37.5$ \\
Real captured & ST-VAE reconstruction & $22$ & -- & -- & $6.0$ \\
Real captured & LDM prediction & $22$ & -- & -- & $874.9$ \\
\bottomrule
\end{tabular}
\end{table}

\subsection{Fine-tuning with the unchanged pipeline}
\label{app:agnostic_ft}
\begin{figure}[!htb]
\centering
\setlength{\tabcolsep}{2pt}\renewcommand{\arraystretch}{0.6}
\resizebox{\textwidth}{!}{\begin{tabular}{@{}c@{\,}c@{\hspace{5pt}}c@{}}
\rotatebox{90}{\scriptsize GT} & \makebox[0pt][l]{\includegraphics[height=\qualh]{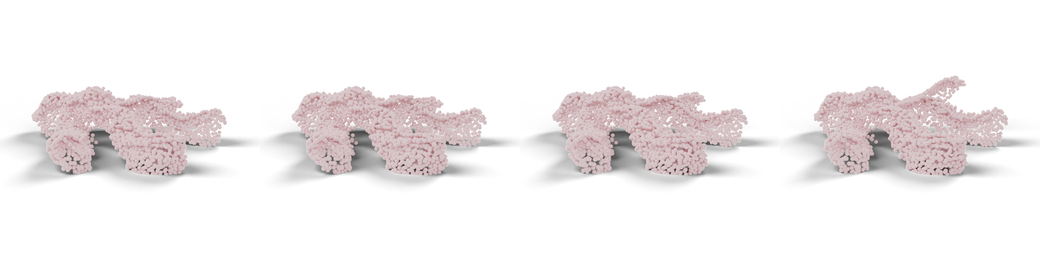}}\makebox[0pt][l]{\raisebox{1.5pt}{\hspace{1.5pt}\scriptsize $N{=}4460$}}\hspace{4.0000\qualh} & \makebox[0pt][l]{\includegraphics[height=\qualh]{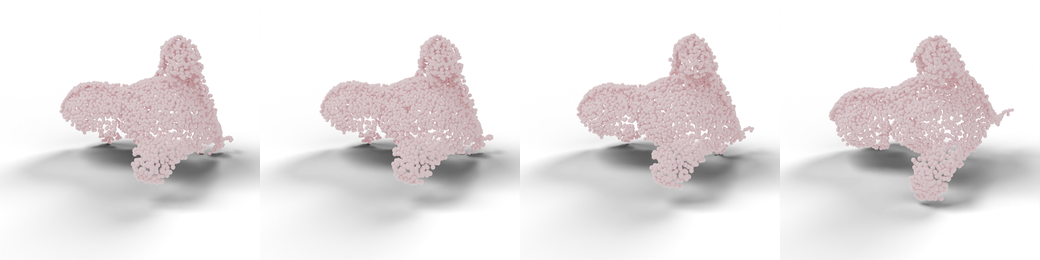}}\makebox[0pt][l]{\raisebox{1.5pt}{\hspace{1.5pt}\scriptsize $N{=}3967$}}\hspace{4.0000\qualh} \\
\rotatebox{90}{\scriptsize \shortstack{\textbf{ST-VAE}\\\textbf{recon.}}} & \includegraphics[height=\qualh]{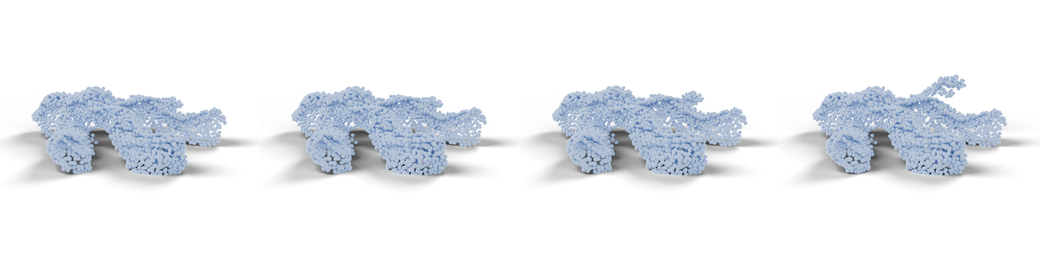} & \includegraphics[height=\qualh]{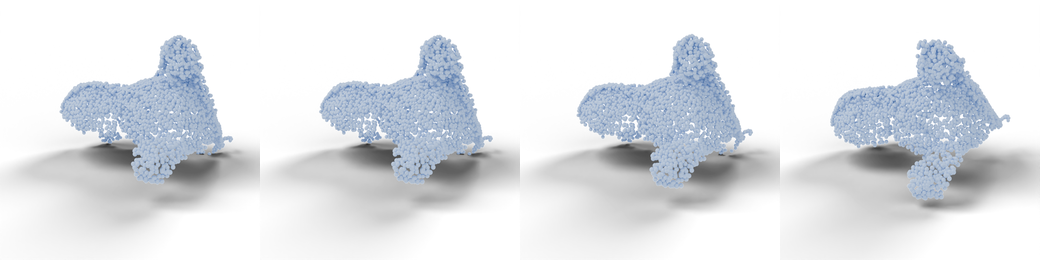} \\
\rotatebox{90}{\scriptsize \shortstack{LDM\\zero-shot}} & \includegraphics[height=\qualh]{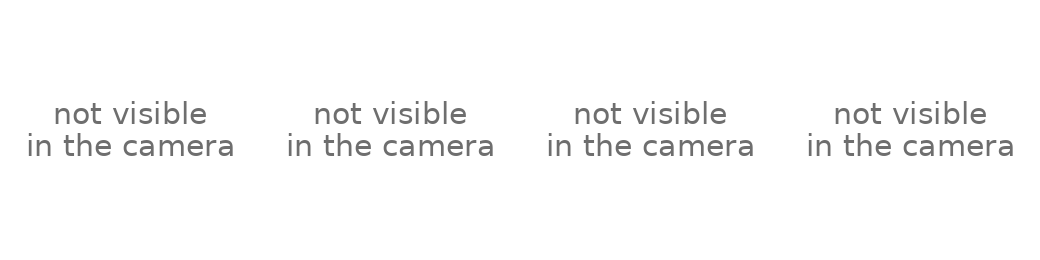} & \includegraphics[height=\qualh]{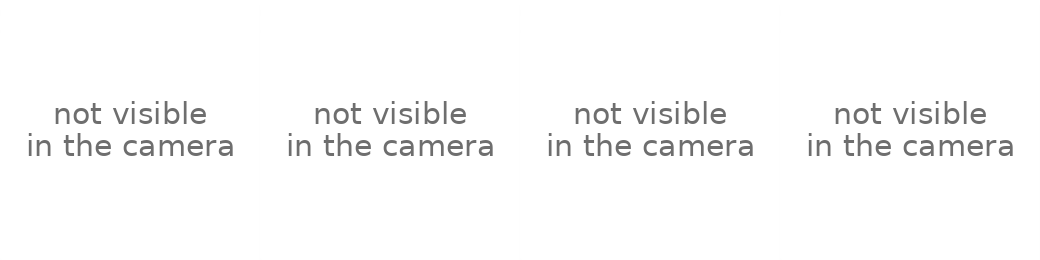} \\
\rotatebox{90}{\scriptsize \shortstack{LDM\\fine-tuned}} & \includegraphics[height=\qualh]{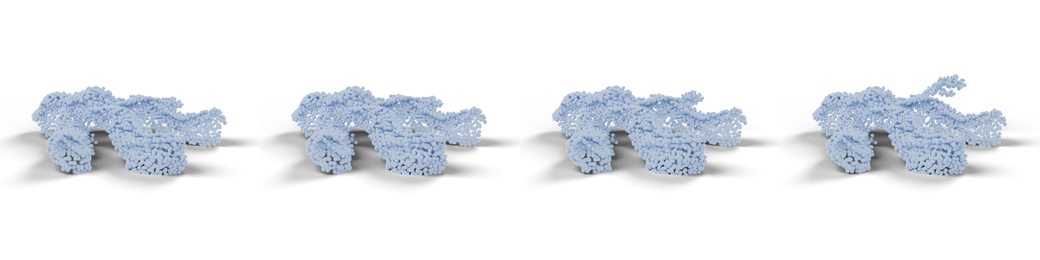} & \includegraphics[height=\qualh]{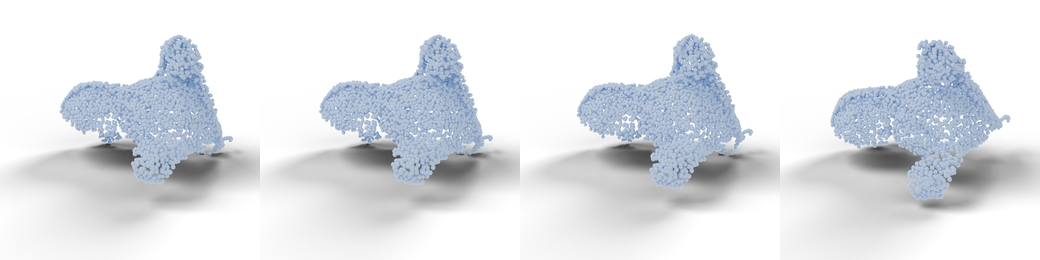} \\
\end{tabular}}\\[4pt]
\resizebox{\textwidth}{!}{\begin{tabular}{@{}c@{\,}c@{\hspace{5pt}}c@{}}
\rotatebox{90}{\scriptsize GT} & \makebox[0pt][l]{\includegraphics[height=\qualh]{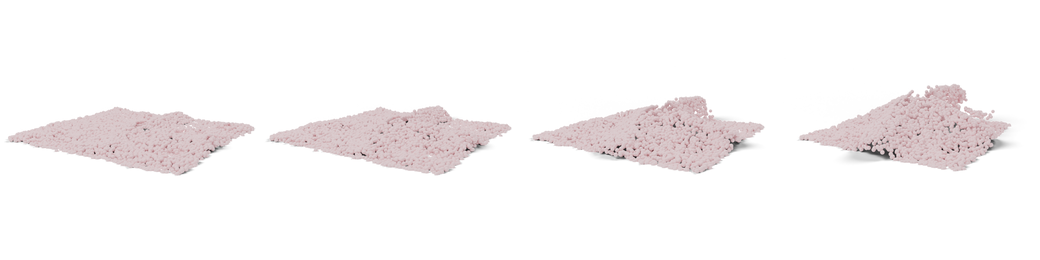}}\makebox[0pt][l]{\raisebox{1.5pt}{\hspace{1.5pt}\scriptsize $N{=}4848$}}\hspace{4.0000\qualh} & \makebox[0pt][l]{\includegraphics[height=\qualh]{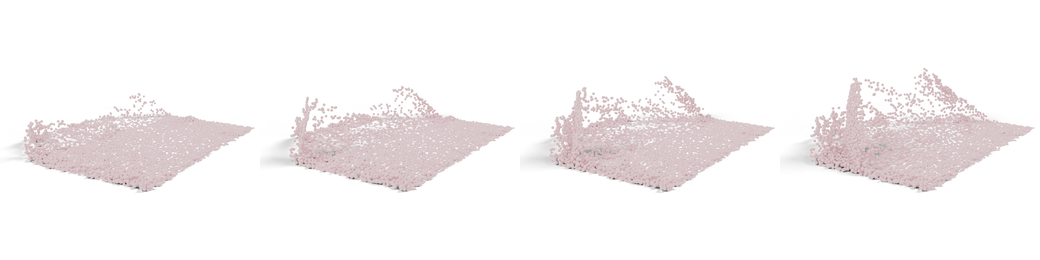}}\makebox[0pt][l]{\raisebox{1.5pt}{\hspace{1.5pt}\scriptsize $N{=}5944$}}\hspace{4.0000\qualh} \\
\rotatebox{90}{\scriptsize \shortstack{\textbf{ST-VAE}\\\textbf{recon.}}} & \includegraphics[height=\qualh]{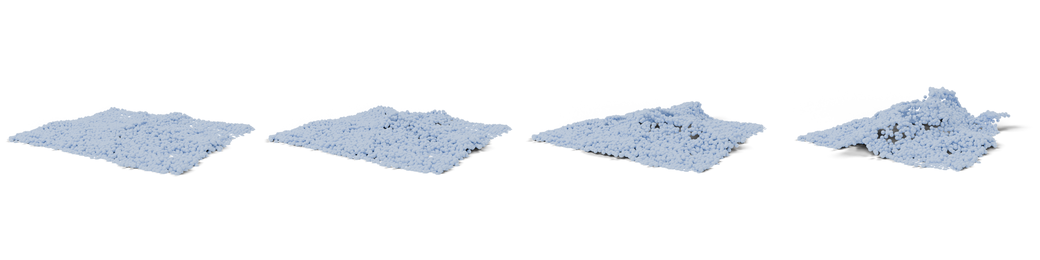} & \includegraphics[height=\qualh]{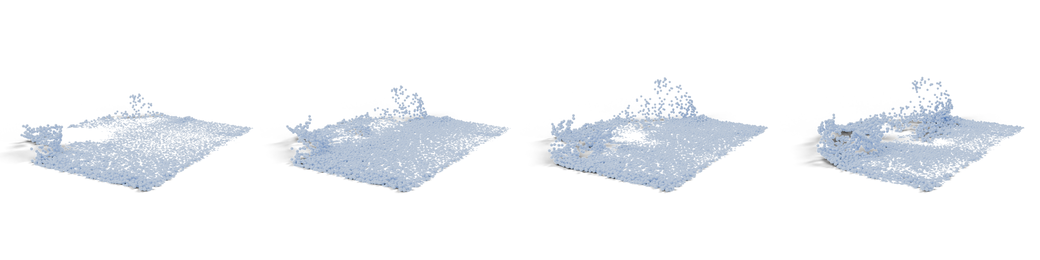} \\
\rotatebox{90}{\scriptsize \shortstack{LDM\\zero-shot}} & \includegraphics[height=\qualh]{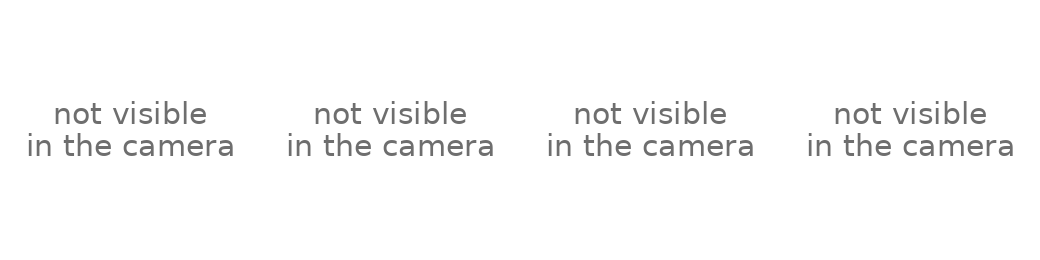} & \includegraphics[height=\qualh]{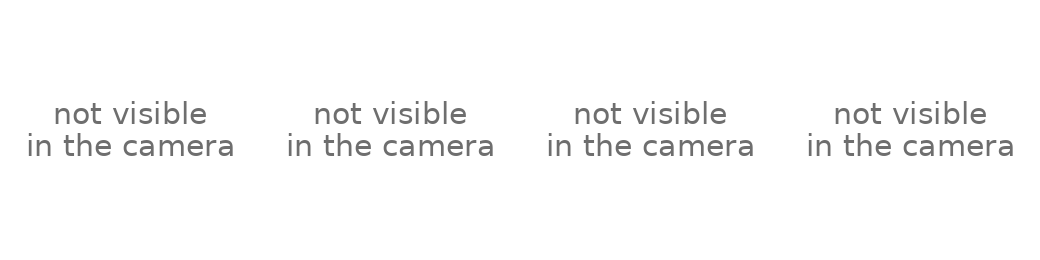} \\
\rotatebox{90}{\scriptsize \shortstack{LDM\\fine-tuned}} & \includegraphics[height=\qualh]{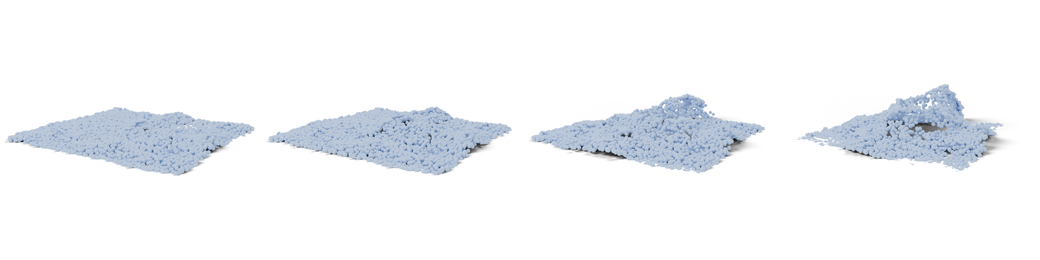} & \includegraphics[height=\qualh]{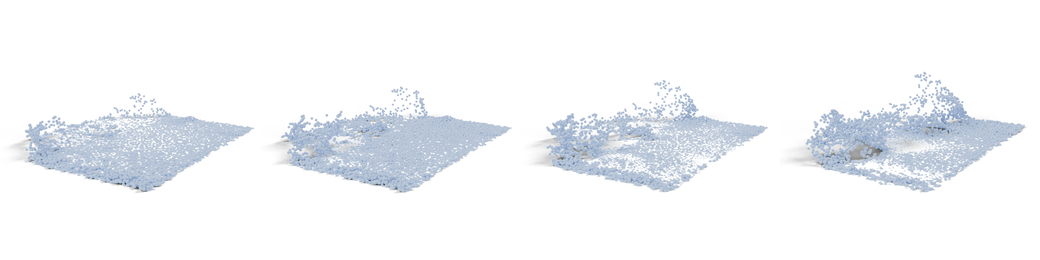} \\
\end{tabular}}\\[4pt]
\resizebox{\textwidth}{!}{\begin{tabular}{@{}c@{\,}c@{\hspace{5pt}}c@{}}
\rotatebox{90}{\scriptsize GT} & \makebox[0pt][l]{\includegraphics[height=\qualh]{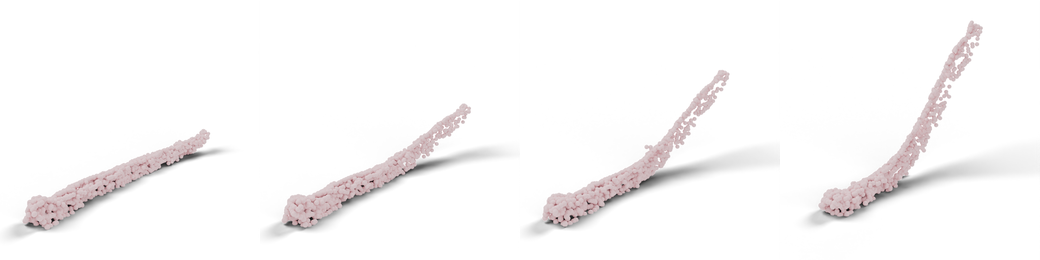}}\makebox[0pt][l]{\raisebox{1.5pt}{\hspace{1.5pt}\scriptsize $N{=}1127$}}\hspace{4.0000\qualh} & \makebox[0pt][l]{\includegraphics[height=\qualh]{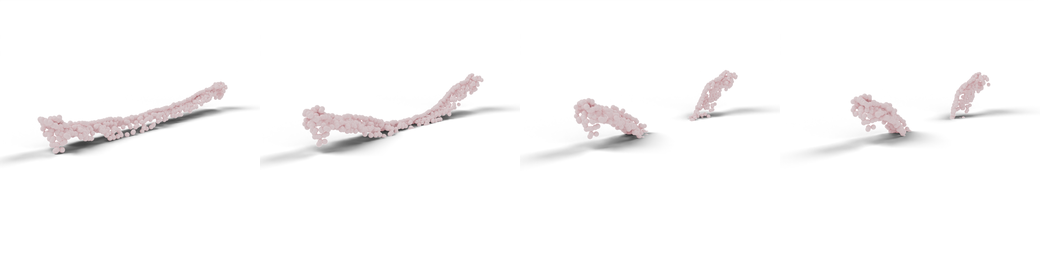}}\makebox[0pt][l]{\raisebox{1.5pt}{\hspace{1.5pt}\scriptsize $N{=}848$}}\hspace{4.0000\qualh} \\
\rotatebox{90}{\scriptsize \shortstack{\textbf{ST-VAE}\\\textbf{recon.}}} & \includegraphics[height=\qualh]{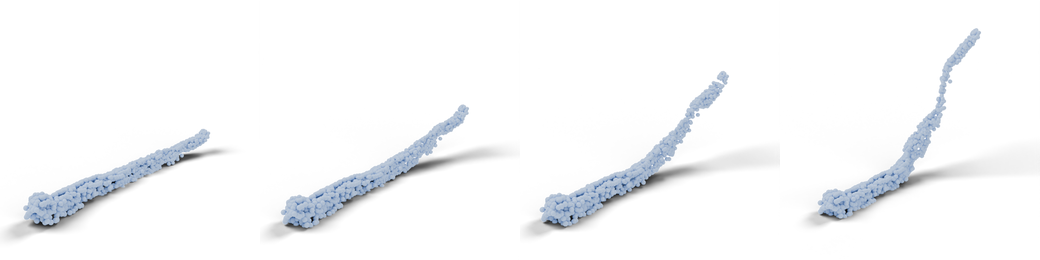} & \includegraphics[height=\qualh]{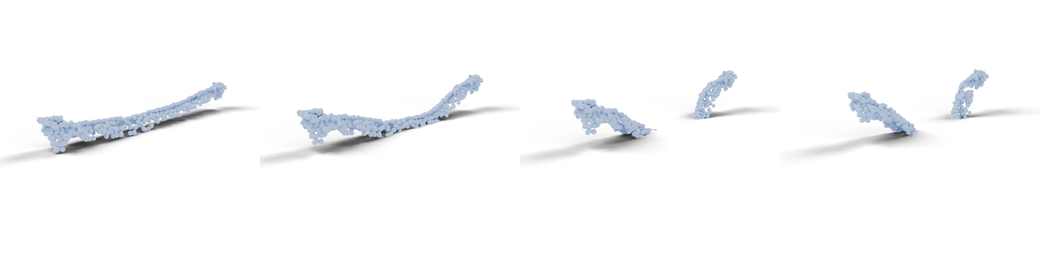} \\
\rotatebox{90}{\scriptsize \shortstack{LDM\\zero-shot}} & \includegraphics[height=\qualh]{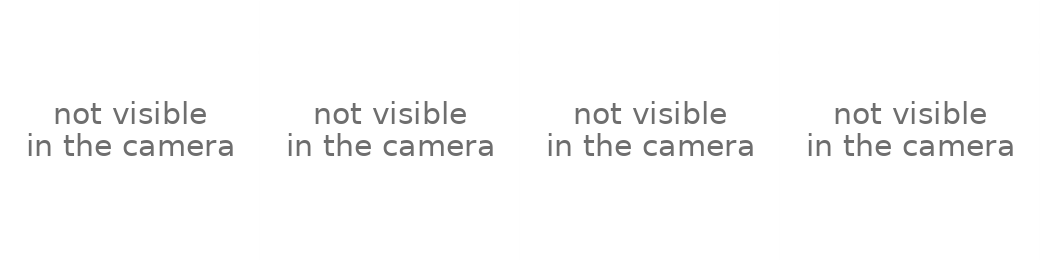} & \includegraphics[height=\qualh]{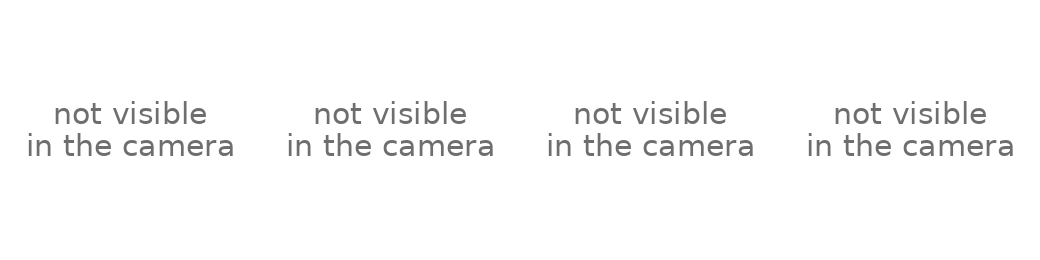} \\
\rotatebox{90}{\scriptsize \shortstack{LDM\\fine-tuned}} & \makebox[0pt][l]{\includegraphics[height=\qualh]{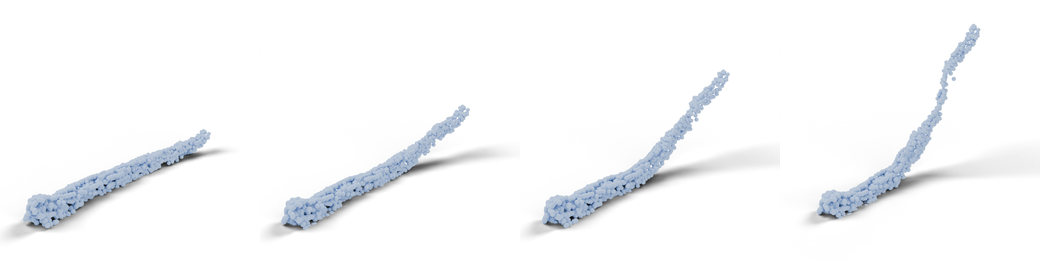}}\makebox[0pt][l]{\hspace{3pt}\tikz[baseline=0pt]{\node[inner sep=0pt,inner xsep=3pt,font=\small\itshape,text=black!65] (t) at ($(2.0000\qualh,0)+(-1.5pt,0)$) {time};\draw[black!65,line width=0.8pt,line cap=round] (0,0) -- (t.west);\draw[black!65,line width=0.8pt,line cap=round,-{Straight Barb[length=5.5pt,width=6.5pt,angle'=60]}] (t.east) -- ($(4.0000\qualh,0)+(-3pt,0)$);}}\hspace{4.0000\qualh} & \includegraphics[height=\qualh]{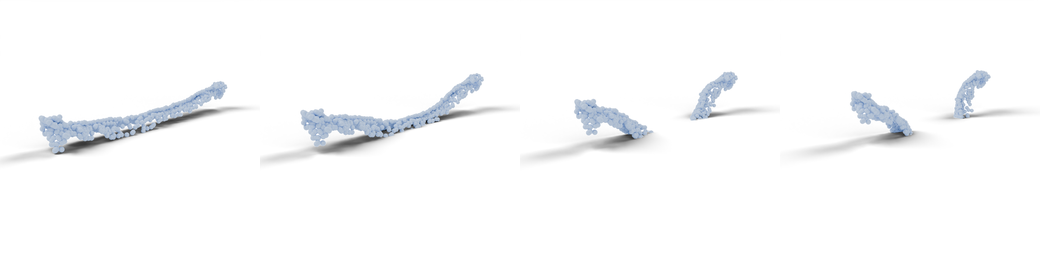} \\
\end{tabular}}

\caption{PhysLDM on real captured point clouds \citep{DBLP:conf/iccv/JiangHZYWL25}: a plush sloth and a plush dinosaur (top), two pieces of cloth (middle), and two ropes (bottom), each at four frames sampled along the trajectory and rendered as points. Rows of each block: the tracked points (rose), the ST-VAE reconstruction, the LDM prediction of the pretrained model without adaptation (zero-shot), and the LDM prediction after fine-tuning (Table~\ref{tab:agnostic_ft}). The zero-shot prediction lets the resting object fall through the table, so it is not visible in the camera.}
\label{fig:phystwin}
\end{figure}

To adapt PhysLDM to new data, we only replace the training data with the new trajectories and fine-tune the LDM from the pretrained weights with the same trainer. The ST-VAE stays frozen. The same procedure applies to both energy models and to the real point clouds. Table~\ref{tab:agnostic_ft} reports the results on the fine-tuning subsets, so the values before fine-tuning differ slightly from Table~\ref{tab:agnostic_zs}. For both explicit energy models, the prediction error drops by about two thirds, to $7.6$\,mm and $8.8$\,mm. For ARAP, the kinetic energy ratio moves from $0.80$ to $0.92$, so the LDM learns the softer material from the trajectories alone. On the real point clouds, the error drops from $921$ to $6.2$\,mm: the LDM learns from the tracked points that a resting object stays on the table and how the points move (Fig.~\ref{fig:phystwin}).

In summary, the same PhysLDM, with the same code and losses, handles simulations with different energy models, real captured trajectories without an explicit energy model, and other data formats such as point clouds. Its ST-VAE transfers to all of them zero-shot, and its LDM adapts to them by fine-tuning. Visual results of PhysLDM on the real captured data are shown in Fig.~\ref{fig:phystwin}.

\begin{table}[!htb]\centering\small
\caption{Fine-tuning the LDM of PhysLDM on each data source with the unchanged trainer. Medians over the scenes of each fine-tuning subset, before and after fine-tuning.}
\label{tab:agnostic_ft}
\begin{tabular}{@{}l c cc cc cc@{}}
\toprule
& & \multicolumn{2}{c}{Vol.\ (\%)\,$\downarrow$} & \multicolumn{2}{c}{$KE_{\mathrm{def}}/KE_{\mathrm{rig}}$\,$\to1$} & \multicolumn{2}{c}{Err (mm)\,$\downarrow$} \\
\cmidrule(lr){3-4}\cmidrule(lr){5-6}\cmidrule(lr){7-8}
Data & Scenes & before & after & before & after & before & after \\
\midrule
Stable Neo-Hookean & $32$ & $8.9$ & $6.2$ & $0.82$ & $0.91$ & $22.9$ & $7.6$ \\
ARAP & $32$ & $4.9$ & $3.1$ & $0.80$ & $0.92$ & $34.4$ & $8.8$ \\
Real captured & $16$ & -- & -- & -- & -- & $921.3$ & $6.2$ \\
\bottomrule
\end{tabular}
\end{table}

\end{document}